\documentclass{article}

\usepackage[preprint]{neurips_2025}

\usepackage[utf8]{inputenc} 
\usepackage[T1]{fontenc}    
\usepackage{url}            
\usepackage{booktabs}       
\usepackage{amsfonts}       
\usepackage{nicefrac}       
\usepackage{microtype}      
\usepackage{xcolor}         
\usepackage{algorithm, algorithmic}
\usepackage{wrapfig}
\usepackage{setspace}
\usepackage{cancel}
\usepackage[backref]{hyperref} 
\usepackage{url}
\let\AND\relax
\usepackage{amsmath, amsfonts, bm, amsthm, amssymb, relsize, bbm, graphicx}
\usepackage[capitalize]{cleveref}
\usepackage{algorithm, algorithmic}
\usepackage{cancel}
\usepackage{subcaption}
\usepackage{setspace}
\usepackage{enumitem}
\usepackage{multirow} 
\usepackage[dvipsnames]{xcolor}
\usepackage{placeins}
\hypersetup{
    colorlinks = true,
    linkcolor  = BrickRed,
    urlcolor   = teal,
    citecolor  = RoyalBlue
}

\def\d{{\mathrm{d}}}
\newcommand{\bb}[1]{\mathbb{#1}}
\newcommand{\x}[1]{\boldsymbol{#1}}
\newcommand{\dX}[1]{\mathrm{d}{#1}}
\def\id{{\bb{I}}}
\def\R{{\bb{R}}}

\def\E{{\bb{E}}}
\def\P{{\bb{P}}}
\def\d{{\mathrm{d}}}
\def\X{{\x{X}}}
\def\Y{{\x{Y}}}
\def\Z{{\x{Z}}}
\def\N{{\mathcal{N}}}

\def\1{{\mathbbm{1}}}
\def\Xt{{\x{X}_t}}

\def\Yt{{\x{Y}_t}}

\def\Wt{{\x{W}_t}}
\def\Xk{{\x{X}_k}}
\def\Xprev{{\x{X}_{k-1}}}
\def\Xnext{{\x{X}_{k+1}}}

\def\Zk{{\x{Z}_k}}
\def\scox{{\nabla \log p_{\Xt}(\Xt, t)}}
\def\scoxk{{\nabla \log p_{\x{X}_{t}}(\x{X}_k, k\delta)}}
\def\scoy{{\nabla \log p_{\Yt}(\Yt, t)}}
\def\normal{{\mathcal{N}(\x{0}, \bb{I})}}
\theoremstyle{plain}
\newtheorem{theorem}{Theorem}[section]

\theoremstyle{definition}

\theoremstyle{remark}

\title{Dale meets Langevin: A Multiplicative Denoising Diffusion Model}

\author{%
  Nishanth Shetty\\
  Department of Electrical Engineering\\
  Indian Institute of Science\\
  Bengaluru 560012 \\
  \texttt{nishanths@iisc.ac.in} \\
  \And
  Madhava Prasath \\
  Department of Electrical Engineering\\
  Indian Institute of Science\\
  Bengaluru 560012 \\
  \texttt{madhavprasath088@gmail.com} \\
  \AND
  Chandra Sekhar Seelamantula \\
  Department of Electrical Engineering\\
  Indian Institute of Science\\
  Bengaluru 560012 \\
  \texttt{css@iisc.ac.in}
}

\begin{document}

\maketitle

\begin{abstract}
Exponentiated gradient descent (EGD), a biologically motivated optimization algorithm that respects Dale's law, produces log-normally distributed synaptic weights at convergence, in alignment with experimental observations in neuroscience. Since the marginal distribution of geometric Brownian motion (GBM) at any fixed time is log-normal, this convergence property reveals a natural connection between EGD and GBM-based stochastic processes. We propose a multiplicative score-based generative model with GBM as a forward noising process and derive its corresponding reverse-time SDE in both the ambient space and in the $\log$-transformed space. We derive two multiplicative samplers by discretizing the corresponding reverse-time SDEs: a \emph{sign-agnostic} sampler obtained directly from the ambient-space reverse-time SDE, and a \emph{sign-preserving} sampler, which we refer to as the \textit{Dale-Langevin sampler}, obtained via the Lamperti transform. We further connect the framework to Mirrored Langevin Dynamics, showing that the convex generator driving EGD in optimization precisely governs the Dale-Langevin sampler. While the standard {\it Stein score}, defined as $\nabla \log p_{\x{X}}(\x{x})$ for a random vector $\x{X}$ evaluated at $\x{x}$, comes up naturally in the additive noise based diffusion models, in the multiplicative setting, we encounter a modified version of the Stein score for sampling, which we refer to as the {\it Hyv\"arinen score}: $\x{x} \circ \nabla \log p_{\x{X}}(\x{x})$. In order to estimate the Hyv\"arinen score, we propose a new multiplicative denoising score-matching objective (M-DSM) and prove its equivalence to the multiplicative explicit score-matching loss (M-ESM). This development subsumes the non-negative score matching loss of \citet{Hyvaerinen2007} as a special case. Experimental results on MNIST, Fashion-MNIST, Kuzushiji-MNIST, and CIFAR-10 validate the generative capability of the proposed framework.
\end{abstract}

\section{Introduction}
\label{sec:intro}

Several studies in computational neuroscience \citep{song2005, loewenstein2011multiplicative, Buzsaki2014, Melander2021, pogodin2024synaptic} have confirmed that synaptic weight distributions in biological neural networks are log-normally distributed, and that neurons obey Dale's law~\citep{Eccles1954}, which states that a neuron is either excitatory or inhibitory, and does not flip between the two states during learning. Standard gradient descent does not guarantee sign preservation. Recent efforts to define mathematical structures that respect Dale's law have been fruitful. In particular, \citet{cornford2024} showed that exponentiated gradient descent (EGD) applied to neural network training respects Dale's law and produces log-normally distributed weights. EGD is the canonical instance of mirror descent \citep{bubeck2015} with the negative entropy as the mirror map, and its natural domain is $\mathbb{R}^d_+$ rather than $\mathbb{R}^d$.

Sampling from probability density functions and optimization are deeply intertwined and can be viewed as two sides of the same coin. It has been shown that {\it Langevin dynamics} is gradient flow in the space of probability measures under the Wasserstein-2 metric, and the corresponding discrete-time update matches with {\it gradient descent} on the Kullback-Leibler (KL) divergence~\citep{Wibisono18a}. This raises the question: \textit{What is the sampling analogue of EGD, in the same way that Langevin dynamics can be viewed as the sampling analogue of gradient descent?} In this paper, we answer that question, and in doing so propose a multiplicative generative model and a corresponding multiplicative score-matching loss, establish the link with EGD through Mirrored Langevin dynamics (MLD)~\citep{mld} and experimentally demonstrate image generation within the proposed framework.

We start with geometric Brownian motion (GBM), a diffusion process that preserves positivity, and has a marginal distribution that is log-normal~\citep{Oksendal1985}. We derive the reverse-time stochastic differential equation (SDE) in the ambient space and the $\log$-transformed space and derive two multiplicative samplers: a \emph{sign-agnostic} sampler obtained directly from the ambient-space reverse-time SDE, and a \emph{sign-preserving} sampler, which we refer to as the \textit{Dale-Langevin sampler}, obtained via the Lamperti transform and an appropriate change of variables. The Dale-Langevin sampler preserves the sign of every coordinate across all iterations by scaling samples in each update by a score-dependent factor. This is the sampling analogue of EGD, in the same sense that Langevin dynamics is the sampling analogue of gradient descent. Additionally, we show that the convex function driving EGD in optimization precisely matches the one used in the MLD framework to derive the Dale-Langevin sampler. 

In standard additive noise based diffusion models~\citep{DDPM20, NCSNv3_21}, the {\it Stein score} comes up naturally during sampling. In the multiplicative setting, the Stein score does not suffice for sampling. We require a modified version of the Stein score, which we refer to as the {\it Hyv\"arinen score}. In order to estimate the Hyv\"arinen score, we propose a new multiplicative denoising score-matching objective (M-DSM) and prove its equivalence to the multiplicative explicit score-matching loss (M-ESM) up to a constant. This is analogous to the equivalence between the denoising score-matching loss and the associated explicit score matching loss in the additive case~\citep{PascalVincent2011}.

\begin{figure}[t]
    \centering
    \includegraphics[width=0.9\linewidth]{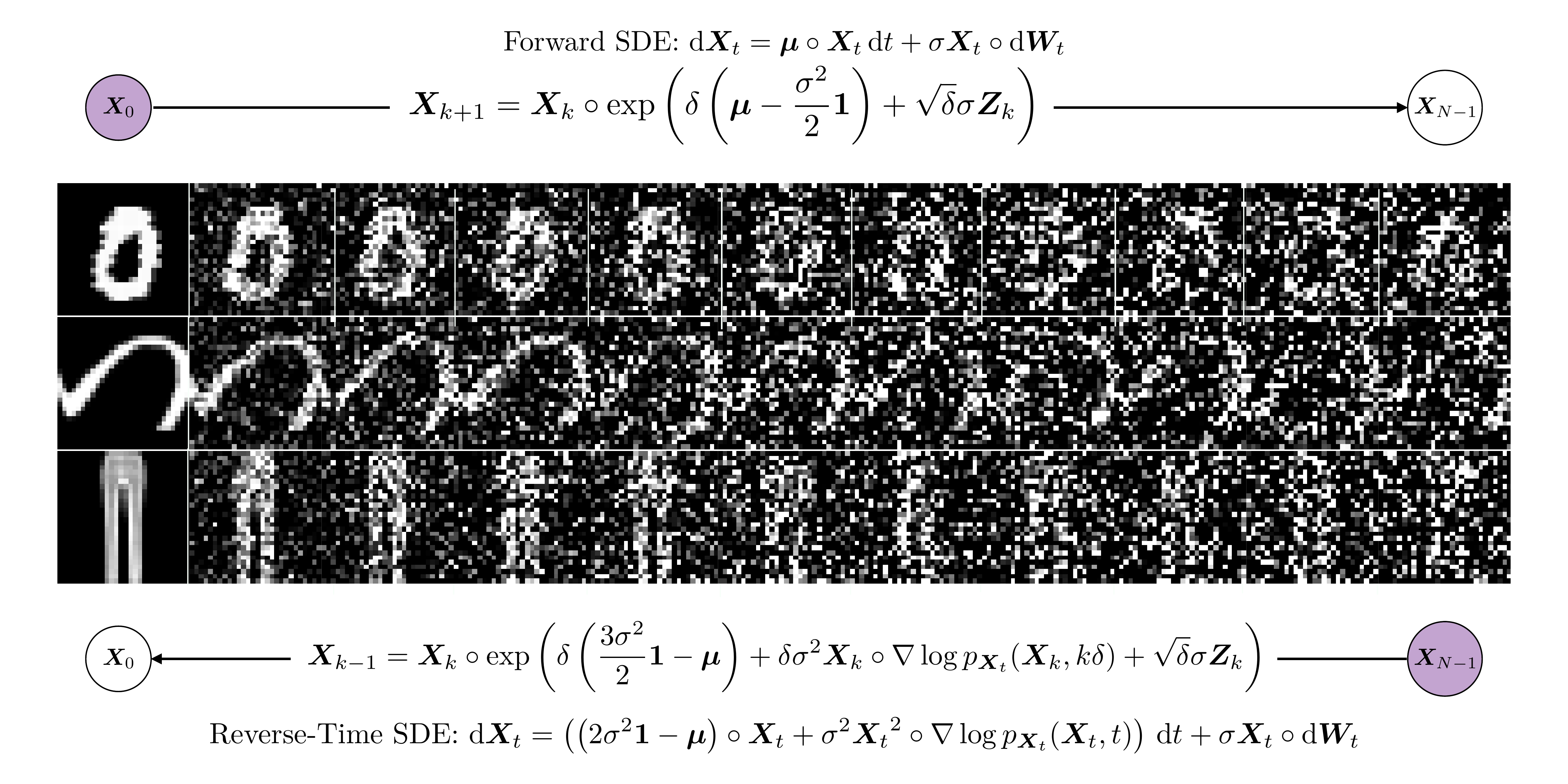}
    \caption{The forward and reverse-time SDEs for Geometric Brownian Motion (GBM). The forward SDE describes the evolution of a clean image sample to a noisy one that eventually becomes log-normally distributed, while the reverse-time SDE captures the dynamics of the process and generates new samples from the unknown density starting from log-normal noise. This is enabled by the knowledge of the unknown density manifesting through the score function.}
    \label{fig:gbm-gen-schematic}
    \vspace{-1.5em}
\end{figure}

The score-matching loss proposed by \citet{Hyvaerinen2007} for non-negative data emerges as a special case of M-ESM at $t = 0$. We validate the resulting algorithm on MNIST~\citep{lecun-mnisthandwrittendigit-2010}, Fashion-MNIST \citep{xiao2017/online}, Kuzushiji-MNIST \citep{Clanuwat2018}, and CIFAR-10 ~\citep{cifar10}. \cref{fig:gbm-gen-schematic} provides a schematic of the generation and noising process.

\subsection{Related Works}
Recent advances in generative modelling with score-based generative models~\citep{DDPM20, NCSNv1_19, NCSNv2_20, NCSNv3_21} have yielded impressive results across modalities such as images, video, and audio. The dynamical systems based diffusion probabilistic model of~\citet{sohl-dickstein15}, inspired by non-equilibrium thermodynamics, provided the foundational framework for these advances. \citet{DDPM20} were the first to demonstrate the generative power of denoising diffusion probabilistic models (DDPMs) by achieving high-quality image synthesis competitive with the then state-of-the-art generative models~\citep{karras2018progressive, Karras2020}. Progress in score-matching~\citep{song2019sliced, NCSNv1_19, NCSNv2_20, DMsbeatGANs2021} showed score-based models to outperform Generative Adversarial Networks (GANs)~\citep{SGAN14, StyleGANv1_19, StyleGANv2_20, StyleGAN3}. \citet{song2021scorebased} unified the discrete-time conditional density framework of \citet{DDPM20} with the SDE framework built on standard Brownian motion. Several works have since explored departures from Brownian motion. \citet{Bansal2023} formulate generative models around generic degradation operators such as blurring, masking, etc. and their corresponding restoration maps. \citet{rissanen2023} view generation as the time-reversal of a heat equation, identify disentanglement of shape and colour, and analyze spectral inductive biases. \citet{Santos2023} develop a discrete state-space diffusion via a pure-death random process.\par

The Gaussian noise assumption, while analytically convenient, is not compatible with data carrying structural constraints such as positivity, boundedness, or discreteness. \citet{nachmani2021denoisingdiffusiongammamodels} factor the positivity constraint by replacing the additive Gaussian noise with additive Gamma noise. \citet{zhou2023beta} propose Beta Diffusion for interval-valued data. To handle count data, \citet{2023poisson_diffusion_model} build a Poisson-process framework. Deviating from the Euclidean paradigm, \citet{fishman2023diffusion} provide a systematic construction of diffusion models on Riemannian manifolds. \citet{vuong2024perceptionbasedmultiplicativenoiseremoval} also consider the GBM SDE and solve a multiplicative denoising problem by converting it to an additive one using the $\log$ transform. They leverage the corresponding reverse-time SDE for sequential denoising but do not explore the potential for image generation.
 
We explain why the $\log$ transform is appropriate, and additionally close the generative loop with reverse-time SDEs and multiplicative samplers for unconditional generation. More recently, \citet{gruhlke2026multiplicative} independently proposed multiplicative diffusion motivated by skew-symmetric multiplicative noise working with Stratonovich interpretation of the SDE. Their framework is motivated by physical principles of conservative forward-backward dynamics to preserve the distribution of norm of the data and show that their framework is better suited for heavy-tailed and anisotropic distributions. They train their score model with standard sliced score-matching, which is inherently limited to lower dimensions. Their experiments are focussed on fluid dynamics and not image generation. In contrast, we consider the GBM SDE, where the marginal density of the forward process at any time instant is log-normal. Our score-matching formalism is new and scales well with the dimension of the data, as evidenced by results on image generation. 

Score matching was originally proposed as a technique for estimation of parameters of model densities~\citep{Hyvarinen05a}. More recently, for estimation of graphical models with densities from the exponential family, \citet{yu_generalized} generalize Hyv\"arinen's score-matching loss for non-negative data. \citet{truncated_liu} study score matching for densities defined on truncated domains. They do not address the question of generative modelling.

\subsection{Main Contributions}
\label{ssec:contributions}
We introduce a multiplicative diffusion framework built on GBM and derive reverse-time SDEs both in the ambient space (without the $\log$-transform) and in the $\log$-transformed space resulting in corresponding multiplicative samplers (\cref{sec:GBM}). We encounter a variant of Stein score in the multiplicative setting, which we refer to as the Hyv\"arinen score. We propose multiplicative explicit and denoising score matching losses and prove that they are equivalent (\cref{thm:gbm_sm}), thus, generalizing the score-matching loss proposed by \citet{PascalVincent2011} in the additive noise setting, to a multiplicative one. We also show that the non-negative score-matching loss \citep{Hyvaerinen2007} emerges as a special case of our formalism (\cref{sec:msm}). We further show that the convex function driving exponential gradient descent in optimization is identical to what results in the Dale-Langevin sampler under the Mirrored Langevin Dynamics~\citep{mld} framework (\cref{asec:Dale}). Finally, we present experimental validation on MNIST, Fashion-MNIST, Kuzushiji-MNIST, and CIFAR-10 datasets (\cref{sec:experiments}). The GBM forward process, the reverse-time SDEs, the multiplicative samplers,  an appropriately defined score-matching loss to train neural networks, demonstration on image generation, together constitute a complete generative framework within the multiplicative scenario.\par

\subsection{Notation}
\label{asec:notation}
Random variables are denoted in uppercase (e.g., $X, Y$), and random vectors in boldface uppercase (e.g., $\x{X}, \x{Y}$). Their realizations are denoted by the corresponding lowercase letters (e.g., $x, \x{x}$). The probability density function (p.d.f.) of a random variable $X$ is written as $p_{X}(x)$; for a random vector $\x{X}$, it is written as $p_{\x{X}}(\x{x})$. The conditional p.d.f.\ of $\x{X}_t$ given $\x{X}_0$ is written $p_{\x{X}_t \mid \x{X}_0}(\x{x}_t \mid \x{x}_0)$. The domain is $\R_{+}^{d}$ for all random vectors throughout the paper. The Stein score of the random vector $\x{X}$ evaluated at $\x{x}$ is $\nabla \log p_{\x{X}}(\x{x})$. The notation $s_{\x{\theta}}(\x{X}_t, t)$ denotes a neural network approximation of the score at time $t$. The symbol $\circ$ denotes element-wise (Hadamard) multiplication. $\x{1}$ denotes a vector of all ones, and $\mathbb{I}$ denotes the identity matrix of appropriate dimension. $\R^d$ denotes the $d$-dimensional real Euclidean space and $\|\cdot\|_2$ denotes the Euclidean ($\ell_2$) norm. We write $\x{X} \sim \mathcal{N}(\x{\mu}, \sigma^2 \mathbb{I})$ for a Gaussian random vector with mean $\x{\mu}$ and isotropic covariance $\sigma^2 \mathbb{I}$, and $\x{X} \sim \mathcal{LN}(\x{\mu}, \sigma^2 \mathbb{I})$ for the corresponding log-normal random vector. A quick recap of the lognormal distribution and its properties is provided in \cref{asec:ln} of the appendix. We write $t \sim \mathcal{U}[0,1]$ for a random variable uniformly distributed over $[0,1]$. $\mathbb{E}$ denotes the expectation operator. A stochastic process is written $\{\x{X}_t\}_{t \geq 0}$. $\x{W}_t$ denotes a standard Wiener process (Brownian motion). Differentials are written as $\mathrm{d}\x{X}_t$ and $\mathrm{d}\x{W}_t$. We follow the It\={o} interpretation of SDEs throughout \citep{Oksendal1985}. Convergence in distribution of a sequence of random vectors $\x{X}_n$ to $\x{X}$ is written $\x{X}_n \xrightarrow[]{\text{d}} \x{X}$. In the discrete-time samplers, $\delta > 0$ denotes the step-size, $k$ indexes the time-step, $\x{X}_k$ denotes the iterate at step $k$, and $\x{Z}_k \sim \mathcal{N}(\x{0}, \mathbb{I})$ is an independent noise vector drawn at each step. The annealing coefficient at step $k$ is $\kappa$, decayed geometrically by factor $\chi \in (0,1]$ at each outer iteration; $L$ denotes the number of Langevin inner steps per noise level. The parameters $\x{\mu} \in \R^d$ and $\sigma > 0$ are the drift and diffusion coefficients, respectively, of the underlying GBM. The Bregman divergence of a strictly convex function $h : \R^d \to \R$ is denoted by $D_h(\x{X}, \x{X}_k)$. The notation $\nabla_{\x{X}} \ell(\x{X}) \Bigr|_{\x{X} = \x{X}_k}$ denotes the gradient of $\ell$ evaluated at $\x{X}_k$.

\section{Stochastic Differential Equations and Generative Modelling}
\label{sec:background}
Diffusion models~\citep{DDPM20, DDIM21} and score-based models~\citep{NCSNv3_21} rely on an SDE framework with a forward SDE that transforms data into noise and a corresponding reverse-time SDE that transforms noise into data. These models have been successful in generating realistic samples across different data modalities such as images~\citep{song2021scorebased}, video~\citep{ho2022video} and audio~\citep{Richter2025}. The key idea is to construct a stochastic process such that one starts with samples from the true, unknown density and progressively transforms them to samples from a noisy, easy-to-sample-from density such as the isotropic Gaussian. Generating new samples requires inverting the forward process. Because the dynamics are stochastic, inversion goes beyond simple time-reversal and requires matching the marginal densities of the forward and reverse processes at every time instant $t$. While standard formulations use additive Gaussian noise with an isotropic Gaussian terminal density, the framework admits any forward SDE for which the terminal density is tractable. In the present formulation, the marginal density is log-normal. \citet{Anderson1982} and \citet{Castanon1982} independently showed that, for a forward SDE of the form 
\begin{align}
    \dX{\x{X}_t} = h(\x{X}_t, t)\,\dX{t} + g(\x{X}_t, t)\,
    \dX{\x{W}_t}, \label{eqn:forward_sde}
\end{align}
there exists a corresponding reverse-time SDE, which has immediate relevance for generation. We adopt the multivariate version of the reverse-time SDE formula given by \citet{NCSNv3_21}
\begin{eqnarray}
        \dX{\x{X}_t} = -\left( h(\x{X}_t, t) - \nabla \cdot [g(\x{X}_t, t) g(\x{X}_t,t)^\top]  - g(\x{X}_t,t)g(\x{X}_t,t)^{\top}  \nabla \log p_{\x{X}_{t}}(\x{X}_t, t) \right) \dX{t} + g(\x{X}_t, t) \dX{\bar{\x{W}}_t},
        \label{eq:reverse_sde}
\end{eqnarray}
where $\bar{\x{W}}_t$ is the reverse-time Brownian motion and $\nabla\cdot F(\x{x}) := (\nabla \cdot f^1(\x{x}),\ \nabla\cdot f^2(\x{x}), \cdots, \ \nabla\cdot f^d(\x{x}))^{\top}$ is the row-wise divergence of the matrix-valued function $F(\x{x}) := (f^1(\x{x}), f^2(\x{x}), \cdots, f^d(\x{x}))^{\top}\in \R^{d\times d}$. To generate new samples from \cref{eq:reverse_sde}, we must have access to the time-dependent score function $\nabla \log p_{\x{X}}(\x{X}_t, t)$, which, in practice, is approximated by a neural network, $s_{\x{\theta}}:\R^d\times [0,1]\to \R^d$, trained to minimize the denoising score-matching loss~\citep{song2021scorebased}
\begin{equation}
    \mathcal{L}(\x{\theta}) = \underset{t \sim \mathcal{U}[0, 1]}{\E}\Bigg[\underset{\substack{\x{X}_0 \sim p_{\x{X}_0}\\\x{X}_t \sim p_{\x{X}_{t}|\x{X}_0}}}{\E}\left[\lambda(t)\left\|s_{\x{\theta}}(\x{X}_t, t) - \nabla \log p_{\x{X}_t |\x{X}_{0}}(\x{X}_t|\x{X}_0) \right\|^2_2\right]\Bigg],
    \label{dsm_loss}
\end{equation}
where $\nabla \log p_{\x{X}_t |\x{X}_{0}}(\x{X}_t|\x{X}_0)$ is determined by the forward SDE~\citep{sarkka2019applied} and $\lambda(t)$ is designed to stabilise training.

\section{Geometric Brownian Motion and its Time-reversal as a Generative Model}
\label{sec:GBM}
Geometric Brownian Motion (GBM) models scenarios where \textit{relative} increments of a stochastic process, rather than the absolute ones, follow a Brownian motion. This is in contrast to standard additive processes such as the Ornstein-Uhlenbeck SDE~\citep{Doob1942_OU}, $\mathrm{d}Y_t = \mu\,\dX{t} + \sigma\,\dX{W_t}$, whose solution $Y_t = Y_0 + \mu t + \sigma W_t$ is Gaussian with mean $\mu$ and variance $\sigma^2$. GBM was pioneered by~\citet{Fisher-Black1973_SDE} for modeling stock prices, where proportional rather than absolute changes are of interest. The marginal distribution is log-normal rather than Gaussian, making it a natural SDE of choice for positive-valued data. Formally, a random process $X_t$ is said to follow a GBM if it satisfies the SDE:
\begin{equation}
    \mathrm{d}X_t = \mu X_t \,\dX{t} + \sigma X_t \,\dX{W_t} \label{eq:gbm},
\end{equation}
where $W_t$ is the Wiener process, and  $\mu$ and $\sigma$ are known as the {\it percentage drift} representing a general trend and {\it volatility coefficients} representing the inherent stochasticity, respectively. The time evolution of $X_t$ follows a log-normal distribution with parameters $\mu$ and $\sigma^2$, i.e., $$X_t = X_0 \exp\left(\left(\mu -\frac{1}{2}\sigma^2\right)t + \sigma W_t\right).$$ 
There exist several multivariate extensions of GBM~\citep{Hu2000}. We consider the element-wise extension of~\cref{eq:gbm} for image data with the forward SDE for time $t \in [0, 1]$:
\begin{align}
    \d \x{X}_t = \x{\mu} \circ \x{X}_t \,\dX{t} + \sigma \x{X}_t \circ \dX{\x{W}_t},\label{eqn:gbmsde_forward}
\end{align}
where $\circ$ denotes element-wise multiplication, $\x{\mu} \in \R^d$ is the drift term, $\sigma > 0$, $\sigma\,\text{diag}(\Xt)$ is the diffusion matrix, and $\x{W}_t$ denotes the multivariate Wiener process. 

The distribution of $\x{X}_t$ given $\x{X}_0$, as it evolves according to \cref{eqn:gbmsde_forward}, has i.i.d. entries that are log-normally distributed. Starting from $\x{X}_0$ from the unknown density $p_{\x{X}_0}$, the solution to~\cref{eqn:gbmsde_forward} is 
\begin{equation}
    \label{eq:sde-solution}
    \x{X}_t = \x{X}_0 \circ \exp\left(\left(\x{\mu} - \frac{\sigma^2}{2}\x{1}\right)t + \sigma \Wt\right).
\end{equation} 
The closed-form expression allows us to generate samples from the forward process for any time instant $t \in [0, 1]$. The samples at the end of the forward process are log-normally distributed. We derive a corresponding reverse-time SDE that would enable us to generate samples from the unknown density $p_{\x{X}_0}$ starting from samples following the log-normal density.
For the forward SDE in \cref{eqn:gbmsde_forward}, we invoke \cref{eq:reverse_sde} to get the following reverse-time SDE
\begin{align}
    \dX{\Xt} = \left(\left(2\sigma^2\x{1} - \x{\mu}\right) \circ \Xt + \sigma^2 \Xt^{2} \circ \scox \right)\,\dX{t} + \sigma \Xt \circ \dX{\Wt},
    \label{eqn:gbmsde_backward_x}
\end{align}
where $\scox$ is the score function corresponding to $\x{X}_t$ and $\x{1}$ is a vector of all ones. 

\subsection{Lamperti Transform for GBM}
\label{sec:lamperti}
The forward SDE in \cref{eqn:gbmsde_forward} has a state-dependent diffusion term $\sigma\mathbf{X}_t$, which makes the derivation of the reverse-time SDE cumbersome. We show that the state-dependent diffusion is precisely what gives rise to the multiplicative structure of the sampler. The Lamperti transform~\citep{lamperti}, here, the element-wise $\log$ map, removes the state-dependence by converting the GBM SDE into an SDE with constant diffusion, to which the standard reverse-time formula in \cref{eq:reverse_sde} and the score change-of-variables formula~\citep{robbins2024scorechangevariables} can be applied. The forward SDE in \cref{eqn:gbmsde_forward} can be written equivalently, using the Lamperti transform $\x{Y}_t = \log \x{X}_t$ invoking It$\hat{\text{o}}$'s lemma, as follows:
\begin{align}
    \d \log \x{X}_t = \left(\x{\mu} - \frac{\sigma^2}{2}\x{1}\right)\,\dX{t} + \sigma\dX{\x{W}_t},\label{aeqn:gbmsde_forward_eq}
\end{align}
where $\log$ is applied element-wise. The distribution of $\x{X}_t$, as it evolves according to~\cref{aeqn:gbmsde_forward_eq}, has i.i.d. entries that are log-normally distributed with parameters $\x{\mu}$ and $\sigma^2\bb{I}$, $\bb{I}$ being the $d\times d$ identity matrix (identical to the solution in \cref{eq:sde-solution}). We can rewrite~\cref{aeqn:gbmsde_forward_eq} as
\begin{align}
    \d \x{Y}_t = \left(\x{\mu} - \frac{\sigma^2}{2}\x{1}\right)\,\dX{t} + \sigma\dX{\x{W}_t}.
    \label{eqn:gbmsde_forward_y}
\end{align}
The reverse-time SDE corresponding to the forward SDE in~\cref{eqn:gbmsde_forward_y} is given by (cf.~\cref{eq:reverse_sde})
\begin{align}
    \dX{\Yt} = - \left(\x{\mu} - \frac{\sigma^2}{2}\x{1} - \sigma^2 \scoy \right)\,\dX{t} + \sigma \dX{\Wt},
    \label{eqn:gbmsde_backward_y}
\end{align}
where $\nabla \log p_{Y}(\x{Y}_{t}, t)$ is the score function corresponding to $\x{Y}_t$ and $\x{1}$ is a vector of all ones. We leverage the score change-of-variables formula~\citep{robbins2024scorechangevariables} to represent $\scoy$ in terms of $\scox$ as $\scoy = \x{1} + \Xt \circ \scox$. Thus, we rewrite~\cref{eqn:gbmsde_backward_y} in terms of $\Xt$ as
\begin{align}
    \dX{\log \Xt} = -\left(\x{\mu} - \frac{3\sigma^2}{2}\x{1} - \sigma^2 \Xt \circ \scox \right)\,\dX{t} + \sigma \dX{\Wt}.
    \label{eqn:gbmsde_backward_y2}
\end{align}
\cref{eqn:gbmsde_backward_y2} is an SDE in the $\log$ space with a state-independent diffusion coefficient $\sigma$.

\subsection{Discretization of the Reverse-Time SDE}
\label{ssec:discretization}
The two reverse-time SDEs derived above, \cref{eqn:gbmsde_backward_x}, directly in the ambient space, and \cref{eqn:gbmsde_backward_y2} via the Lamperti transform in the log-space, yield two distinct samplers upon Euler-Maruyama discretization~\citep{Higham2001}. Both turn out to be multiplicative, yet they differ in a fundamental way: the sampler from \cref{eqn:gbmsde_backward_x} is \textit{sign-agnostic}, i.e., the sign of the entries in $\x{X}_t$ may not be preserved over updates, whereas the sampler from \cref{eqn:gbmsde_backward_y2} is \textit{sign-preserving}, i.e., the sign of the entries in $\x{X}_t$ are preserved over time. The sign-preserving update is reminiscent of Dale's law, which states that synaptic flips do not occur during learning. The interval $[0,1]$ is discretized into $N$ steps with step-size $\delta = \frac{1}{N}$. Let $\x{X}_{k\delta}$ be denoted as $\x{X}_k$ for $k = 1, 2, \dots, N-1$.

The discretization of \cref{eqn:gbmsde_backward_x} resulting in the \textbf{sign-agnostic multiplicative sampler} is given by
\begin{align}
\boxed{
    \Xprev = \Xk \circ \left((1 + 2\delta\sigma^2)\x{1} - \delta\x{\mu} + \delta\sigma^2 \Xk \circ \scoxk + \sqrt{\delta}\sigma \Zk\right),}
    \label{eqn:gbmsde_backward_sampling}
\end{align}
where $\x{Z}_k \sim \mathcal{N}(\x{0}, \mathbb{I})$ and $\scoxk$ is the score function evaluated at $\Xk$ and $t = k\delta$. The update is multiplicative, $\Xprev$ is obtained from $\Xk$ after scaling by a data-dependent factor, and because that factor may be negative, the sign of the entries of $\Xk$ may not be preserved across updates.

We refer to the discretization of \cref{eqn:gbmsde_backward_y2} in the log-space and its point-wise exponentiation as the \textbf{Dale-Langevin sampler}:
\begin{align}
\boxed{
    \Xprev = \Xk \circ \exp\left(-\delta\left(\x{\mu} - \frac{3\sigma^2}{2}\x{1}\right) + \delta\sigma^2 \Xk \circ \scoxk + \sqrt{\delta}\sigma \x{Z}_k\right).}
    \label{eqn:gbmsde_sampling_a}
\end{align}
Here, the scaling factor is an exponential, which is strictly positive, and therefore, the sign of every entry of $\Xk$ is preserved across updates. This update is also structurally equivalent to exponentiated gradient descent, as we shall make precise in \cref{asec:Dale}. In both cases, $\scox$ must be estimated from the data and \cref{sec:msm} describes the multiplicative score-matching objective used to train the score network.

\section{Multiplicative Score Matching}
\label{sec:msm}
We identify $\x{X}_t \circ \nabla \log p_{\x{X}_t}(\x{X}_t)$ as the natural score term in the multiplicative noise setting based on the structure of \cref{eqn:gbmsde_backward_sampling} and \cref{eqn:gbmsde_sampling_a}. We define the multiplicative explicit score-matching (M-ESM) and multiplicative denoising score-matching (M-DSM) counterparts of the explicit score matching (ESM) and denoising score matching (DSM) losses for the additive noise case~\citep{PascalVincent2011} to train a neural network $s_{\x{\theta}}$, $\x{\theta} \in \x{\Theta}$ (the parameter space), to approximate the score function as follows:
\begin{equation}
\boxed{
\mathcal{L}_{\text{M-ESM}}(\x{\theta}) = \underset{\x{X}_t \sim p_{\x{X}_t}}{\E} \left[ \frac{1}{2} \| \x{X}_t \circ \nabla \log p_{\x{X}_t}(\x{X}_t) - \x{X}_t \circ s_{\x{\theta}}(\x{X}_t,t) \|^2_2 \right], \quad}
\label{eq:mesm}
\end{equation}
\begin{equation}
\boxed{
\mathcal{L}_{\text{M-DSM}}(\x{\theta}) = \underset{\substack{\x{X}_0 \sim p_{\x{X}_0}\\ \x{X}_t \sim p_{\x{X}_t\mid \x{X}_0}}}{\E} \left[ \frac{1}{2} \| \x{X}_t \circ \nabla \log p_{\x{X}_t|\x{X}_0}(\x{X}_t|\x{X}_0) - \x{X}_t \circ s_{\x{\theta}}(\x{X}_t,t) \|^2_2 \right].}
\label{eq:mdsm}
\end{equation}
Generating samples from the marginal density $p_{\x{X}_t}$ is intractable, which makes $\mathcal{L}_{\text{M-ESM}}(\x{\theta})$ impossible to work with directly. We formalize the relationship between $\mathcal{L}_{\text{M-ESM}}(\x{\theta})$ and $\mathcal{L}_{\text{M-DSM}}(\x{\theta})$ in Theorem~\ref{thm:gbm_sm}. This allows us to train score models with $\mathcal{L}_{\text{M-DSM}}(\x{\theta})$. Before proceeding further with the key result, we state below the assumptions on the density and the data and model score functions over the positive orthant $\R_d^+$:
\begin{itemize}
    \item \textbf{Assumption 1} (Regularity of score functions). The p.d.f. $p_{\x{X}_t}$ is differentiable, the expectations $\E_{\x{X}_t \sim p_{\x{X}_t}}\big[\| \x{X}_t \circ \nabla \log p_{\x{X}_t}(\x{X}_t) \|_{2}^2 \big]$ and $\E_{\x{X}_t \sim p_{\x{X}_t}} \big[\| \x{X}_t \circ s_{\x{\theta}}(\x{X}_t, t) \|_{2}^2 \big]$ are finite for $\x{\theta} \in \x{\Theta}$ and $t \in [0, 1]$.
    \item \textbf{Assumption 2} (Boundary conditions). The quantity $p_{\x{X}_t}(\x{X}_t) (\x{X}_t \circ s_{\x{\theta}}(\x{X}_t, t))$ vanishes for $\x{\theta} \in \x{\Theta}$ and $t \in [0, 1]$ as $\| \x{X}_t \| \to \infty$.
\end{itemize}
\begin{theorem}[Multiplicative Denoising Score-Matching]\label{thm:gbm_sm}
Under the assumptions of regularity and appropriate boundary conditions stated above, the M-ESM loss given in \cref{eq:mesm} and the M-DSM loss given in \cref{eq:mdsm} are equivalent up to a constant, i.e., $\mathcal{L}_{\text{M-DSM}}(\x{\theta}) = \mathcal{L}_{\text{M-ESM}}(\x{\theta}) + C$, where $C$ is independent of $\x{\theta}$.
\end{theorem}
The proof is provided in Section~\ref{ssec:scorematching} of the appendix. Since $\mathcal{L}_{\text{M-ESM}}$ is intractable, the true marginal score $\nabla \log p_{\x{X}_t}(\x{X}_t)$ is unknown. The theorem provides a means to optimize $s_{\x{\theta}}$ via the conditional score $\nabla \log p_{\x{X}_t|\x{X}_0}(\x{X}_t|\x{X}_0)$, which is tractable from the forward SDE. Evaluating at discrete time $t = k\delta$ gives the training target:
\begin{equation}
   \x{X}_t \circ \nabla \log p_{\x{X}_t|\x{X}_0}(\x{X}_t|\x{X}_0)\Bigr|_{t=k \delta} = -\left(\x{1} + \dfrac{1}{\sigma^2 k \delta} \left(\log \x{X}_k - \log\x{X}_0 - k\delta\left(\x{\mu} - \frac{\sigma^2}{2}\x{1}\right) \right)\right).
    \label{eq:gbm_score}
\end{equation}
This is the multiplicative counterpart of the denoising score-matching target proposed by~\citet{song2021scorebased} for additive noise. The score-matching loss proposed by~\citet{Hyvaerinen2007} for non-negative data ($\mathcal{L}_{\text{NN}}(\x{\theta})$) emerges as a special case of $\mathcal{L}_{\text{M-ESM}}(\x{\theta})$ at $t=0$:
\begin{align}
    \mathcal{L}_{\text{NN}}(\x{\theta}) = \dfrac{1}{2}\underset{\x{X}_0 \sim p_{\x{X}_0}}{\E} \left[\|\x{X}_0\circ \nabla \log p_{\x{X}_0}(\x{X}_0) - \x{X}_0\circ s_{\x{\theta}}(\x{X}_0)\|_{2}^{2}\right].
    \label{eq:hyvaerinen_nn}
\end{align}
The formulation of~\citet{Hyvaerinen2007} is static as there is no temporal dependency. It was primarily motivated by a need to avoid the singularity at the origin for non-negative data in the context of parameter estimation for model densities. Our framework based on the reverse-time SDE discretization shows that the multiplicative score term pops up naturally. The log-normal structure of GBM implicitly restricts samples to the positive orthant, and the loss in~\citet{Hyvaerinen2007} is recovered by evaluating \cref{eq:mesm} at $t=0$. The M-DSM framework is thus a generalization of the result of ~\citet{Hyvaerinen2007} to the multiplicative setting.

\section{Dale's Law, Exponentiated Gradient Descent, and Mirrored Langevin Dynamics}
\label{asec:Dale}
The Dale-Langevin sampler, exponentiated gradient descent, and Mirrored Langevin Dynamics (MLD) are unified by a single convex generator: $h(\x{X}) = \sum_{i=1}^{d} X_i \log X_i - X_i$, where $X_i$ is the $i^{\text{th}}$ entry of the vector $\x{X}$. In optimization, $h$ defines the Bregman divergence that yields EGD and enforces Dale's law. In sampling, the same $h$ defines the mirror map that yields MLD on the positive orthant. The Dale-Langevin dynamics derived from the GBM reverse-time SDE sits precisely at the intersection between the two perspectives. The difference, however, is that Dale's law acts on the weights of the network, whereas the multiplicative samplers act on the state vector $\x{X}_{k}$.

\begin{figure}[t]
    \centering
    \includegraphics[width=0.87\textwidth]{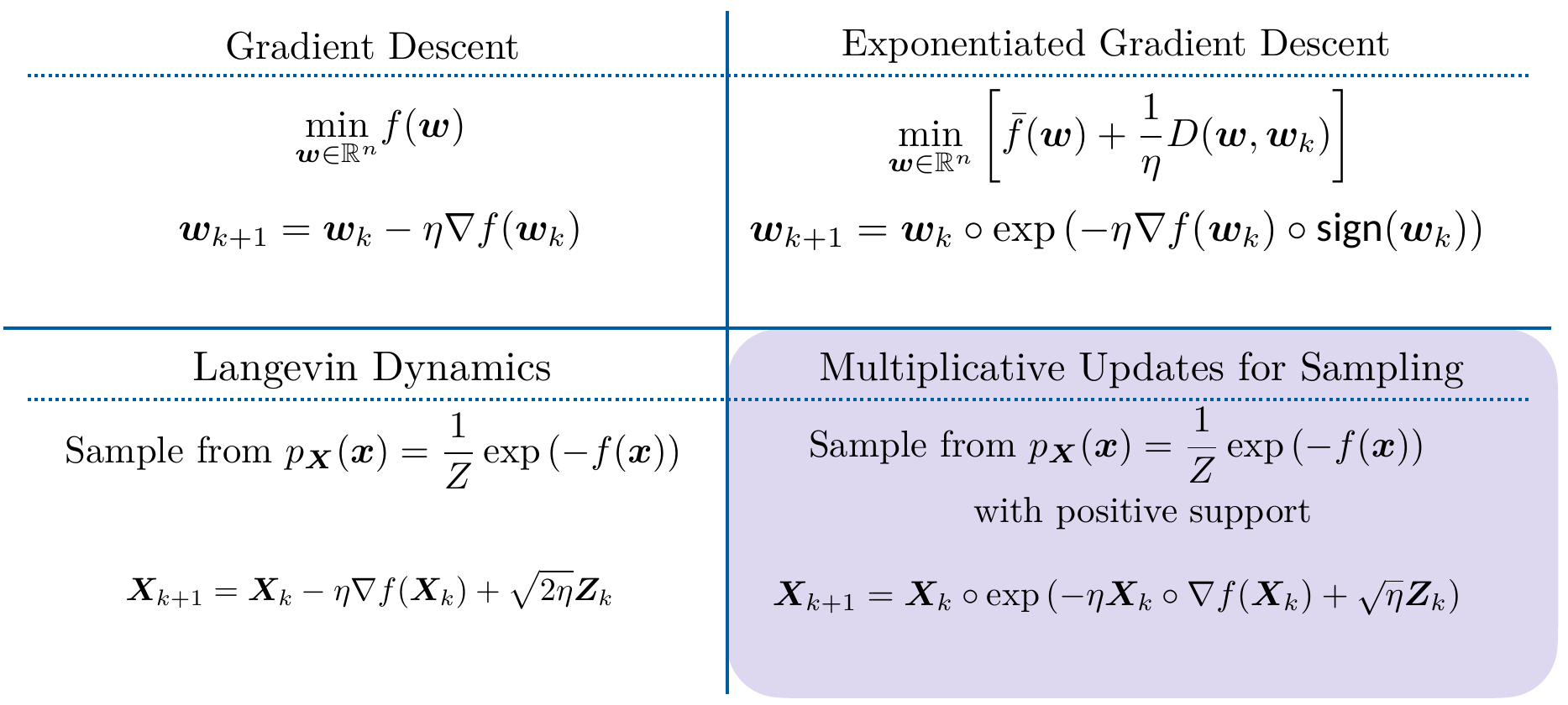}
    \caption{The convex generator $h$ that produces EGD in optimization and the Dale-Langevin sampler derived from the GBM reverse-time SDE share the same multiplicative $\exp(\cdot)$ update structure, establishing a formal duality between biologically plausible learning and score-based generative sampling.}
    \label{fig:mld_egd}
\end{figure}
\subsection{Exponentiated Gradient Descent and Dale's Law}
\label{ssec:egd}
\citet{cornford2024} demonstrated that standard gradient descent is a suboptimal phenomenological fit to learning experiments in biologically relevant settings: it violates Dale's law~\citep{Eccles1954} by allowing synaptic flips, which leads to weight distributions that are not log-normal, contradicting experimental observations. Additionally, they showed that exponentiated gradient descent (EGD)~\citep{Kivinen1997} respects Dale's law, produces log-normally distributed weights, and is superior to gradient descent for synaptic pruning. The learning task can be formulated via mirror descent~\citep{bubeck2015} as a minimization of a linear combination of task error and synaptic change penalty. For loss $\ell:\R^n\to\R_{+}$, we obtain the following update rule for synaptic weights $\Xk$, $k \in \bb{N}$:
\begin{eqnarray}
\label{aeq:bregman}
    \x{X}_{k+1} = \arg\min_{\x{X}} \left[ \ell(\x{X}) + \dfrac{1}{\eta} D_{h}(\x{X}, \x{X}_{k}) \right],
\end{eqnarray}
where $\ell(\x{X}) = \ell(\x{X}_{k}) + \nabla \ell(\x{X})^{\top}\Bigr|_{\x{X}=\x{X}_{k}}\left(\x{X} - \x{X}_k\right)$ is the first-order Taylor-series expansion of $\ell$ about $\x{X}_k$, and $D_{h}$ is the Bregman divergence of a strictly convex function $h:\R^d\to\R$. In particular, $h(\x{X}) = \|\x{X}\|_2^2$ yields standard gradient descent: $\x{X}_{k+1} = \x{X}_{k} - \eta \nabla \ell(\x{X})\Bigr|_{\x{X} = \x{X}_{k}}$, which does not preserve sign and therefore synaptic flips are possible~\citep{cornford2024}. Specifically, \citet{cornford2024} choose $h(\x{X}) = \sum\limits_{i=1}^{d}|X^{(i)}| \log |X^{(i)}| - |X^{(i)}|$, yielding $D_{h}$ as the unnormalised relative entropy: $D_{h}(\x{X}, \x{X}_{k}) = \sum\limits_{i=1}^{d} {X^{(i)}} \log \dfrac{X^{(i)}}{X^{(i)}_k} - X^{(i)} + X^{(i)}_k.$
For this choice, \cref{aeq:bregman} takes the form 
\begin{equation}
    \x{X}_{k+1} = \x{X}_{k} \circ \exp\left(-\eta \nabla \ell(\x{X})\Bigr|_{\x{X} = \x{X}_{k}} \circ \text{ sign}(\x{X}_{k}) \right)
    \label{eq:egd}
\end{equation}
where $\circ$ denotes element-wise multiplication. This update differs from standard gradient descent in three respects: it is multiplicative rather than additive, involves exponentiation, and preserves the sign of every entry of $\x{X}_k$ across the iterations. By design, EGD does not allow synaptic flips, automatically respects Dale's law, and produces log-normally distributed weights~\citep{pogodin2024synaptic}, which is precisely the terminal distribution of the GBM forward process. Comparing \cref{eq:egd} with the Dale-Langevin sampler in \cref{eqn:gbmsde_sampling_a}, the structural equivalence is immediate: the multiplicative updates have the same form and are sign-preserving across iterations, with the role of $\nabla \ell$ in EGD played by the multiplicative score $\x{X}_k \circ \scoxk$ in the sampler. This connection is illustrated in \cref{fig:mld_egd} and formalized next.

\subsection{Connection to Mirrored Langevin Dynamics}
\label{sec:mld}
Mirrored Langevin Dynamics (MLD)~\citep{mld} is the sampling analogue of mirror descent: it circumvents constrained sampling in the primal space by working in the unconstrained dual space. For a target density $p_{\X}(\x{x}) \propto \exp(-V(\x{x}))$, the canonical SDE is
\begin{equation}
    \dX{\Xt} = -\nabla V(\Xt)\,\dX{t} + \sqrt{2} \dX{\Wt}.
    \label{eq:langevin_sde}
\end{equation}
Under the mirror map $\Yt = \nabla h(\Xt)$ for a convex $h$ satisfying certain assumptions (cf.~\citep{mld}), the corresponding SDE in the dual space is given by
\begin{equation}
    \dX{\Yt} = -(\nabla^{2}h(\Xt))^{-1}\left(\nabla V(\Xt) + \nabla \log \det \nabla^{2}h(\Xt)\right)\,\dX{t} + \sqrt{2} \dX{\Wt},
    \label{eq:mld_sde}
\end{equation}
where $\nabla^{2}h(\Xt)$ denotes the Hessian of $h$ evaluated at $\Xt$. For the convex function $h(\x{X}) = \sum_{i=1}^{d} X_i \log X_i - X_i$, the same $h$ that yields EGD in Section~\ref{ssec:egd}, we have $\nabla h(\x{X}) = \log \x{X}$ and $\nabla^{2}h(\x{X}) = \mathrm{diag}(1/\x{X})$. Upon substitution into \cref{eq:mld_sde}, we get
\begin{equation}
    \dX{\log \Xt} = \left(\x{1} + \Xt \circ \scox \right)\,\dX{t} + \sqrt{2} \dX{\Wt}.
    \label{eq:mld_gbm}
\end{equation}
\cref{eq:mld_gbm} is a sampling SDE on the positive orthant, the multiplicative analogue of the standard Langevin dynamics, and is recovered directly from the GBM's reverse-time SDE (\cref{eqn:gbmsde_backward_y2}) via the Lamperti transform. In the appendix (Sec.~\ref{asec:mld_lognormal}), we specialize MLD to a log-normal target and establish convergence guarantees analogous to those in~\citet{Wibisono18a}.
\begin{figure}[ht]
    \centering
    \begin{subfigure}[b]{0.32\textwidth}
        \includegraphics[width=\linewidth]{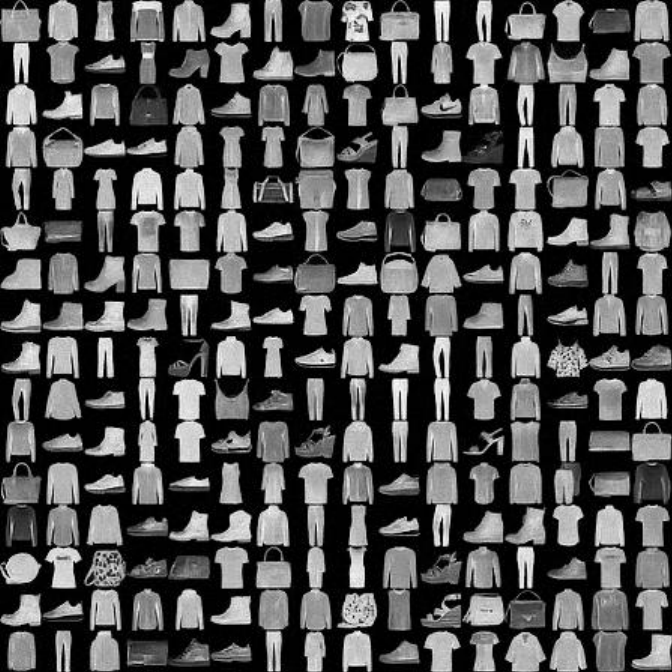}
        \caption{Fashion-MNIST}
        \label{fig:fmnncsn}
    \end{subfigure}\hfill
    \begin{subfigure}[b]{0.32\textwidth}
        \includegraphics[width=\linewidth]{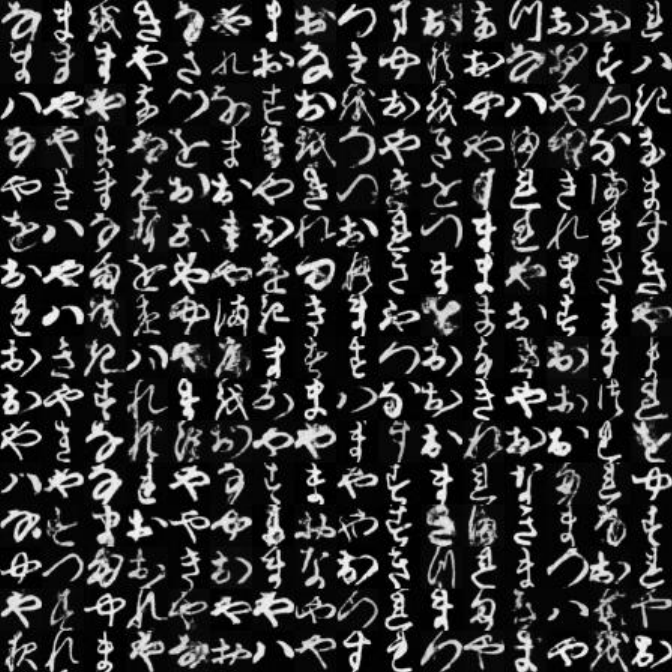}
        \caption{Kuzushiji MNIST}
        \label{fig:kmnncsn}
    \end{subfigure}\hfill
    \begin{subfigure}[b]{0.32\textwidth}
        \includegraphics[width=\linewidth]{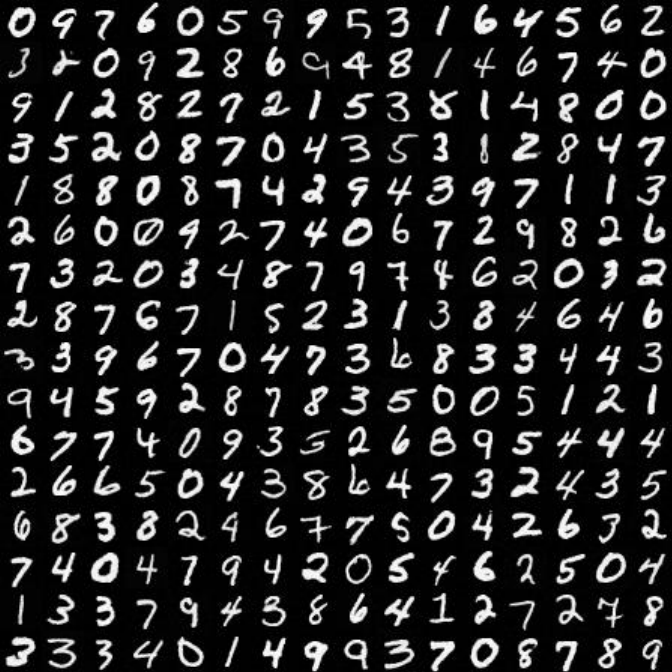}
        \caption{MNIST}
        \label{fig:Mnncsnd}
    \end{subfigure}
    \caption{Samples generated via the NCSN-Deeper architecture using Annealed-Dale-Langevin sampler \cref{algo:gbm_sampler_anneal} for (a) Fashion-MNIST, (b) Kuzushiji MNIST, and (c) MNIST. The sampling process was initialized by applying the forward process to class-averaged images, following the configurations ($L,\delta , \chi$) in \cref{tab:parameters_sampling}.}
    \label{fig:ncsndeeper_class_avg}
\end{figure}
\section{Experiments}
\label{sec:experiments}
We evaluate the generative performance of the proposed framework on three grayscale image datasets: MNIST (handwritten digits), Fashion-MNIST (clothing items), Kuzushiji-MNIST~\citep{cornford2021} (handwritten cursive Japanese characters); and one multi-channel dataset: CIFAR-10~\footnote{Code is available at \url{https://anonymous.4open.science/r/Dale-Meets-Langevin-65F8}}. For the grayscale datasets, each dataset contains $70{,}000$ images of size $28 \times 28$, split into $60{,}000$ training and $10{,}000$ test images. The CIFAR-10~\citep{Krizhevsky09learningmultiple} dataset contains $60{,}000$ images of size $32 \times 32 \times 3$, split into $50{,}000$ training and $10{,}000$ test images. The pixel values are rescaled to the range $[1, 2]$ to satisfy the positivity constraint. We use $N = 1000$ discretization steps for the forward SDE in \cref{eqn:gbmsde_forward}, corresponding to a step-size of $\delta = 1/N = 0.001$, with $\x{\mu} = \frac{\sigma^2}{2}\x{1}$ and $\sigma = 0.8$. We provide additional details in \cref{ssec:train-details}.

\subsection{Score Network Architecture}
The score network is based on the NCSNDeeper architecture~\citep{NCSNv1_19} with dilated convolutions, adapted for image generation with $N$ discrete time-steps. The network follows an encoder-decoder structure with skip connections and class conditioning is applied via conditional normalization layers. The encoder consists of an initial convolutional layer followed by five stages of residual blocks with progressive downsampling, where each block incorporates class information. The decoder mirrors this structure with five conditional refine blocks that progressively upsample features while incorporating skip connections from the corresponding encoder stages. For CIFAR-10, we use DDPM++ architecture from~\cite{NCSNv3_21}, which was the additional architectural inclusion from StyleGAN2~\citep{StyleGANv2_20} with a pre-existing DDPM architecture~\citep{DDPM20}.

\subsection{Noise Initialization Strategies}
\label{sec:init}
We propose two strategies for initializing the reverse process.

\textbf{Class-Averaged Initialization.} We compute the per-class mean image of the training set and pass it through the forward diffusion process. The resulting terminal state serves as the starting point for sampling.

\textbf{lognormal Noise Initialization.} We apply a logarithmic transformation to the terminal samples obtained from the forward process and fit a Gaussian to the resulting pixel-value histograms, estimating mean $\mu$ and standard deviation $\sigma$. Sampling noise is then drawn from this Gaussian distribution and exponentiated to synthesise the lognormal noise. For CIFAR-10, $\mu$ and $\sigma$ are estimated channel-wise; for the other three datasets, a single-channel estimate is used. The fitted parameters are reported in \cref{tab:mu_sigma_values} in the appendix.

\begin{figure}[t]
    \centering
    \begin{minipage}[t]{0.48\textwidth}
        \begin{algorithm}[H]
            \caption{Annealed Dale-Langevin multiplicative sampler (ADLS)}
            \label{algo:gbm_sampler_anneal}
            \begin{algorithmic}
                \small
                \REQUIRE{$\sigma, \delta, \x{\mu}, L, \kappa, \chi, n, \hat{\x{\mu}}, \hat{\sigma}$, trained score network $s_{\x{\theta}}, n$ , \texttt{initclass\_average}}
                \STATE{$\kappa \leftarrow 1$}
                \IF{$ \texttt{initclass\_average} == \texttt{TRUE}$}
                    \STATE $\mathbf{X}_{N-1} \leftarrow \texttt{get\_class\_average()}$
                    \STATE $\mathbf{X}_{N-1} \leftarrow \texttt{apply\_forward\_process($\mathbf{X}_{N-1}$,$\mu$,$\sigma$)}$
                \ELSE
                    \STATE $\mathbf{X}_{N-1} \sim \mathcal{LN}(\hat{\mathbf{\mu}}\1,\, \hat{\sigma}^2\mathbf{I})$
                \ENDIF
                \FOR{$k \gets N-1$ \textbf{to} $1$}
                    \FOR{$j \gets 1$ \textbf{to} $L$}
                        \STATE{$\x{Z}_{k,j} \sim \mathcal{N}(\x{0}, \bb{I})$}
                        \STATE{$\x{X}_{k-1} = \x{X}_{k} \circ \exp\Bigl(-\delta \bigl(\x{\mu} - \tfrac{3\sigma^2}{2}\x{1}\bigr)$}
                        \STATE{\qquad\qquad\qquad $+ \delta\sigma^2 \x{X}_{k} \circ s_{\x{\theta}}(\x{X}_k, k) + \kappa\sigma\sqrt{\delta}\,\x{Z}_{k,j}\Bigr)$}
                    \ENDFOR
                    \STATE{$\kappa \leftarrow \kappa \times \chi$}
                \ENDFOR
                \STATE{$\x{X}_{1} \gets \x{X}_{0}$}
                \FOR{$l \gets 1$ \textbf{to} $n$}
                    \STATE{$\x{X}_{l+1} = \x{X}_{l} \circ \exp\Bigl(-\delta \bigl(\x{\mu} - \tfrac{3\sigma^2}{2}\x{1}\bigr) + \delta\sigma^2 \x{X}_{l} \circ s_{\x{\theta}}(\x{X}_l, l)\Bigr)$}
                \ENDFOR
            \end{algorithmic}
        \end{algorithm}
    \end{minipage}
    \hfill
    \begin{minipage}[t]{0.48\textwidth}
        \begin{algorithm}[H]
            \caption{Annealed sign-agnostic multiplicative sampler (ASAMS)}
            \label{algo:gbm_sampler_nlamp_anneal}
            \begin{algorithmic}
                \small
                \REQUIRE{$\sigma, \delta, \x{\mu}, L, \kappa, \chi, n, \hat{\x{\mu}}, \hat{\sigma}$, trained score network $s_{\x{\theta}}, n$ , initclass average}
                \STATE{$\kappa \leftarrow 1$}
                
                \IF{$\texttt{initclass\_average} == \texttt{TRUE}$}
                    \STATE $\mathbf{X}_{N-1} \leftarrow \texttt{get\_class\_average()}$
                    \STATE $\mathbf{X}_{N-1} \leftarrow \texttt{apply\_forward\_process($\mathbf{X}_{N-1}$,$\mu$,$\sigma$)}$
                
                \ELSE
                    \STATE $\mathbf{X}_{N-1} \sim \mathcal{LN}(\hat{\mathbf{\mu}}\1,\, \hat{\sigma}^2\mathbf{I})$
                \ENDIF
                \FOR{$k \gets N-1$ \textbf{to} $0$}
                    \FOR{$j \gets 1$ \textbf{to} $L$}
                        \STATE{$\x{Z}_{k,j} \sim \mathcal{N}(\x{0}, \bb{I})$}
                        \STATE{$\x{X}_{k-1} = \x{X}_{k} \circ \Bigl((1 + 2\delta\sigma^2)\mathbf{1} - \delta\x{\mu}$}
                        \STATE{\qquad\qquad\qquad $- \delta\sigma^2 \x{X}_{k} \circ s_{\x{\theta}}(\x{X}_k, k) + \kappa\sigma\sqrt{\delta}\,\x{Z}_{k,j}\Bigr)$}
                    \ENDFOR
                    \STATE{$\kappa \leftarrow \kappa \times \chi$}
                \ENDFOR
                \STATE{$\x{X}_{1} \gets \x{X}_{0}$}
                \FOR{$l \gets 1$ \textbf{to} $n$}
                    \STATE{$\x{X}_{l+1} = \x{X}_{l} \circ \Bigl((1 + 2\delta\sigma^2)\mathbf{1} - \delta\x{\mu}-\delta\sigma^2 \x{X}_{l} \circ s_{\x{\theta}}(\x{X}_l, l)\Bigr)$}
                \ENDFOR
            \end{algorithmic}
        \end{algorithm}
    \end{minipage}
\end{figure}

\subsection{Sampling Algorithms}
\label{sec:sampling}
We derived samplers for the reverse-time SDEs via Euler-Maruyama discretization~\citep{Higham2001}: \cref{eqn:gbmsde_backward_sampling} and \cref{eqn:gbmsde_sampling_a}. We propose annealed variants in which the noise coefficient $\kappa$ is decayed geometrically by a factor $\chi \in (0, 1]$ across noise levels. Setting $\chi = 1.0$ recovers standard Euler-Maruyama, while $\chi < 1$ progressively reduces noise injection over the $L$ repeated sampling steps at each noise level, consistently improving sample quality. We also set $\x{\mu} = \frac{\sigma^2}{2}\x{1}$ to simplify the update rule. We present the two annealed samplers in \cref{algo:gbm_sampler_anneal} and \cref{algo:gbm_sampler_nlamp_anneal}. 

For single-channel grayscale datasets (MNIST, Fashion-MNIST, Kuzushiji-MNIST), no additional denoising was required. For CIFAR-10, the multi-channel dataset, we found that applying several ($n$) terminal denoising steps at the end of \cref{algo:gbm_sampler_anneal} and \cref{algo:gbm_sampler_nlamp_anneal}, using the parameter configurations in \cref{tab:mu_sigma_values} improved the quality of generated samples.

Both sampling algorithms (\cref{algo:gbm_sampler_anneal} and \cref{algo:gbm_sampler_nlamp_anneal}) can be used with the initialization strategies proposed in \cref{sec:init}. In particular, setting \texttt{initclass\_average} to \texttt{TRUE} initializes sampling with class-averaged images using \texttt{get\_class\_average()} function, which are then passed through the forward process. Otherwise, the initialization is done with lognormal noise with parameters $\hat{\mu},\hat{\sigma}$ with appropriate broadcasting in the multichannel case. 

The values of $\delta$, $L$, and $\chi$ are selected empirically via a grid-search, drawing $1024$ samples per configuration to minimize FID and KID. The resulting configurations are reported in \cref{tab:parameters_sampling,tab:sign_agnostic_class_avg} in \cref{asec:exp-details} of the appendix.

\begin{table}[t]
    \caption{FID and KID scores computed over $50$k generated and $10$k real test samples.
    ADLS = Annealed Dale-Langevin Sampler (\cref{algo:gbm_sampler_anneal});
    ASAMS = Annealed sign-agnostic Sampler (\cref{algo:gbm_sampler_nlamp_anneal}).
    w/o denotes sampling without the terminal denoising step; $(n)$ denotes $n$ terminal denoising steps.
    CIFAR-10 results use exponential moving-averaged checkpoints every $50$k iterations.}
    \centering
    \footnotesize
    \renewcommand{\arraystretch}{1.1}
    \begin{tabular}{l l l l | c c }
        \hline
        \textbf{Dataset} & \textbf{Sampler} & \textbf{Architecture} & \textbf{Initialization} & \textbf{FID} $(\downarrow)$ & \textbf{KID} $(\downarrow)$ \\
        \hline
        MNIST & ADLS  & NCSN-Deeper & class average & $11.4858$ & $0.0663 \pm 0.0021$ \\
        MNIST & ADLS  & NCSN-Deeper & lognormal  & $30.0030$ & $0.0385 \pm 0.0016$ \\ \cline{2-6}
        MNIST & ASAMS & NCSN-Deeper & class average & $11.3855$ & $0.0859 \pm 0.0022$ \\ 
        MNIST & ASAMS & NCSN-Deeper & lognormal  & $14.6595$ & $0.0489 \pm 0.0020$  \\ \cline{2-6}
        \hline\hline

        Kuzushiji-MNIST & ADLS  & NCSN-Deeper & class average & $16.3796$ & $0.0176 \pm 0.0010$ \\
        Kuzushiji-MNIST & ADLS  & NCSN-Deeper & lognormal  & $27.5636$ & $0.0106 \pm 0.0008$ \\ \cline{2-6}
        Kuzushiji-MNIST & ASAMS & NCSN-Deeper & class average & $26.8998$ & $0.0186 \pm 0.0009$ \\ 
        Kuzushiji-MNIST & ASAMS & NCSN-Deeper & lognormal  & $21.6120$ & $0.0689 \pm 0.0019$ \\\cline{2-6}
        \hline\hline

        Fashion-MNIST & ADLS  & NCSN-Deeper & class average & $21.2460$ & $0.0233 \pm 0.0007$ \\
        Fashion-MNIST & ADLS  & NCSN-Deeper & lognormal  & $90.7468$ & $0.0654 \pm 0.0020$ \\ \cline{2-6}
        Fashion-MNIST & ASAMS & NCSN-Deeper & class average & $21.4585$  & $0.0364 \pm 0.0010$ \\ 
        Fashion-MNIST & ASAMS & NCSN-Deeper & lognormal  & $103.2147$ & $0.1091 \pm 0.0021$ \\\cline{2-6}
        \hline\hline
        CIFAR-10 w/o   & ADLS  & DDPM++      & class average & $88.3223$  & $0.1575 \pm 0.0140$ \\
        CIFAR-10 $(25)$ & ADLS  & DDPM++      & class average & $38.5620$  & $0.0437 \pm 0.0071$\\ 
        CIFAR-10 w/o   & ADLS  & DDPM++      & lognormal  & $86.3112$  & $0.1585 \pm 0.0034$ \\
        CIFAR-10 $(25)$ & ASAMS & DDPM++      & lognormal  & $51.6470$  & $0.0645 \pm 0.0086$\\\cline{2-6}
        CIFAR-10 w/o   & ASAMS & DDPM++      & class average & $87.9819$  & $0.1629 \pm 0.0132$ \\
        CIFAR-10 $(16)$ & ASAMS & DDPM++      & class average & $45.0849$  & $0.0700 \pm 0.0094$ \\ 
        CIFAR-10 $(25)$ & ADLS  & DDPM++      & lognormal  & $46.1921$  & $0.0520 \pm 0.0022$ \\
        CIFAR-10 w/o   & ASAMS & DDPM++      & lognormal  & $100.7164$ & $0.1812 \pm 0.0148$ \\\cline{2-6}
        \hline
    \end{tabular}
    \label{tab:fid_kid_combined}
\end{table}

\subsection{Evaluation Metrics}
We evaluate generative quality using Fr\'echet Inception Distance (FID)~\citep{Frechet17} and Kernel Inception Distance (KID)~\citep{binkowski2018demystifying}, computed via the \texttt{torcheval} and \texttt{torchmetrics} libraries~\citep{detlefsen2022torchmetrics}. Scores are computed using 50,000 generated samples and 50,000 real samples from the test set. For grayscale image datasets, the images are replicated across the three channels and resized to 299 $\times$ 299 to match the input requirements of the InceptionV3 network~\citep{inceptionv3}.

We evaluate samples generated by both samplers (\cref{algo:gbm_sampler_anneal,algo:gbm_sampler_nlamp_anneal}) under both initialization strategies (lognormal noise and class-averaged), as initialization is often a critical factor affecting generated sample quality. The FID and KID scores for the annealed versions of the Dale-Langevin sampler (ADLS) and the sign-agnostic multiplicative sampler (ASAMS) are reported in \cref{tab:fid_kid_combined}. We observe that, in general, class-averaged initialization gives rise to comparable or better FID and KID compared to the lognormal initialization. Taken together, both metrics establish that multiplicative samplers offer a viable generative modeling paradigm.

\section{Conclusions and Outlook}
\label{sec:conclusions}
We introduced a novel diffusion framework grounded in Geometric Brownian Motion, an SDE for positive-valued data, which preserves strict positivity throughout the trajectory and admits the log-normal as its marginal distribution. A core contribution of this work is the theoretical unification of multiplicative diffusion, score matching, and optimization. We derive two multiplicative samplers --- a sign-agnostic one and a sign-preserving one, termed as the Dale-Langevin sampler. We showed that the Dale-Langevin sampler, obtained by discretizing the GBM's reverse-time SDE, is equivalent to standard additive Langevin dynamics on the log-transformed variable via the Lamperti transform, establishing GBM-based diffusion as the multiplicative-space interpretation of score-based diffusion on $\mathbb{R}^d_+$. To match the multiplicative forward process, we derived the multiplicative denoising score-matching (M-DSM) loss and proved its equivalence to the multiplicative explicit score-matching (M-ESM) loss, generalizing the score-matching framework of~\citet{PascalVincent2011} to the multiplicative setting. Non-negative score matching of ~\citep{Hyvaerinen2007} naturally emerges as a special case at $t = 0$. Finally, we bridged these sampling dynamics with optimization, showing that the convex function driving exponential gradient descent exactly dictates Dale-Langevin dynamics under the Mirrored Langevin Dynamics~\citep{mld} framework. Experiments on MNIST, Fashion-MNIST, Kuzushiji-MNIST, and CIFAR-10 demonstrate the generative capability of score-based generative models designed with multiplicative noise.

Beyond unconditional generation, the proposed multiplicative score-matching framework can be adapted for image denoising and restoration tasks where the forward noising process is inherently multiplicative, as opposed to the widely assumed additive noise. While this work focused on log-normal noise, exploring other distributions, such as the gamma distribution, and their associated SDEs remains a compelling future direction. This would broaden the applicability to domains where various types of multiplicative noise are inherent, such as optical coherence tomography~\citep{oct} and synthetic aperture RADAR~\citep{sar}, enabling more robust and versatile restoration capabilities. Future research directions include 
high-resolution image synthesis as well as to non-image domains, such as financial time-series, where proportional changes and multiplicative dynamics are intrinsically fundamental.

\section*{Limitations}
\label{sec:limitations}
Like most score-based generative modeling paradigms, the proposed model also requires a large amount of training data and computational resources to achieve good performance, which can be a constraint in some applications. Further, the choice of hyperparameters, such as the noise schedule and learning rate, which are carefully tuned, can affect the performance of the model. However, this limitation is true of all deep generative models and not unique to ours.\par

\subsubsection*{Broader Impact Statement}
This work advances generative modeling by introducing a Geometric Brownian Motion (GBM) based diffusion framework with multiplicative dynamics substantiated by experiments conducted on standard image datasets (MNIST, Fashion-MNIST, Kuzushiji-MNIST and CIFAR-10). These experiments did not involve human subjects or sensitive personal data. Potential positive impact includes enabling score-based generative modeling and restoration/denoising in settings where multiplicative noise and non-negative data are intrinsically fundamental (e.g., optical coherence tomography and synthetic aperture radar). The technique could also be applied to financial time-series data. Risks are broadly consistent with those of generative models: the approach could be misused to generate biased, fake, or misleading content, and over-reliance on generated outputs in downstream decision-making could cause harm.

\bibliography{arxiv-release}
\bibliographystyle{abbrvnat}

\appendix

\section{Log-normal Distribution}
\label{asec:ln}
\indent A positive random variable $W$ is said to follow the log-normal distribution if $\log W \sim \N(\mu, \sigma^2)$, that is, $\log W$ follows a Gaussian distribution with mean $\mu$ and variance $\sigma^2$. We denote this as $W \sim \mathcal{LN}(\mu,\sigma^2)$. The log-normal density is given by 
\begin{equation}
f_W(w) = 
\begin{cases}
\dfrac{1}{w \sigma \sqrt{2\pi}} \exp \left( -\dfrac{(\log w - \mu)^2}{2\sigma^2} \right), & w > 0,\\
0, & w\leq 0.
\end{cases}
\label{logndef}
\end{equation}
Note that $\mu$ and $\sigma^2$ are {\bf not} the mean and variance of the log-normal random variable. The mean and variance of the log-normal random variable $W$ are $\E[W] = \exp\left(\mu + \frac{\sigma^2}{2}\right)$ and $\text{Var}(W) = \exp\left(\sigma^2 -1\right) \exp\left(2\mu + \sigma^2\right)$, respectively.

The multivariate log-normal random vector is defined as $\x{W} = \exp\left(\x{\mu} + \sigma \x{Z}\right)$ where $\x{Z} \sim \N(\x{0}, \bb{I})$ and the exponentiation is applied element-wise. Effectively, the entries of $\x{W}$ are independent and identically distributed according to \cref{logndef}. The corresponding density is denoted as $\mathcal{L}\N(\x{\mu}, \sigma^2\bb{I})$.

\section{Equivalence Between Multiplicative Denoising Score-Matching and Multiplicative Explicit Score-Matching}
\label{ssec:scorematching}

Recall from Sec.~\ref{sec:msm} of the main document that the multiplicative explicit score-matching loss is given by
\begin{equation}
    \mathcal{L}_{\text{M-ESM}}(\x{\theta}) = \underset{\x{X}_t \sim p_{\x{X}_t}}{\E} \left[ \frac{1}{2} \Big\| \x{X}_t \circ \nabla \log p_{\x{X}_t}(\x{X}_t) - \x{X}_t \circ s_{\x{\theta}}(\x{X}_t,t) \Big\|^2_2 \right],\label{eq:mesm_new}
\end{equation}
and that the multiplicative denoising score-matching loss is given by
\begin{equation}
    \mathcal{L}_{\text{M-DSM}}(\x{\theta}) = \underset{\substack{\x{X}_0 \sim p_{\x{X}_0}\\ \x{X}_t \sim p_{\x{X}_t\mid \x{X}_0}}}{\E} \left[ \frac{1}{2} \Big\| \x{X}_t \circ \nabla \log p_{\x{X}_t|\x{X}_0}(\x{X}_t|\x{X}_0) - \x{X}_t \circ s_{\x{\theta}}(\x{X}_t,t) \Big\|^2_2 \right].\label{eq:mdsm_new}
\end{equation}
In the following result, we establish the equivalence between multiplicative explicit score-matching and multiplicative denoising score-matching loss. We state below the assumptions on the density and the data and model score functions over the positive orthant $\R_d^+$:
\begin{itemize}
    \item \textbf{Assumption 1} (Regularity of score functions). The p.d.f. $p_{\x{X}_t}$ is differentiable, the expectations $\E_{\x{X}_t \sim p_{\x{X}_t}}\big[\| \x{X}_t \circ \nabla \log p_{\x{X}_t}(\x{X}_t) \|_{2}^2 \big]$ and $\E_{\x{X}_t \sim p_{\x{X}_t}} \big[\| \x{X}_t \circ s_{\x{\theta}}(\x{X}_t, t) \|_{2}^2 \big]$ are finite for $\x{\theta} \in \x{\Theta}$ and $t \in [0, 1]$.
    \item \textbf{Assumption 2} (Boundary conditions). The quantity $p_{\x{X}_t}(\x{X}_t) (\x{X}_t \circ s_{\x{\theta}}(\x{X}_t, t))$ vanishes for $\x{\theta} \in \x{\Theta}$ and $t \in [0, 1]$ as $\| \x{X}_t \| \to \infty$.
\end{itemize}
\begin{theorem}[Multiplicative Denoising Score-Matching]\label{athm:gbm_sm}
Under the assumptions of regularity and appropriate boundary conditions stated above, the M-ESM loss given in \cref{eq:mesm} and the M-DSM loss given in \cref{eq:mdsm} are equivalent up to a constant, i.e., $\mathcal{L}_{\text{M-DSM}}(\x{\theta}) = \mathcal{L}_{\text{M-ESM}}(\x{\theta}) + C$, where $C$ is independent of $\x{\theta}$.
\end{theorem}
\begin{proof} We assume that the densities $p_{\x{X}_t}$ and $p_{\x{X}_t|\x{X}_0}$ (defined in Sec.~\ref{sec:GBM} of the main document) are supported over $\R_{+}^{d}$, and zero elsewhere. Further, we assume that $p_{\x{X}_t}(\x{x}_t) > 0, p_{\x{X}_t|\x{X}_0}(\x{x}_t\mid \x{x}_0) > 0, \ \forall \ \x{x}_t \in \R_{+}^{d}$ for $t \in [0, 1]$. The expectations are evaluated over the support $\R_{+}^{d}$. We expand $\mathcal{L}_{\text{M-ESM}}(\x{\theta})$ to get
\begin{eqnarray}
    \mathcal{L}_{\text{M-ESM}}(\x{\theta}) =  \underset{\x{X}_t \sim p_{\x{X}_t}}{\E} \left[\frac{1}{2}\Big\| \x{X}_t \circ \nabla \log p_{\x{X}_t}(\x{X}_t) \Big\|^2\right] + \underset{\x{X}_t \sim p_{\x{X}_t}}{\E} \left[\frac{1}{2}\Big\| \x{X}_t \circ s_{\x{\theta}}(\x{X}_t,t) \Big\|^2\right] \nonumber \\
    - \underset{\x{X}_t \sim p_{\x{X}_t}}{\E} \left[ (\x{X}_t \circ \nabla \log p_{\x{X}_t}(\x{X}_t))^{\top} (\x{X}_t \circ s_{\x{\theta}}(\x{X}_t,t)) \right].\label{eq:intermediate}
\end{eqnarray}
Consider the cross-term $\underset{\x{X}_t \sim p_{\x{X}_t}}{\E} \left[ (\x{X}_t \circ \nabla \log p_{\x{X}_t}(\x{X}_t))^{\top} (\x{X}_t \circ s_{\x{\theta}}(\x{X}_t,t)) \right]$ and express it as an integral over $\R_{+}^{d}$. For brevity of notation, we don't explicitly indicate the support $\R_{+}^{d}$ in the following integrals. The cross-term is given by\\

$\underset{\x{X}_t \sim p_{\x{X}_t}}{\E} \left[ (\x{X}_t \circ \nabla \log p_{\x{X}_t}(\x{X}_t))^{\top} (\x{X}_t \circ s_{\x{\theta}}(\x{X}_t,t)) \right]$
\begin{eqnarray}
&&= \int (\x{x}_t \circ \nabla \log p_{\x{X}_t}(\x{x}_t))^{\top} (\x{x}_t \circ s_{\x{\theta}}(\x{x}_t,t)) p_{\x{X}_t}(\x{x}_t) \,\mathrm{d}\x{x}_t \nonumber\\ 
&&= \int (\x{x}_t \circ \nabla p_{\x{X}_t}(\x{x}_t))^{\top} (\x{x}_t \circ s_{\x{\theta}}(\x{x}_t,t)) \,\mathrm{d}\x{x}_t.\label{eq:esm1}
\end{eqnarray}

We know that the marginal density $p_{\x{X}_t}(\x{x}_t)$ can be expressed in terms of the conditional density as 
\begin{eqnarray*}
    p_{\x{X}_t}(\x{x}_t) = \int p_{\x{X}_t|\x{X}_0}(\x{x}_t|\x{x}_0) p_{\x{X}_0}(\x{x}_0) \,\mathrm{d}\x{x}_0.
\end{eqnarray*}
Computing the gradient with respect to $\x{x}_t$ on both sides yields
\begin{equation}
\nabla p_{\x{X}_t}(\x{x}_t) = \int \nabla p_{\x{X}_t|\x{X}_0}(\x{x}_t|\x{x}_0) p_{\x{X}_0}(\x{x}_0) \,\mathrm{d}\x{x}_0. \label{eq:integral}    
\end{equation}
Substituting \cref{eq:integral} in \cref{eq:esm1}, multiplying and dividing by $p_{\x{X}_t|\x{X}_0}(\x{x}_t|\x{x}_0)$, we get\\
$\underset{\x{X}_t \sim p_{\x{X}_t}}{\E} \left[ (\x{X}_t \circ \nabla \log p_{\x{X}_t}(\x{X}_t))^{\top} (\x{X}_t \circ s_{\x{\theta}}(\x{X}_t,t)) \right]$
\begin{eqnarray}
    &&=\int \left( \x{x}_t \circ \int \nabla p_{\x{X}_t|\x{X}_0}(\x{x}_t|\x{x}_0)  p_{\x{X}_0}(\x{x}_0) \,\mathrm{d}\x{x}_0  \right)^{\top} (\x{x}_t \circ s_{\x{\theta}}(\x{x}_t,t)) \,\mathrm{d}\x{x}_t \nonumber\\
    &&= \iint (\x{x}_t \circ \nabla \log p_{\x{X}_t|\x{X}_0}(\x{x}_t|\x{x}_0))^{\top} (\x{x}_t \circ s_{\x{\theta}}(\x{x}_t,t)) \,p_{\x{X}_t|\x{X}_0}(\x{x}_t|\x{x}_0)  p_{\x{X}_0}(\x{x}_0) d\x{x}_0  \,\mathrm{d}\x{x}_t,\nonumber\\
    && = \underset{\substack{\x{X}_0 \sim p_{\x{X}_0}\\ \x{X}_t \sim p_{\x{X}_t\mid \x{X}_0}}}{\E} \left[ (\x{X}_t \circ \nabla \log p_{\x{X}_t|\x{X}_0}(\x{X}_t|\x{X}_0))^{\top} (\x{X}_t \circ s_{\x{\theta}}(\x{X}_t,t)) \right]. \label{eqn:intermed}
\end{eqnarray}
Substituting \cref{eqn:intermed} in \cref{eq:intermediate} gives the following equivalent expression for the multiplicative explicit score-matching loss:
\begin{eqnarray}
    \mathcal{L}_{\text{M-ESM}}(\x{\theta}) &&= \cancelto{C_1}{\underset{\x{X}_t \sim p_{\x{X}_t}}{\E}\left[\frac{1}{2} \Big\| \x{X}_t \circ \nabla \log p_{\x{X}_t}(\x{X}_t) \Big\|^2\right]} + \underset{\x{X}_t \sim p_{\x{X}_t}}{\E} \left[\frac{1}{2} \Big\| \x{X}_t \circ s_{\x{\theta}}(\x{X}_t,t) \Big\|^2\right] \nonumber\\
    &&\quad- \underset{\substack{\x{X}_0 \sim p_{\x{X}_0}\\ \x{X}_t \sim p_{\x{X}_t\mid \x{X}_0}}}{\E} \left[ (\x{X}_t \circ \nabla \log p_{\x{X}_t|\x{X}_0}(\x{X}_t|\x{X}_0))^{\top} (\x{X}_t \circ s_{\x{\theta}}(\x{X}_t,t)) \right]\nonumber\\
    &&=\underset{\x{X}_t \sim p_{\x{X}_t}}{\E} \left[\frac{1}{2} \Big\| \x{X}_t \circ s_{\x{\theta}}(\x{X}_t,t) \Big\|^2\right] \nonumber\\
    &&\quad- \underset{\substack{\x{X}_0 \sim p_{\x{X}_0}\\ \x{X}_t \sim p_{\x{X}_t\mid \x{X}_0}}}{\E} \left[ (\x{X}_t \circ \nabla \log p_{\x{X}_t|\x{X}_0}(\x{X}_t|\x{X}_0))^{\top} (\x{X}_t \circ s_{\x{\theta}}(\x{X}_t,t)) \right] + C_1,
    \label{eq:esmloss_newer}
\end{eqnarray}
where $C_1$ is a constant that is not dependent on $\x{\theta}$.\\
We carry out a similar simplification for the multiplicative denoising score-matching loss:
\begin{eqnarray*}
    \mathcal{L}_{\text{M-DSM}}(\x{\theta})&&=\underset{\substack{\x{X}_0 \sim p_{\x{X}_0}\\ \x{X}_t \sim p_{\x{X}_t\mid \x{X}_0}}}{\E} \left[ \frac{1}{2} \Big\| \x{X}_t \circ \nabla \log p_{\x{X}_t|\x{X}_0}(\x{X}_t|\x{X}_0) - \x{X}_t \circ s_{\x{\theta}}(\x{X}_t,t) \Big\|^2_2 \right], \nonumber\\
    &&= \cancelto{C_2}{\underset{\substack{\x{X}_0 \sim p_{\x{X}_0}\\ \x{X}_t \sim p_{\x{X}_t\mid \x{X}_0}}}{\E} \left[ \frac{1}{2} \Big\| \x{X}_t \circ \nabla \log p_{\x{X}_t|\x{X}_0}(\x{X}_t|\x{X}_0)\Big\|_{2}^{2}\right]} + \underset{\substack{\x{X}_0 \sim p_{\x{X}_0}\\ \x{X}_t \sim p_{\x{X}_t\mid \x{X}_0}}}{\E}\left[\dfrac{1}{2}\Big\|\x{X}_t \circ s_{\x{\theta}}(\x{X}_t,t) \Big\|^2_2 \right],\nonumber\\
    &&\quad-\underset{\substack{\x{X}_0 \sim p_{\x{X}_0}\\ \x{X}_t \sim p_{\x{X}_t\mid \x{X}_0}}}{\E} \left[ (\x{X}_t \circ \nabla \log p_{\x{X}_t|\x{X}_0}(\x{X}_t|\x{X}_0))^{\top} (s_{\x{\theta}}(\x{X}_t,t) \circ \x{X}_t) \right],
\end{eqnarray*}
or equivalently,
\begin{eqnarray}
    \mathcal{L}_{\text{M-DSM}}(\x{\theta}) &=& \underset{\substack{\x{X}_t \sim p_{\x{X}_t}}}{\E}\left[\dfrac{1}{2}\Big\|\x{X}_t \circ s_{\x{\theta}}(\x{X}_t,t) \Big\|^2_2 \right]  \nonumber\\
    &&- \underset{\substack{\x{X}_0 \sim p_{\x{X}_0}\\ \x{X}_t \sim p_{\x{X}_t\mid \x{X}_0}}}{\E} \left[ (\x{X}_t \circ \nabla \log p_{\x{X}_t|\x{X}_0}(\x{X}_t|\x{X}_0))^{\top} (s_{\x{\theta}}(\x{X}_t,t) \circ \x{X}_t) \right] + C_2,
    \label{eq:dsmloss}
\end{eqnarray}
where $C_2$ is a constant that is not dependent on $\x{\theta}$.\\
On comparing \cref{eq:esmloss_newer} and \cref{eq:dsmloss}, we get
\begin{equation}
    \mathcal{L}_{\text{M-DSM}}(\x{\theta}) = \mathcal{L}_{\text{M-ESM}}(\x{\theta}) + C_2 - C_1.
\end{equation}
This concludes the proof.
\end{proof}
The implication of the result is as follows: multiplicative explicit score-matching loss is intractable since we do not have access to the true marginal scores, and, this equivalence allows us to optimize the score network parameters by minimizing the multiplicative denoising score-matching loss instead since the conditional scores can be tractably computed from the forward SDE (cf. Sec.~\ref{sec:GBM}).

\section{Time-Varying Drift and Diffusion SDE}
Here, we consider a generalization of the proposed GBM framework wherein the drift and diffusion terms are both time dependent. In particular, consider the following SDE 
\begin{align}
    \d \log \x{X}_t = \left(\x{\mu}_t - \frac{\sigma_t^2}{2}\x{1}\right)\,\dX{t} + \sigma_t\dX{\x{W}_t},\label{aeqn:gbmsde_forward_eq_tv}
\end{align}
where $\log$ is applied element-wise, $\x{\mu}_t$ and $\sigma_t$ are both functions of time. Euler-Maruyama discretization of Eq.~\ref{aeqn:gbmsde_forward_eq_tv} yields
\begin{align}
    \log\Xnext = \log \Xk + \delta \left(\x{\mu}_{k} - \frac{\sigma_{k}^2}{2}\x{1}\right) + \sqrt{\delta}\sigma_k \Zk,
    \label{eqn:gbmsde_forward_tv}
\end{align}
where $\x{Z}_k \sim \N(\x{0}, \mathbb{I})$, $\x{\mu}_k \triangleq \x{\mu}_{k\delta}$ and $\sigma_k \triangleq \sigma_{k\delta}$ for $k=0, \dots, N-1$. Unrolling Eq.~\ref{eqn:gbmsde_forward_tv}, we get the following expression for $\Xk$ in terms of $\x{X}_0$
\begin{align}
    \Xk = \X_0 \circ \exp\left(\delta A_k + \sqrt{\delta}B_k \Z_0\right),
    \label{eq:gbm_tv_forward_discrete}
\end{align}
where $A_k = \sum\limits_{j=0}^{k}\left(\x{\mu}_{k} - \frac{\sigma_{k}^2}{2}\x{1}\right)$, $B_k = \sqrt{\sum\limits_{j=0}^{k}\sigma_k^2}$ and $\x{Z}_0 \sim \N(\x{0}, \mathbb{I})$. This allows us to generate samples for the forward process that is used during training. We set $\sigma_t = \sigma_{\text{max}}\left(\dfrac{\sigma_{\text{max}}}{\sigma_{\text{min}}}\right)^{t}$ which results in a variance exploding SDE (c.f.~\cite{song2021scorebased}) and the discretized version is $\sigma_k = \sigma_{\text{max}}\left(\dfrac{\sigma_{\text{max}}}{\sigma_{\text{min}}}\right)^{\frac{j-1}{N-1}}$ for $j=1, \dots, N-1$. From \cref{eq:gbm_tv_forward_discrete}, we see that $\Xk | \X_0 \sim \mathcal{L}\N(\log \X_0 + \delta A_k, \delta B_k^2 \id)$ and this provides the target term to be used in the loss (Eq.~\ref{eq:mdsm}) as 
\begin{align*}
    \x{x}_k \circ \nabla \log p_{\Xk | \X_0}(\x{x}_k | \x{x}_0) = -\left(\1 + \dfrac{1}{\delta B^{2}_{k}}\left(\log \x{x}_k - \x{x}_0\right)\right).
\end{align*}
The corresponding reverse-time SDE for Eq.~\ref{aeqn:gbmsde_forward_eq_tv} is 
\begin{align}
    \dX{\log \Xt} = -\left(\x{\mu}_t - \frac{3\sigma_t^2}{2}\x{1} - \sigma_t^2 \Xt \circ \scox \right)\,\dX{t} + \sigma_t \dX{\Wt}.
\end{align}
In principle, one could discretize this reverse-time SDE, train a neural network to approximate the score for samples generated using \cref{eq:gbm_tv_forward_discrete} and generate new samples from the corresponding discretized reverse-time SDE.

\section{Mirrored Langevin Dynamics for a Log-normal Target}
\label{asec:mld_lognormal}
\citet{Wibisono18a} show that the unadjusted Langevin algorithm for the Ornstein-Uhlenbeck process leads to a bias, i.e., it does not converge to the target distribution. In particular, they consider a Gaussian target with mean $\x{\mu}$ and covariance matrix $\Sigma$ and the limit measure is a Gaussian with the correct mean but covariance matrix $\hat{\Sigma} = \Sigma (\id - \frac{\epsilon}{2}\Sigma^{-1})^{-1}$ for the step-size $\epsilon$. For a log-normal target, we show that the limiting density obtained from GBM does not converge to the correct target density and is biased. 

In particular, we consider the example with the target $\mathcal{LN}(\x{\mu}, \sigma^2 \bb{I})$, i.e., $\nu(\x{x}) = (2\pi \sigma^2)^{-d/2} \prod\limits_{j=1}^{d}x_{i}^{-1}\exp\left(-\dfrac{1}{2\sigma^2}\left\|\log \x{x} - \x{\mu}\right\|^{2}_{2}\right)$. It can be shown that $\x{1} + \Xt \circ \scox = -\frac{1}{\sigma^2}\left(\log \x{X}_t - \x{\mu}\right)$. Substituting this in \cref{eq:mld_gbm}, we get
\begin{equation}
    \dX{\log \Xt} = -\frac{1}{\sigma^2}\left(\log \x{X}_t - \x{\mu}\right)\,\dX{t} + \sqrt{2} \dX{\Wt}.
\end{equation}
Using the Euler-Maruyama discretization with $\delta$ such that $0<\delta < \sigma^2$, we obtain
\begin{equation*}
    \log \Xk = \log \Xprev -\frac{\delta}{\sigma^2}\left(\log \Xprev - \x{\mu}\right) + \sqrt{2\delta} \Zk,
\end{equation*}
where $\Zk \sim \normal$. Now, subtracting $\x{\mu}$ from both sides and rearranging terms, we get
\begin{equation*}
    \log \Xk - \x{\mu} = \left(1 - \frac{\delta}{\sigma^2}\right)\left(\log \Xprev - \x{\mu}\right) + \sqrt{2\delta} \Zk.
\end{equation*}
Unrolling this equation till $\x{X}_0$, we get 
\begin{equation*}
    \log \Xk - \x{\mu} = \left(1 - \frac{\delta}{\sigma^2}\right)^{k}\left(\log \x{X}_0 - \x{\mu}\right) + \sqrt{\dfrac{2 \sigma^2}{2 - \frac{\delta}{\sigma^2}} \left(1 - \left(1 - \dfrac{\delta}{\sigma^2}\right)^{2k}\right)} \x{Z}_0.
\end{equation*}
This implies that $\log \Xk \xrightarrow[]{\text{d}} \x{\mu} + \sigma \sqrt{\dfrac{2}{2 - \frac{\delta}{\sigma^2}}}\x{Z}_0$, i.e., $\Xk \xrightarrow[]{\text{d}} \exp\left(\x{\mu} + \sigma \sqrt{\dfrac{2}{2 - \frac{\delta}{\sigma^2}}}\x{Z}_0\right)$. The corresponding limit measure is $\nu_{\delta}(\x{x}) = \mathcal{LN}\left(\x{\mu}, \dfrac{\sigma^2}{\left(1 - \frac{\delta}{2\sigma^2}\right)} \bb{I}\right)$. In this case, the integrating the SDE (\cref{eq:mld_gbm}) generates samples from the limit measure. The limit measure has the right parameter $\x{\mu}$ but the covariance matrix of the underlying Gaussian is biased by a factor of $\dfrac{1}{\left(1 - \frac{\delta}{2\sigma^2}\right)}$.

\section{Additional Experimental Details}
\label{asec:exp-details}

\subsection{Training Details}
\label{ssec:train-details}
The models are implemented in PyTorch and optimized using AdamW~\citep{loshchilovdecoupled}. The MNIST model is trained for $300$k iterations; Fashion-MNIST and Kuzushiji-MNIST models are trained for $200$k iterations. The CIFAR-10 model is trained for $1.3$ million iterations, with checkpoints saved every $5$k iterations. For CIFAR-10, we performed exponential moving-average (EMA) for each $50$k, $100$k, $150$k iterations and found that  models with EMA for every $50$k steps gave better results, similar to the setting in \citet{NCSNv2_20}. Training is distributed across two NVIDIA RTX 4090 and two NVIDIA A6000 GPUs.

The loss function for each model is the Monte Carlo estimator of the M-DSM loss (\cref{eq:mdsm}):
\begin{equation}
    \hat{\mathcal{L}}_{\text{M-DSM}}(\x{\theta}) = \frac{1}{NM}\sum_{i=1}^{M}\sum_{k=0}^{N-1} \left[ \frac{1}{2} \Bigl\| \x{x}_{k}^{(i)} \circ \nabla \log p_{\x{X}_{k}|\x{X}_0}\!\left(\x{x}_k^{(i)}\bigm\vert\x{x}_0^{(i)}\right) - \x{x}_{k}^{(i)} \circ s_{\x{\theta}}(\x{x}_{k}^{(i)}, k) \Bigr\|^2_2 \right],
    \label{eq:mdsm_monte_carlo}
\end{equation}
where $k = 0, 1, \dots, N-1$ indexes the discretized time-step and $i = 1, \dots, M$ indexes the training sample.

\begin{table}[b]
    \centering
    \begin{tabular}{c|c|c}
        \hline \hline
        \textbf{Dataset} & $\hat{\mu}$ & $\hat{\sigma}$ \\ \hline
        Fashion-MNIST    & $0.2771$ & $0.8428$ \\ \hline
        Kuzushiji-MNIST  & $0.2140$ & $0.8290$ \\ \hline
        MNIST            & $0.1565$ & $0.8353$ \\ \hline
        CIFAR-10         & $[0.4075,\; 0.3785,\; 0.3693]$ & $[0.8016,\; 0.8004,\; 0.8005]$ \\ \hline
    \end{tabular}
    \caption{Estimated parameters $(\hat{\mu}, \hat{\sigma})$ used for lognormal noise initialization. For CIFAR-10, the $\hat{\mu}$ and $\hat{\sigma}$ values are given channel-wise.}
    \label{tab:mu_sigma_values}
\end{table}

We set $\x{\mu} = \frac{\sigma^2}{2}\x{1}$ to simplify the update rule. The optimal values of $\delta$, $L$, and $\chi$ are selected via an empirical grid search, drawing $1{,}024$ samples per configuration and selecting the setting that minimizes FID and KID. The resulting configurations are reported in \cref{tab:parameters_sampling}. Specifically for the Fashion-MNIST and Kuzushiji-MNIST datasets the optimal values of  $\delta$, $L$, and $\chi$ starting from class-averaged initialization are mentioned in \cref{tab:sign_agnostic_class_avg}, for all other datasets, the optimal $\delta$, $L$, and $\chi$ values are identical to those mentioned in \cref{tab:parameters_sampling}.

\begin{table}[!ht]
    \centering
    \begin{tabular}{|c|c|c|c|c|}
        \hline
        \textbf{Sampler} & \textbf{Dataset} & \textbf{$L$} & \textbf{$\delta$ (approx.)} & \textbf{$\chi$} \\ \hline
        \multirow{4}{*}{ADLS (\cref{algo:gbm_sampler_anneal})}
            & Fashion-MNIST   & $1$ & $5.86 \times 10^{-4}$ & $0.9950$ \\ \cline{2-5}
            & Kuzushiji-MNIST & $3$ & $2.11 \times 10^{-4}$ & $0.9950$ \\ \cline{2-5}
            & MNIST           & $1$ & $6.06 \times 10^{-4}$ & $0.9950$ \\ \cline{2-5}
            & CIFAR-10        & $6$ & $1.14 \times 10^{-4}$ & $0.9995$ \\ \hline
        \multirow{4}{*}{ASAMS (\cref{algo:gbm_sampler_nlamp_anneal})}
            & Fashion-MNIST   & $3$ & $1.43 \times 10^{-4}$ & $0.9995$ \\ \cline{2-5}
            & Kuzushiji-MNIST & $3$ & $2.11 \times 10^{-4}$ & $1.0000$ \\ \cline{2-5}
            & MNIST           & $4$ & $1.43 \times 10^{-4}$ & $0.9950$ \\ \cline{2-5}
            & CIFAR-10        & $5$ & $1.43 \times 10^{-4}$ & $0.9995$ \\ \hline
    \end{tabular}
    \caption{Optimal $L$, $\delta$, and $\chi$ per dataset for lognormal noise initialization, selected to minimize FID and KID. ADLS = Annealed Dale-Langevin Sampler; ASAMS = Annealed sign-agnostic Sampler.}
    \label{tab:parameters_sampling}
\end{table}
\begin{table}[!ht]
    \centering
    \renewcommand{\arraystretch}{1.5}
    \begin{tabular}{|c|c|c|c|c|}
        \hline
        \textbf{Sampler} & \textbf{Dataset} & \textbf{$L$} & \textbf{$\delta$ (approx.)} & \textbf{$\chi$} \\ \hline

        \multirow{4}{*}{ADLS (\cref{algo:gbm_sampler_anneal})}
            & Fashion-MNIST   & $1$ & $5.86 \times 10^{-4}$ & $0.9950$ \\ \cline{2-5}
            & Kuzushiji-MNIST & $3$ & $2.11 \times 10^{-4}$ & $0.9950$ \\ \cline{2-5}
            & MNIST           & $1$ & $6.06 \times 10^{-4}$ & $0.9950$ \\ \cline{2-5}
            & CIFAR-10        & $6$ & $1.14 \times 10^{-4}$ & $0.9995$ \\ \hline
        \multirow{4}{*}{ASAMS (\cref{algo:gbm_sampler_nlamp_anneal})}
            & Fashion-MNIST   & $3$ & $2.50 \times 10^{-4}$ & $0.9995$ \\ \cline{2-5}
            & Kuzushiji-MNIST & $2$ & $3.30 \times 10^{-4}$ & $0.9950$ \\ \cline{2-5}
            & MNIST           & $4$ & $1.43 \times 10^{-4}$ & $0.9950$ \\ \cline{2-5}
            & CIFAR-10        & $5$ & $1.43 \times 10^{-4}$ & $0.9995$ \\ \hline
    \end{tabular}
        
    \caption{Optimal $L$, $\delta$, and $\chi$ for ASAMS (\cref{algo:gbm_sampler_nlamp_anneal}) initialized from class-averaged samples passed through the forward process. ASAMS = Annealed sign-agnostic Sampler.}
    \label{tab:sign_agnostic_class_avg}
\end{table}

\FloatBarrier

\section{Generated Samples}
\label{asec:images}

The samples are generated using the trained model and the sampling algorithms described in \cref{algo:gbm_sampler_anneal} and \cref{algo:gbm_sampler_nlamp_anneal}. We observe that the generated samples are diverse and resemble the training data. They are also noise-free, which goes to show that the Annealed multiplicative sampling update is quite robust. There are some samples that are entirely novel and are not identical to the training data. This effect is more pronounced in MNIST and Kuzushiji MNIST datasets. Samples from the Fashion MNIST dataset are less diverse and seem to have latched on to certain modes of the training data. This is by no means evidence of mode collapse but certain classes are underrepresented in the generation. This is probably because the Fashion MNIST dataset is more complex and has more variability in the images compared to MNIST and Kuzushiji MNIST.

\section{Evaluation Metrics for the Generated Images}
We use the following metrics to evaluate the quality of the generated images:
\begin{itemize}
    \item \textbf{Fréchet Inception Distance (FID)}~\citep{Frechet17}, which measures the distance between the distribution of generated images and real images in the feature space of a pre-trained InceptionV3 network~\citep{inceptionv3}. Lower values indicate better quality.
    \item \textbf{Kernel Inception Distance (KID)}~\citep{binkowski2018demystifying}, which is similar to FID, but uses a kernel to measure the distance between distributions. It is less sensitive to outliers and is more robust for small sample sizes.
    \item \textbf{Nearest neighbours from training data}, which is a qualitative measure of how closely the generated samples resemble the training data and to rule out the possibility of memorization of the training samples. The nearest neighbours are identified by measuring the Euclidean distance between generated samples and images from the training data with distances measured both in the pixel space and InceptionV3 feature space.
\end{itemize}

\subsection{Nearest neighbours}
To verify that the model generalizes rather than memorizes, we retrieve the $10$ nearest neighbours of each generated sample from the training set using the Euclidean ($\ell_2$) distance measured in two complementary spaces: directly in pixel space (raw image intensities) and in the InceptionV3 feature space (semantic embeddings extracted from a pretrained InceptionV3 network). Pixel-space retrieval checks for low-level visual similarity, while feature-space retrieval probes semantic similarity independent of pixel-level variation. Across both metrics, the retrieved neighbours are semantically coherent with the generated samples but not identical, confirming that the model captures the underlying data distribution without collapsing to training set memorization. For our choice of datasets, we have found that the nearest neighbour in pixel space provided a significant interpretative value.

\begin{figure}[htbp]
  \centering
  \subfloat[InceptionV3 feature]{\includegraphics[page=2, width=0.49\textwidth]{figures/dale-langevin-nn-cifar10.pdf}}
  \hfill
  \subfloat[Raw Image as feature]{\includegraphics[page=1, width=0.49\textwidth]{
  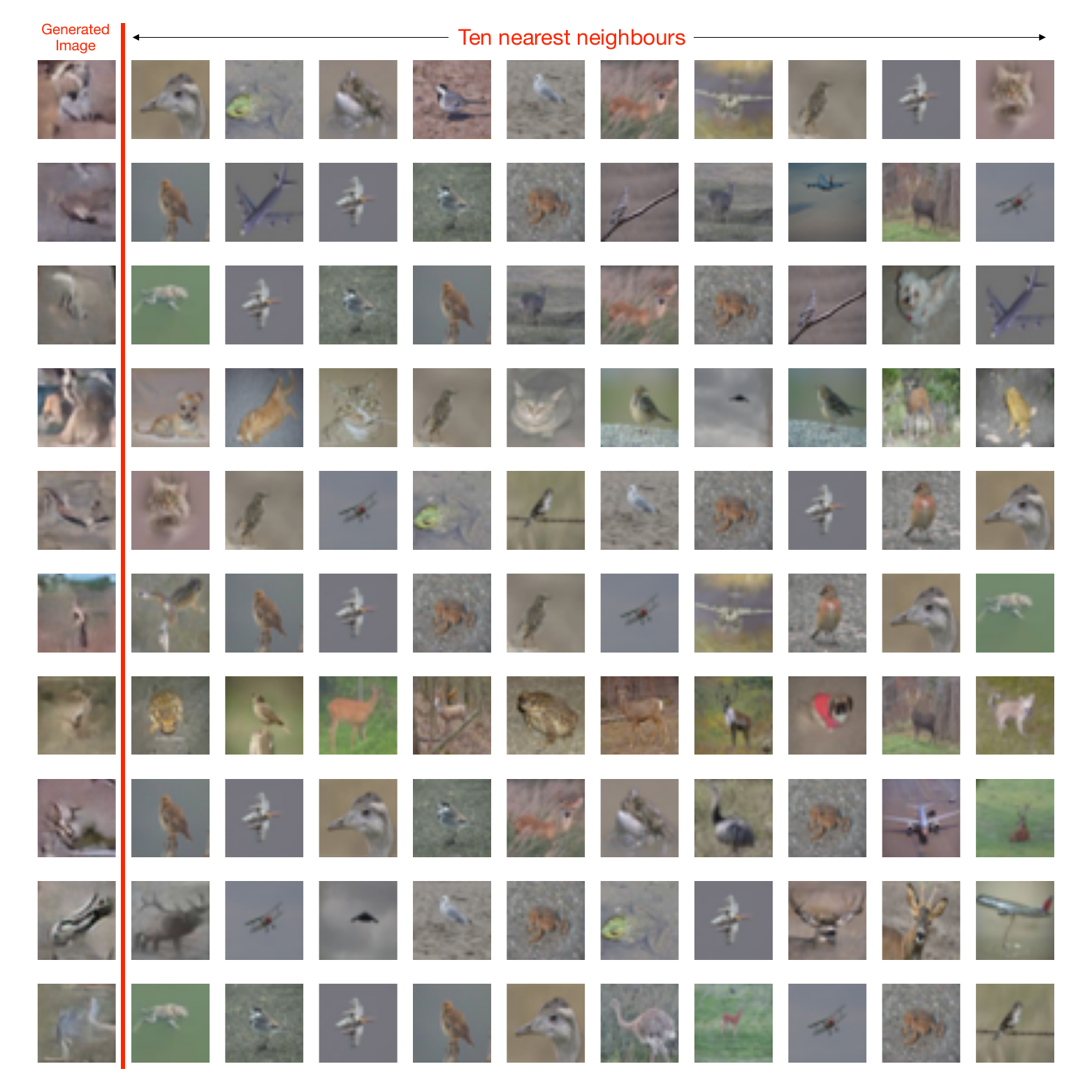}}
  \caption{$10$ nearest neighbours (calculated using Euclidean distance on InceptionV3 features and raw images respectively) from the CIFAR-10 dataset for samples generated with lognormal noise initialization using Dale-Langevin sampler \cref{eqn:gbmsde_sampling_a}.}
\end{figure}

\begin{figure}[htbp]
  \centering
  \subfloat[InceptionV3 feature]{\includegraphics[page=4, width=0.49\textwidth]{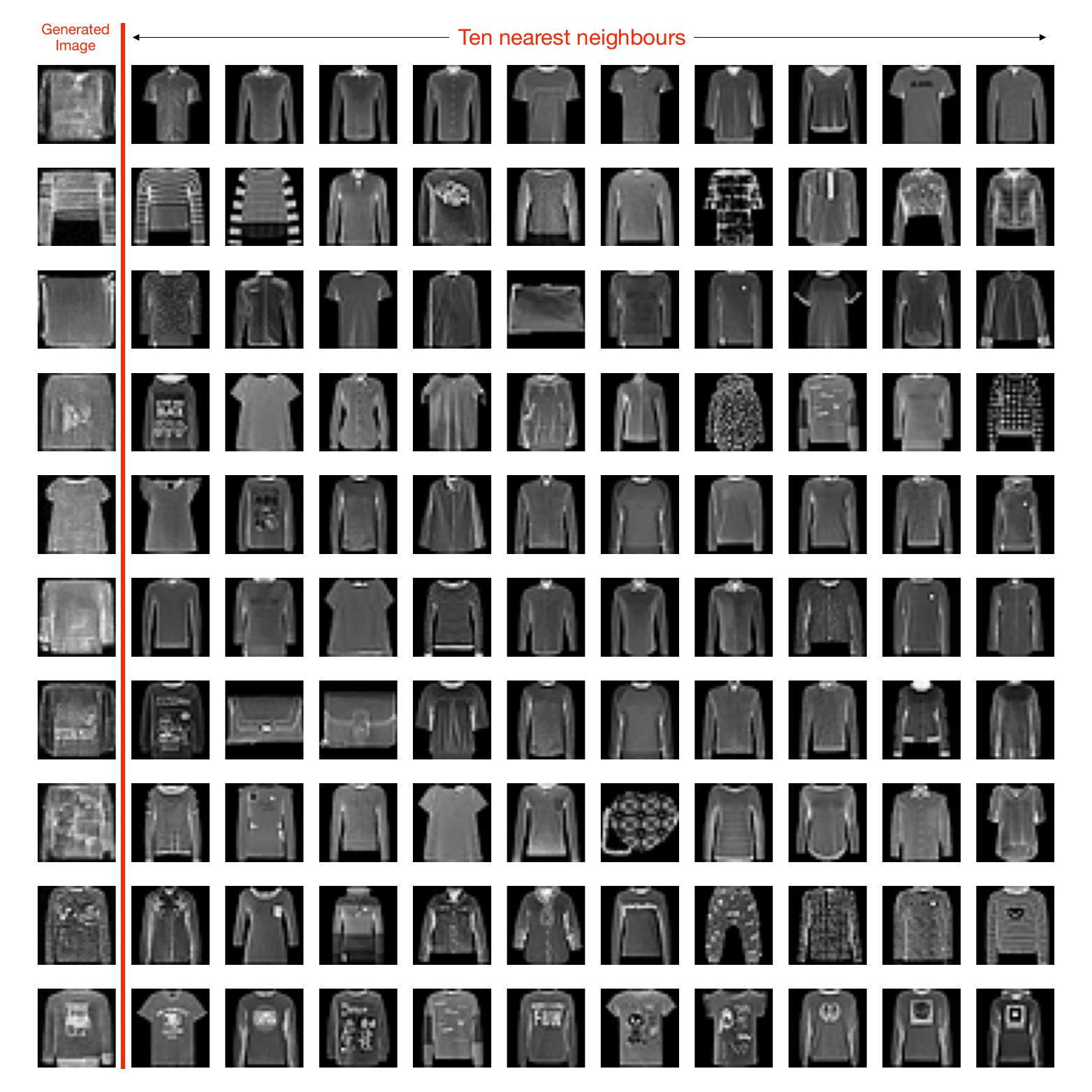}}
  \hfill
  \subfloat[Raw Image as feature]{\includegraphics[page=3, width=0.49\textwidth]{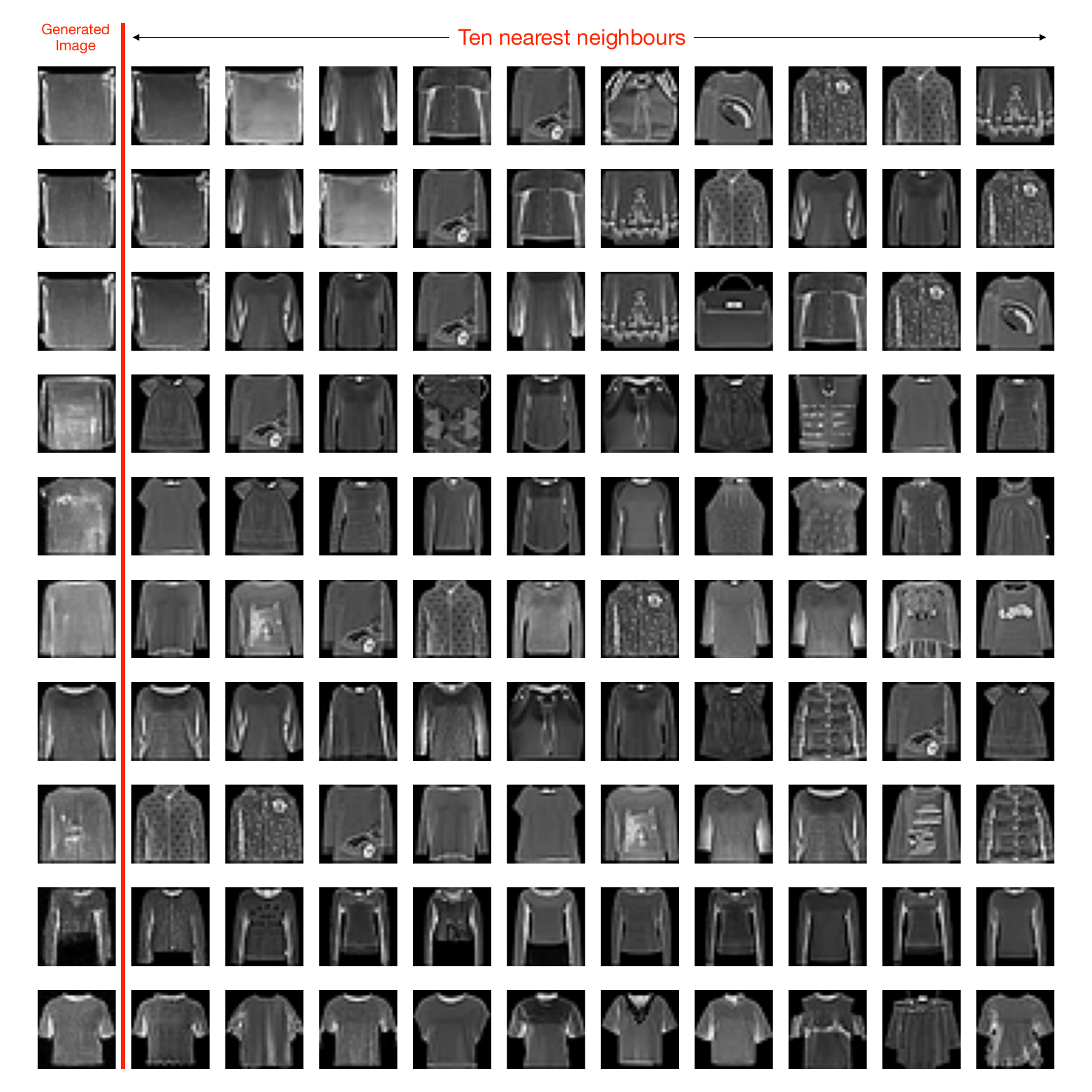}}
  \caption{$10$ nearest neighbours (calculated using Euclidean distance on InceptionV3 features and raw images respectively) from the Fashion MNIST dataset for samples generated with lognormal noise initialization using Dale-Langevin sampler \cref{eqn:gbmsde_sampling_a}.}
\end{figure}

\begin{figure}[htbp]
  \centering
  \subfloat[InceptionV3 feature]{\includegraphics[page=6, width=0.49\textwidth]{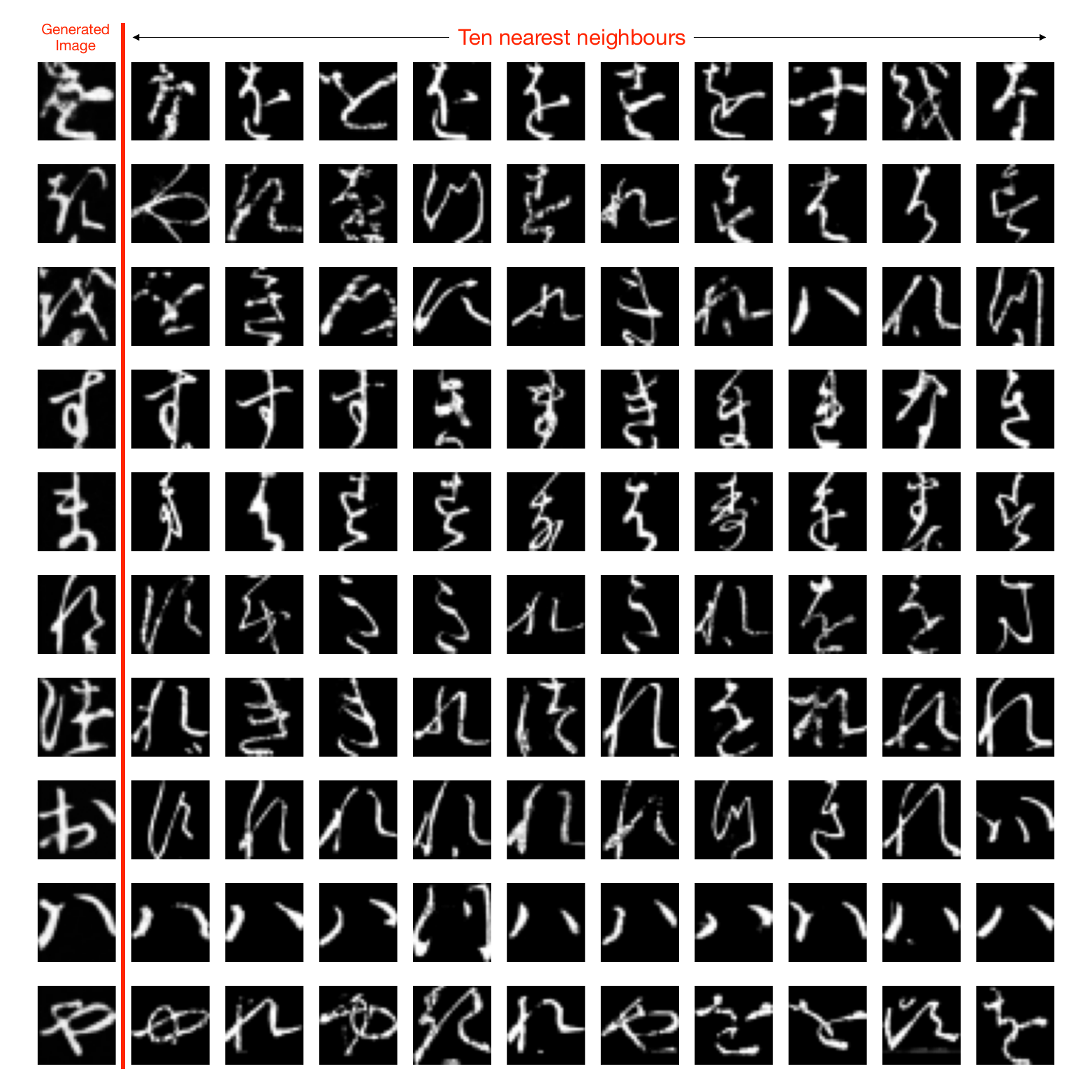}}
  \hfill
  \subfloat[Raw Image as feature]{\includegraphics[page=5, width=0.49\textwidth]{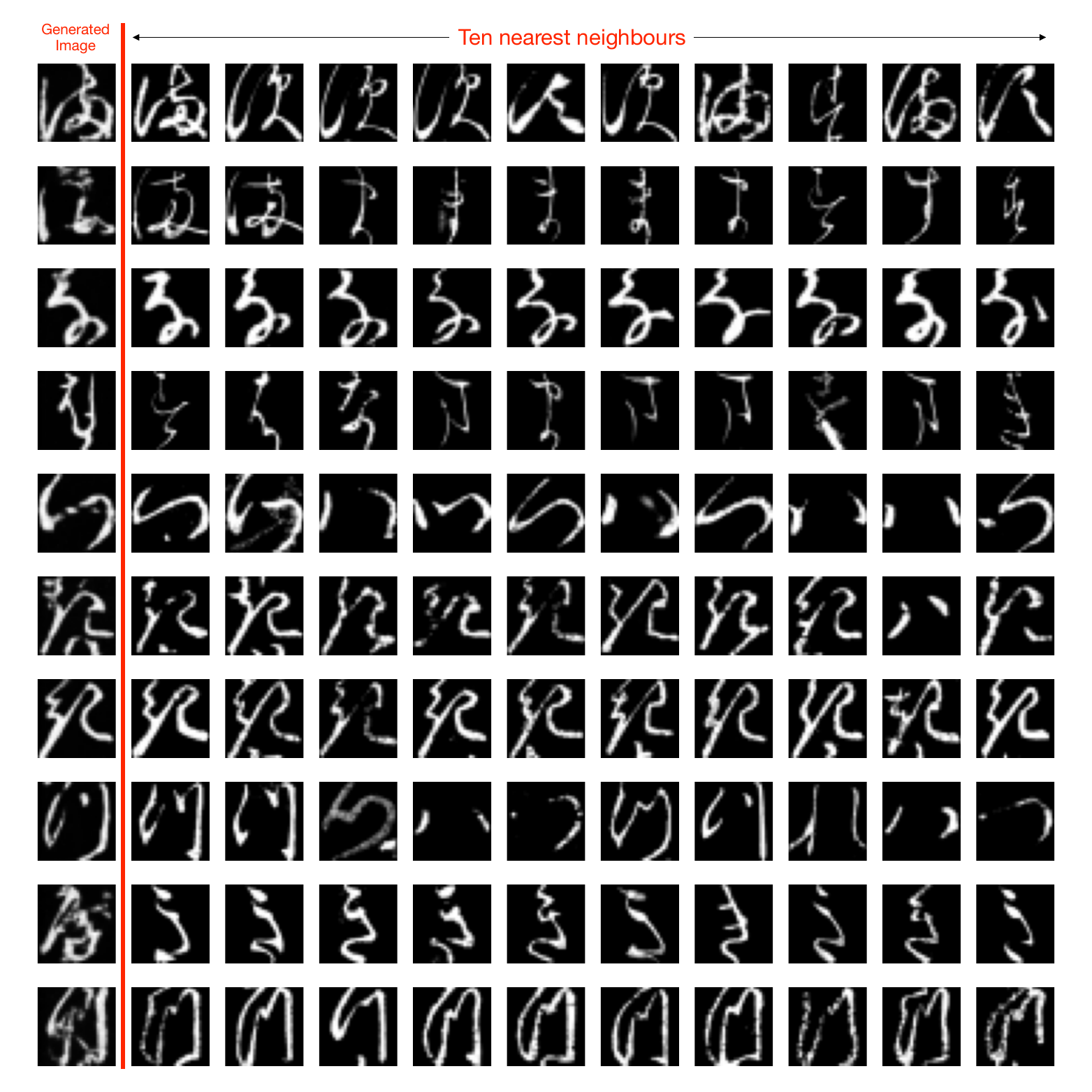}}
  \caption{$10$ nearest neighbours (calculated using Euclidean distance on InceptionV3 features and raw images respectively) from the Kuzushiji MNIST dataset for samples generated with lognormal noise initialization using Dale-Langevin sampler \cref{eqn:gbmsde_sampling_a}.}
\end{figure}

\begin{figure}[htbp]
  \centering
  \subfloat[InceptionV3 feature]{\includegraphics[page=8, width=0.49\textwidth]{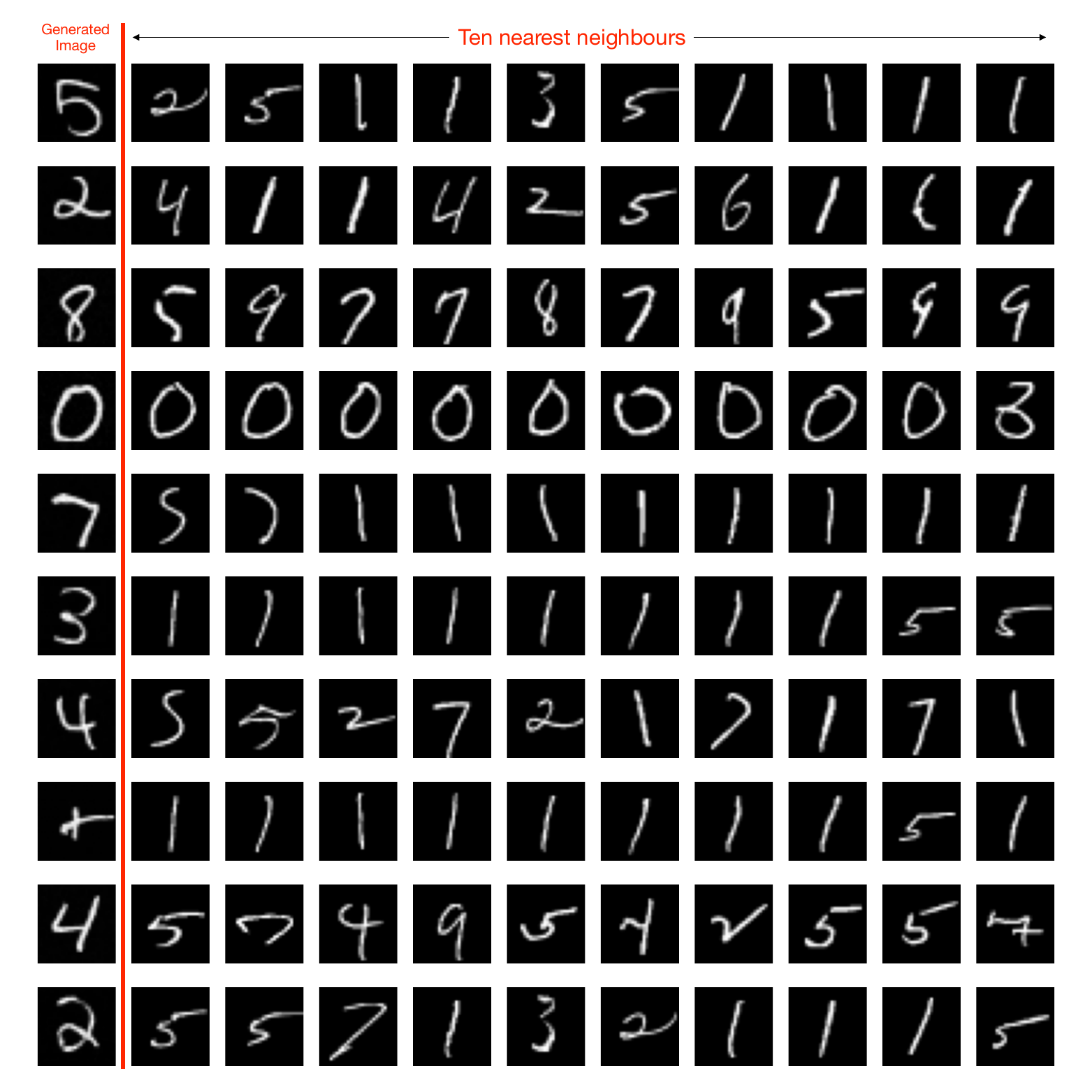}}
  \hfill
  \subfloat[Raw Image as feature]{\includegraphics[page=7, width=0.49\textwidth]{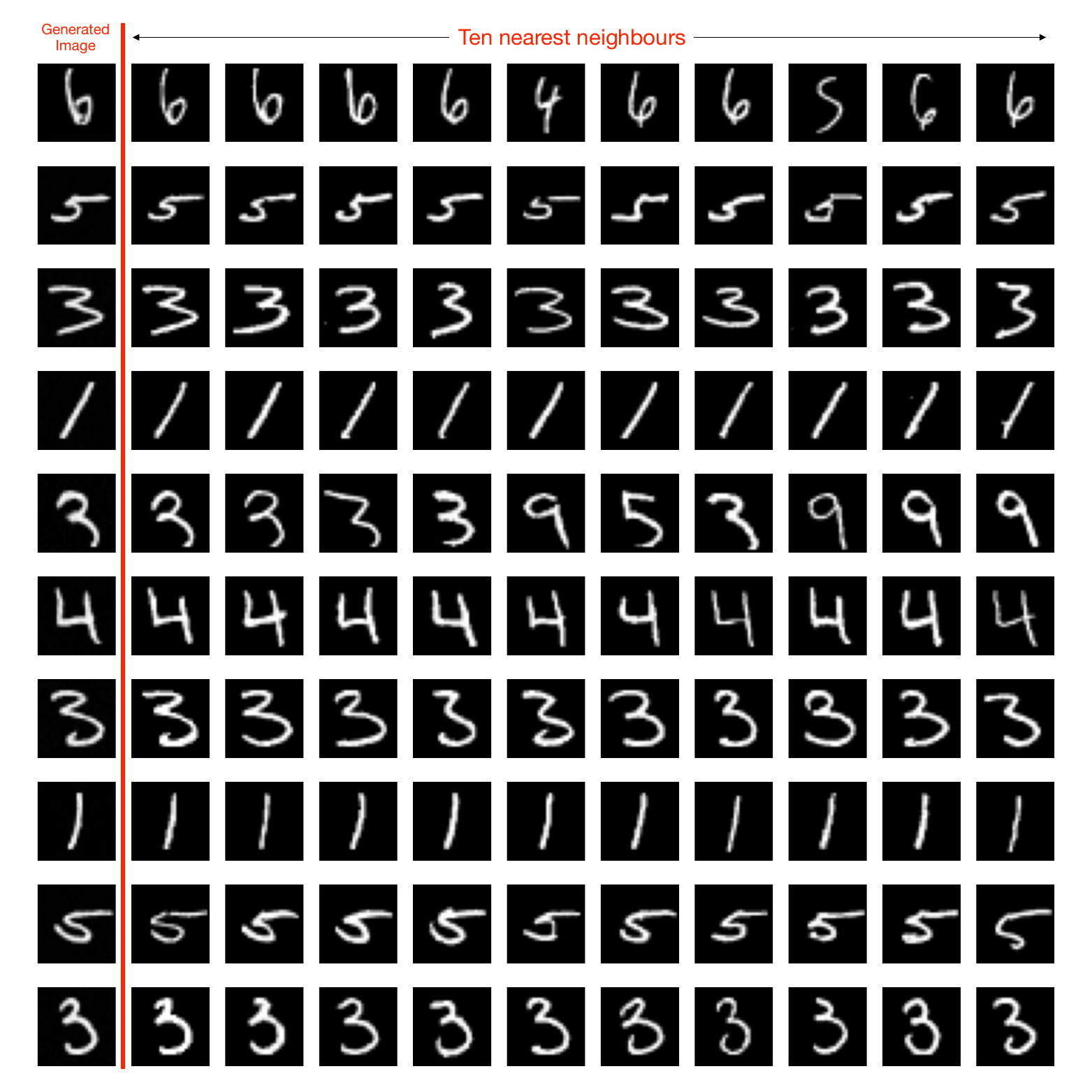}}
  \caption{$10$ nearest neighbours (calculated using Euclidean distance on InceptionV3 features and raw images respectively) from the MNIST dataset for samples generated with lognormal noise initialization using Dale-Langevin sampler \cref{eqn:gbmsde_sampling_a}.}
\end{figure}

\begin{figure}[htbp]
  \centering
  \subfloat[InceptionV3 feature]{\includegraphics[page=17, width=0.49\textwidth]{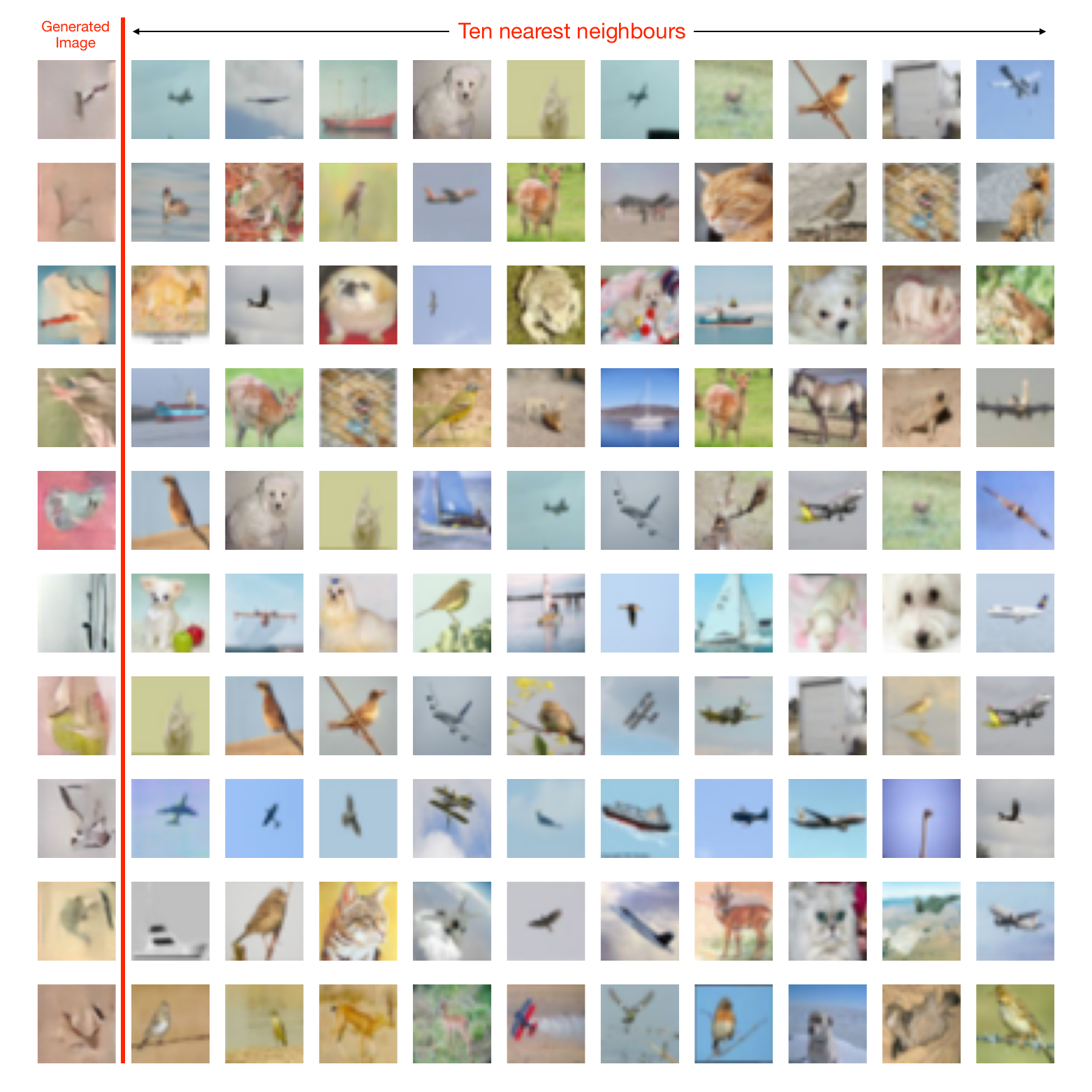}}
  \hfill
  \subfloat[Raw Image as feature]{\includegraphics[page=18, width=0.49\textwidth]{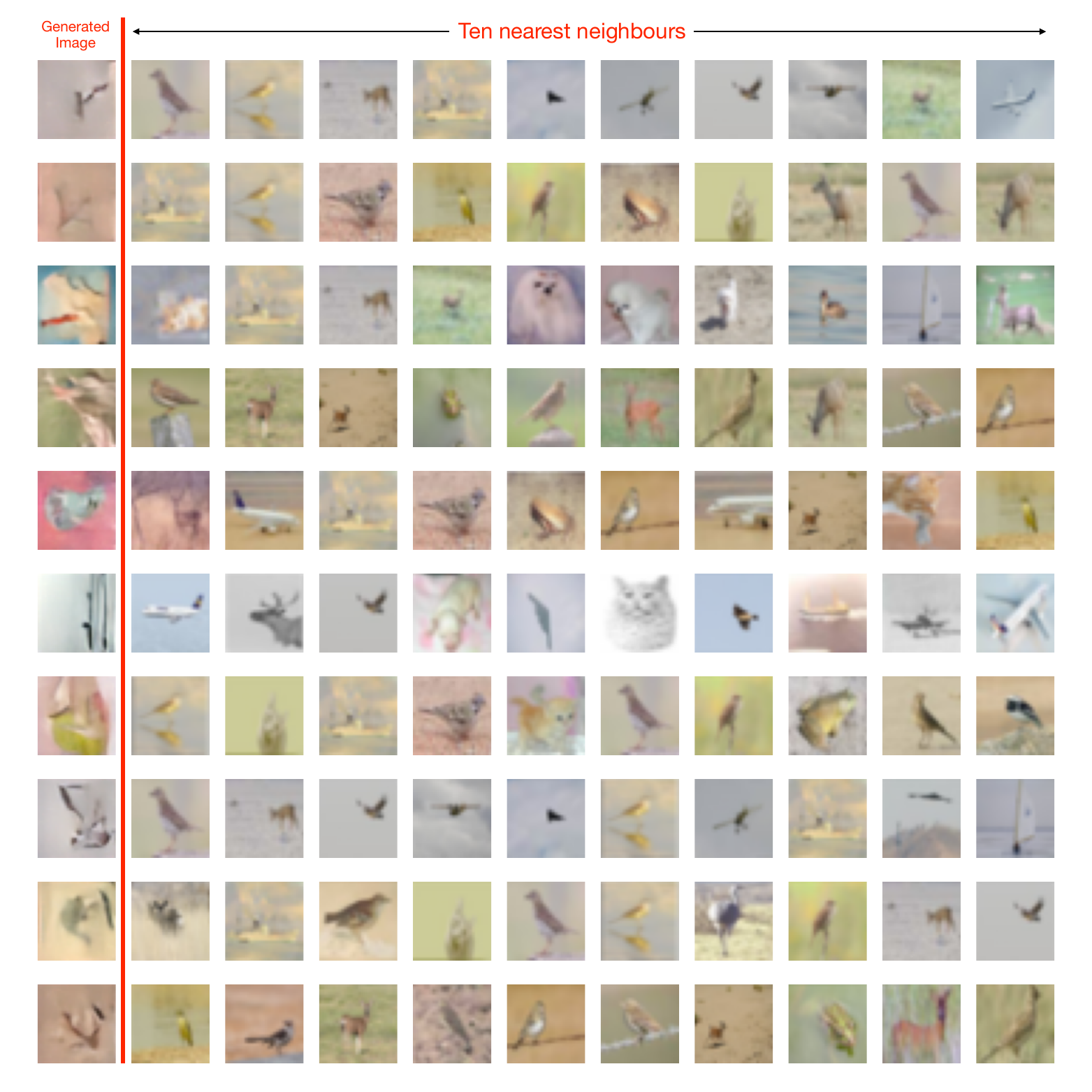}}
  \caption{$10$ nearest neighbours (calculated using Euclidean distance on InceptionV3 features and raw images respectively) from the CIFAR-10 dataset for samples generated with class averaged image passed through forward process as initialization using Dale-Langevin sampler \cref{eqn:gbmsde_sampling_a}.}
\end{figure}

\begin{figure}[htbp]
  \centering
  \subfloat[InceptionV3 feature]{\includegraphics[page=10, width=0.49\textwidth]{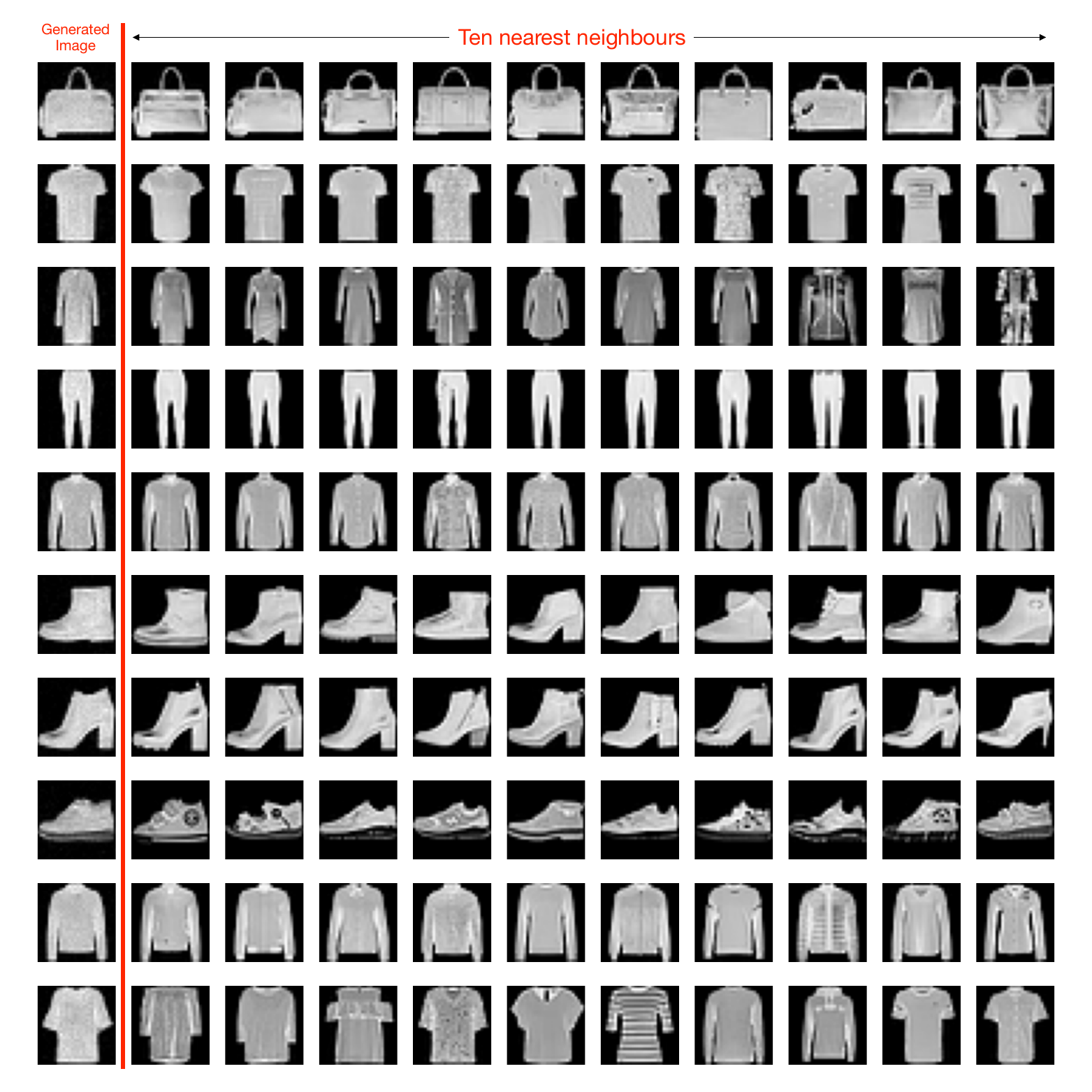}}
  \hfill
  \subfloat[Raw Image as feature]{\includegraphics[page=9, width=0.49\textwidth]{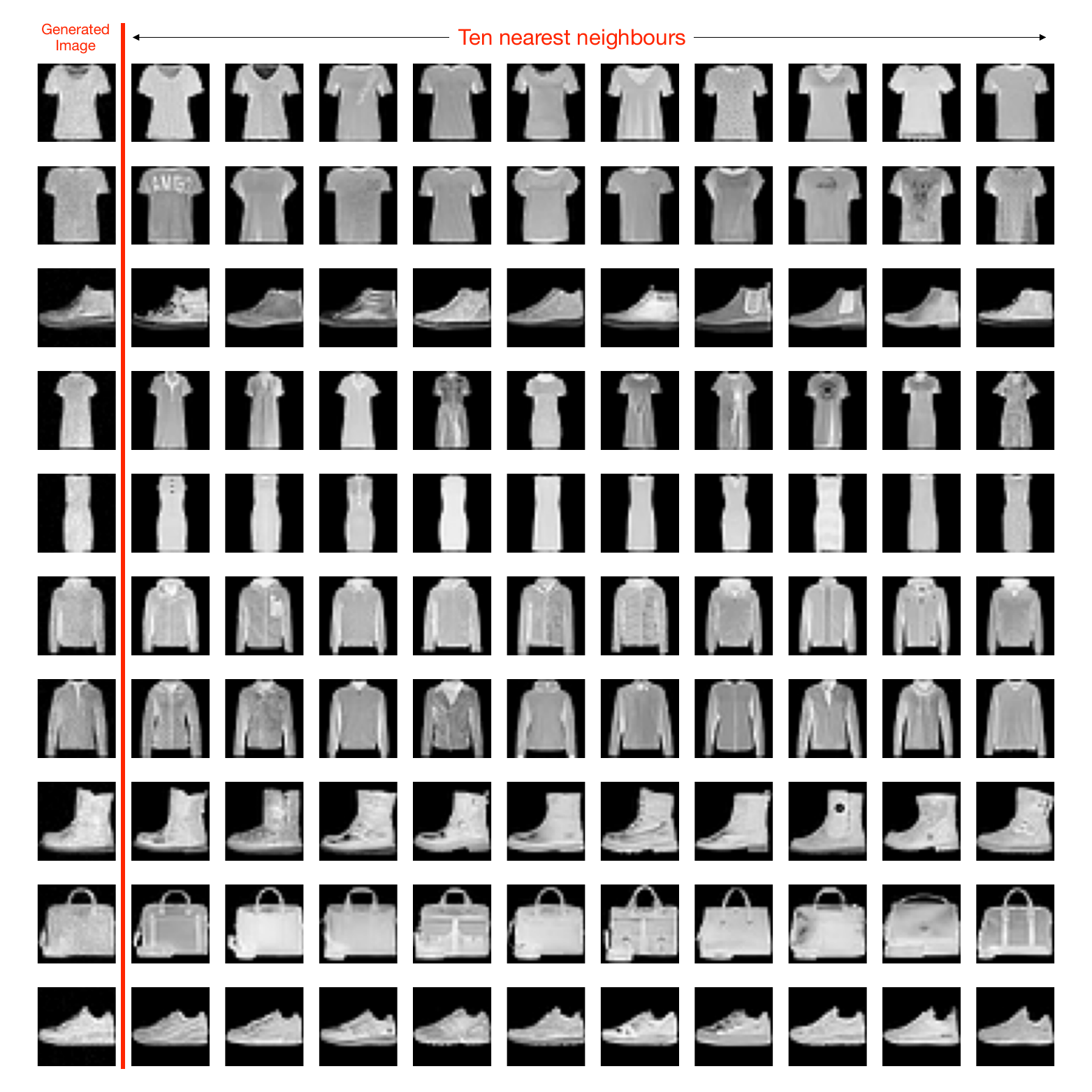}}
  \caption{$10$ nearest neighbours (calculated using Euclidean distance on InceptionV3 features and raw images respectively) from the Fashion MNIST dataset for samples generated with class averaged image passed through forward process as initialization using Dale-Langevin sampler \cref{eqn:gbmsde_sampling_a}.}
\end{figure}

\begin{figure}[htbp]
  \centering
  \subfloat[InceptionV3 feature]{\includegraphics[page=12, width=0.49\textwidth]{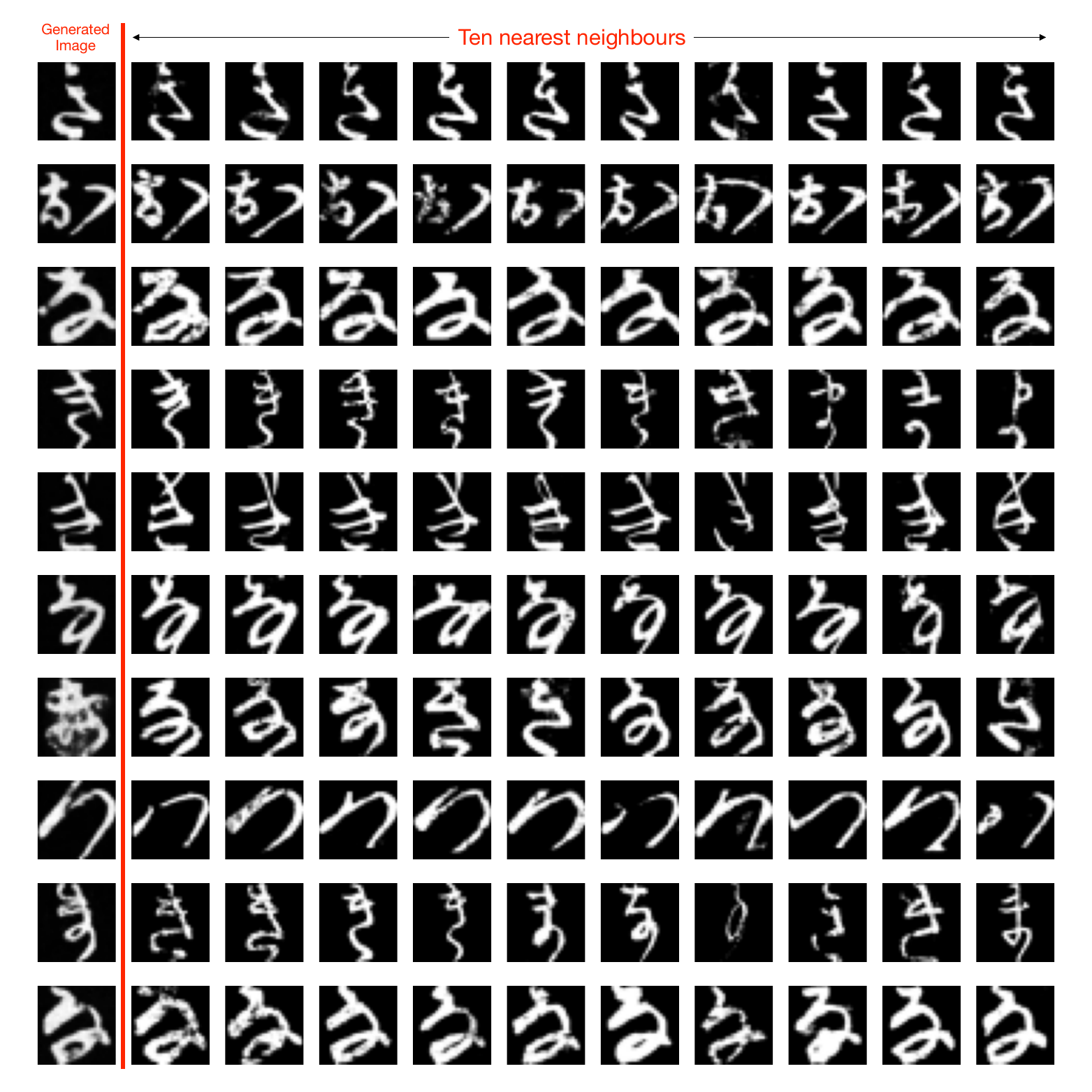}}
  \hfill
  \subfloat[Raw Image as feature]{\includegraphics[page=11, width=0.49\textwidth]{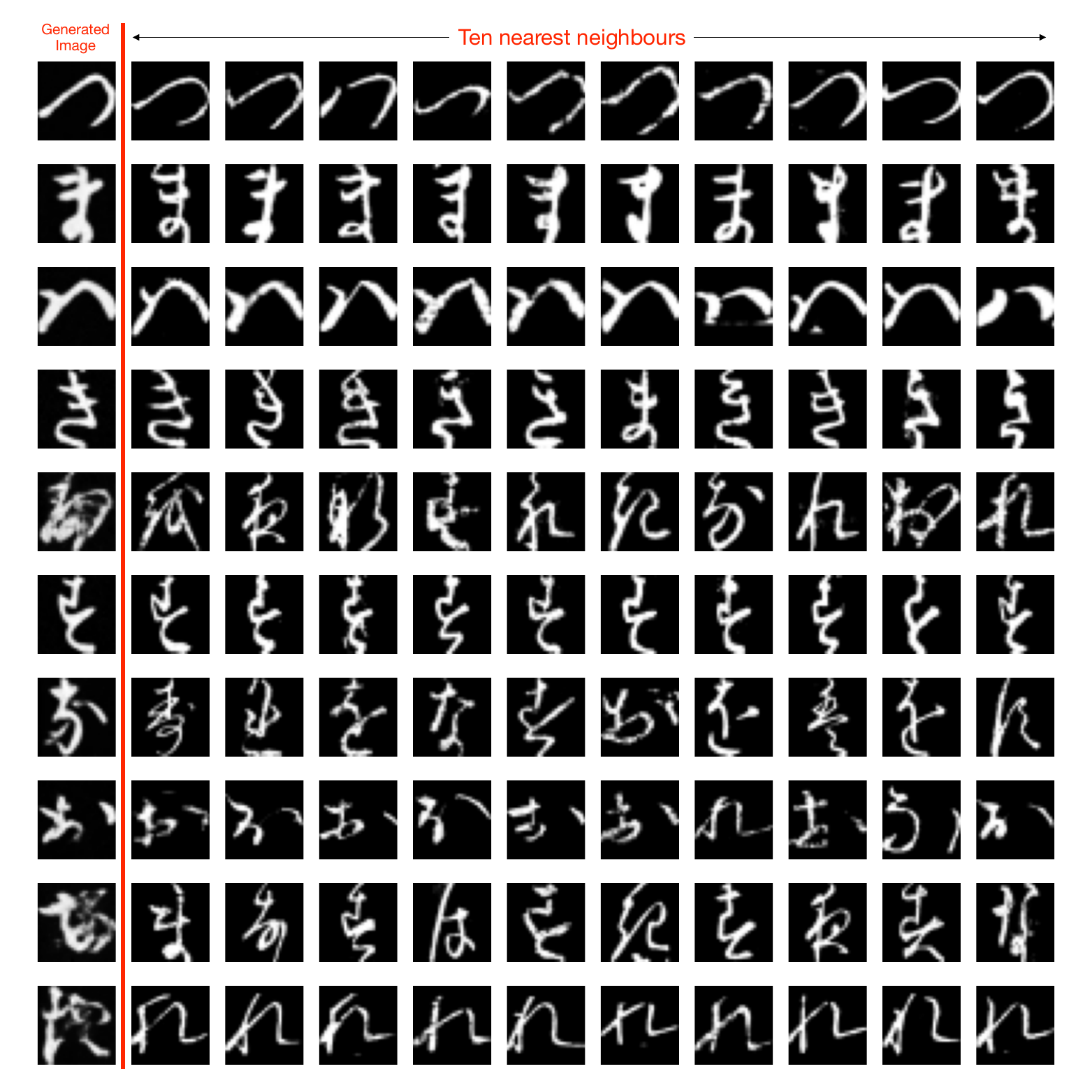}}
  \caption{$10$ nearest neighbours (calculated using Euclidean distance on InceptionV3 features and raw images respectively) from the Kuzushiji MNIST dataset for samples generated with class averaged image passed through forward process as initialization using Dale-Langevin sampler \cref{eqn:gbmsde_sampling_a}.}
\end{figure}

\begin{figure}[htbp]
  \centering
  \subfloat[InceptionV3 feature]{\includegraphics[page=14, width=0.49\textwidth]{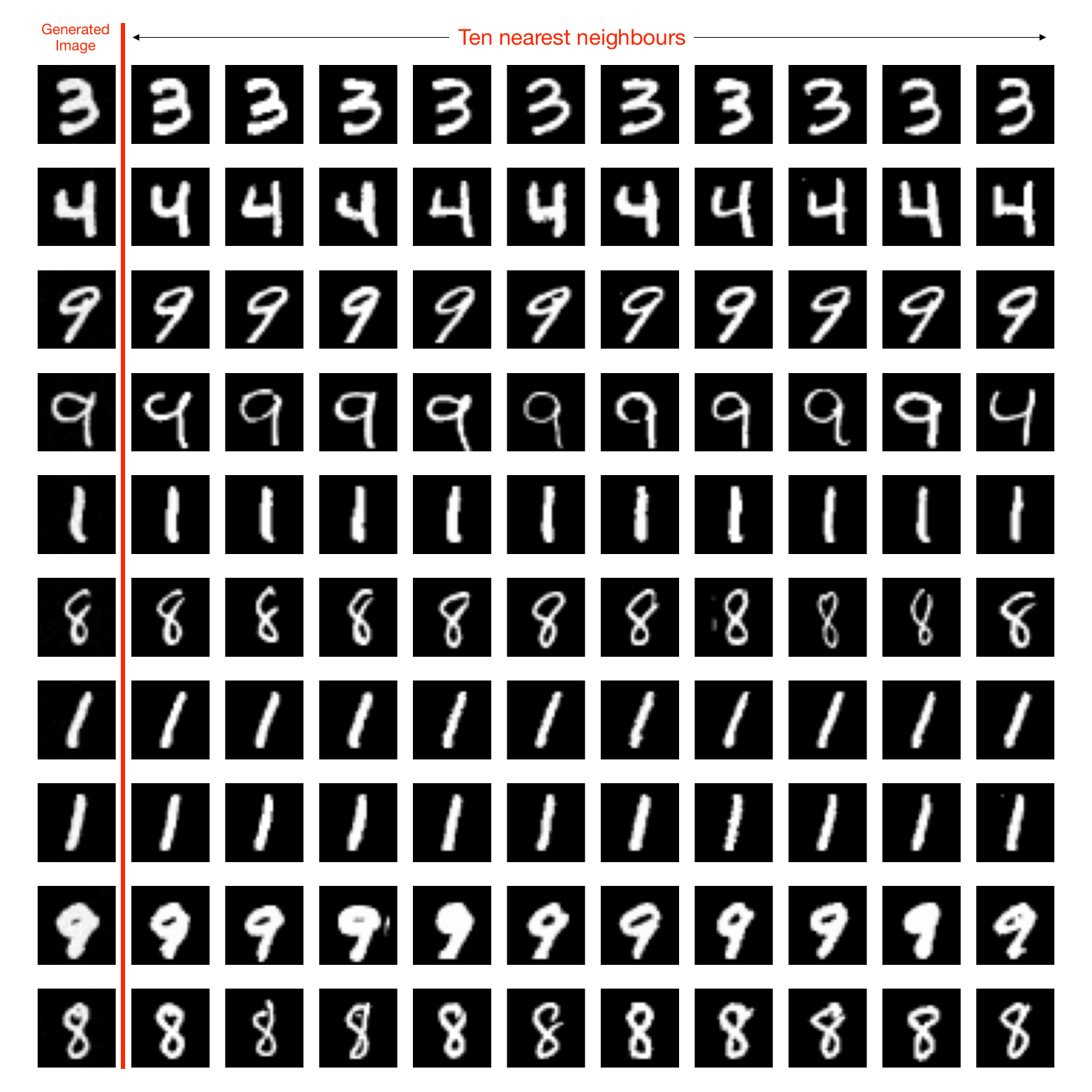}}
  \hfill
  \subfloat[Raw Image as feature]{\includegraphics[page=13, width=0.49\textwidth]{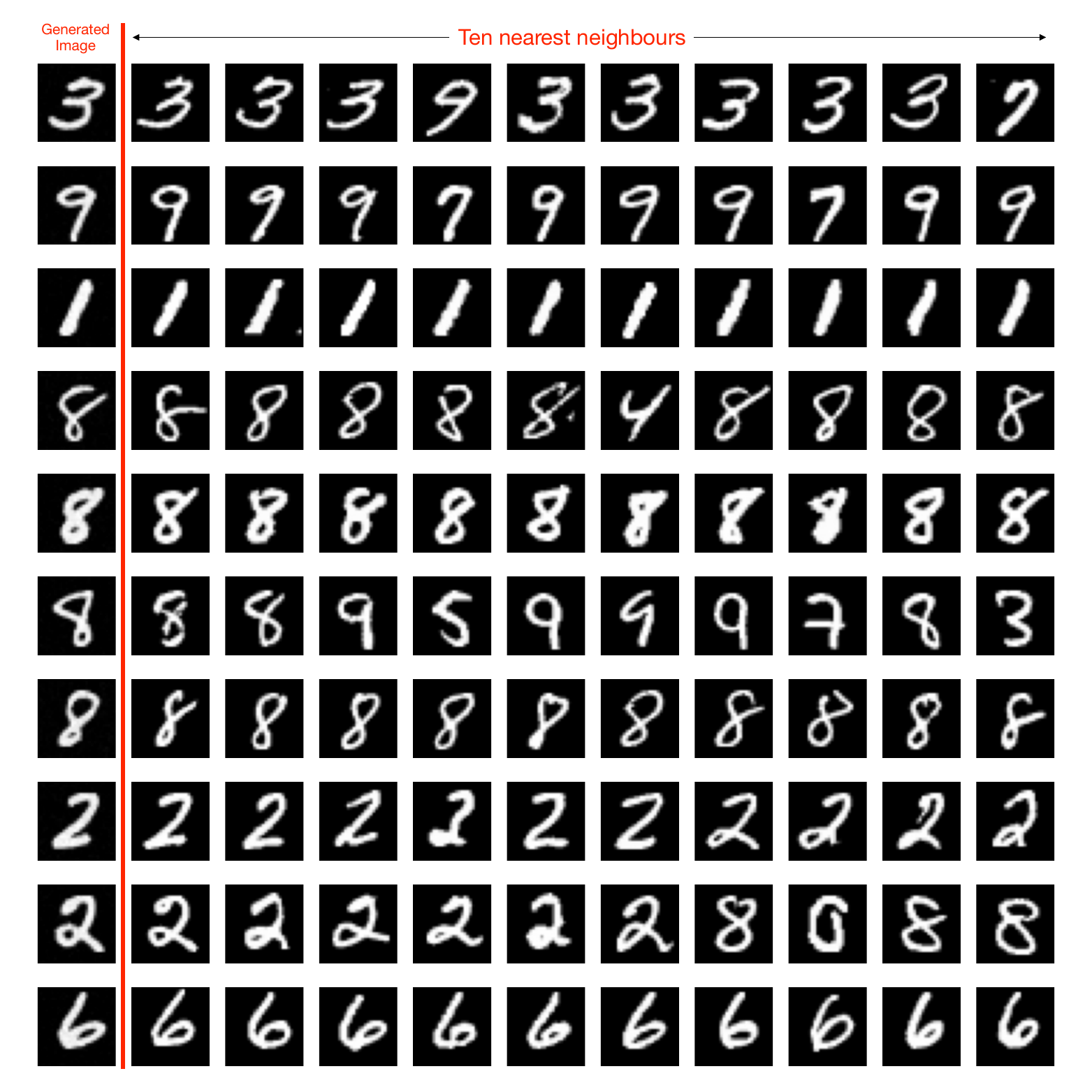}}
  \caption{$10$ nearest neighbours (calculated using Euclidean distance on InceptionV3 features and raw images respectively) from the MNIST dataset for samples generated with class averaged image passed through forward process as initialization using Dale-Langevin sampler \cref{eqn:gbmsde_sampling_a}.}
\end{figure}

\begin{figure}[htbp]
  \centering
  \subfloat[InceptionV3 feature]{\includegraphics[page=14,width=0.49\textwidth] {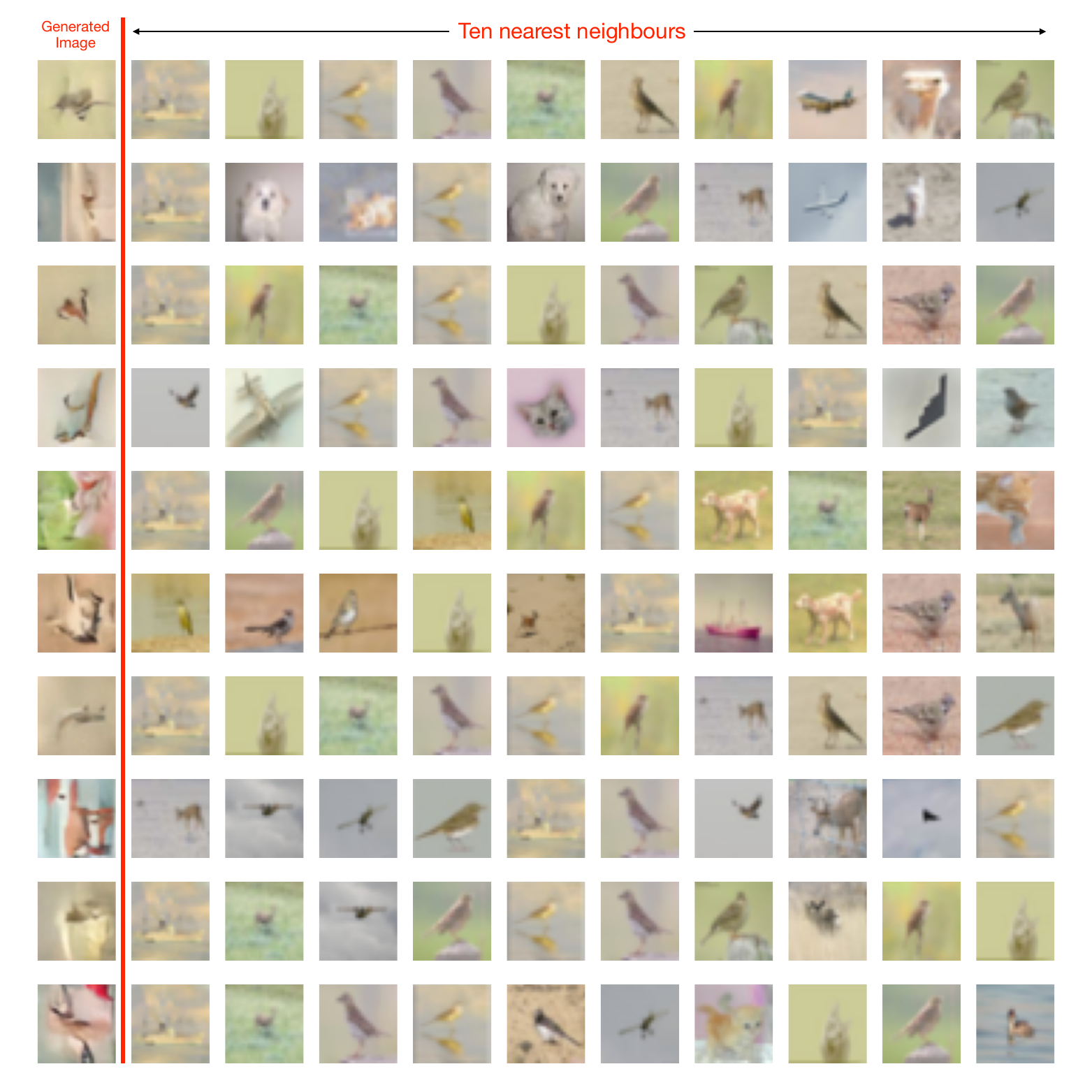}}
  \hfill
  \subfloat[Raw Image as feature]{\includegraphics[page=15,width=0.49\textwidth] {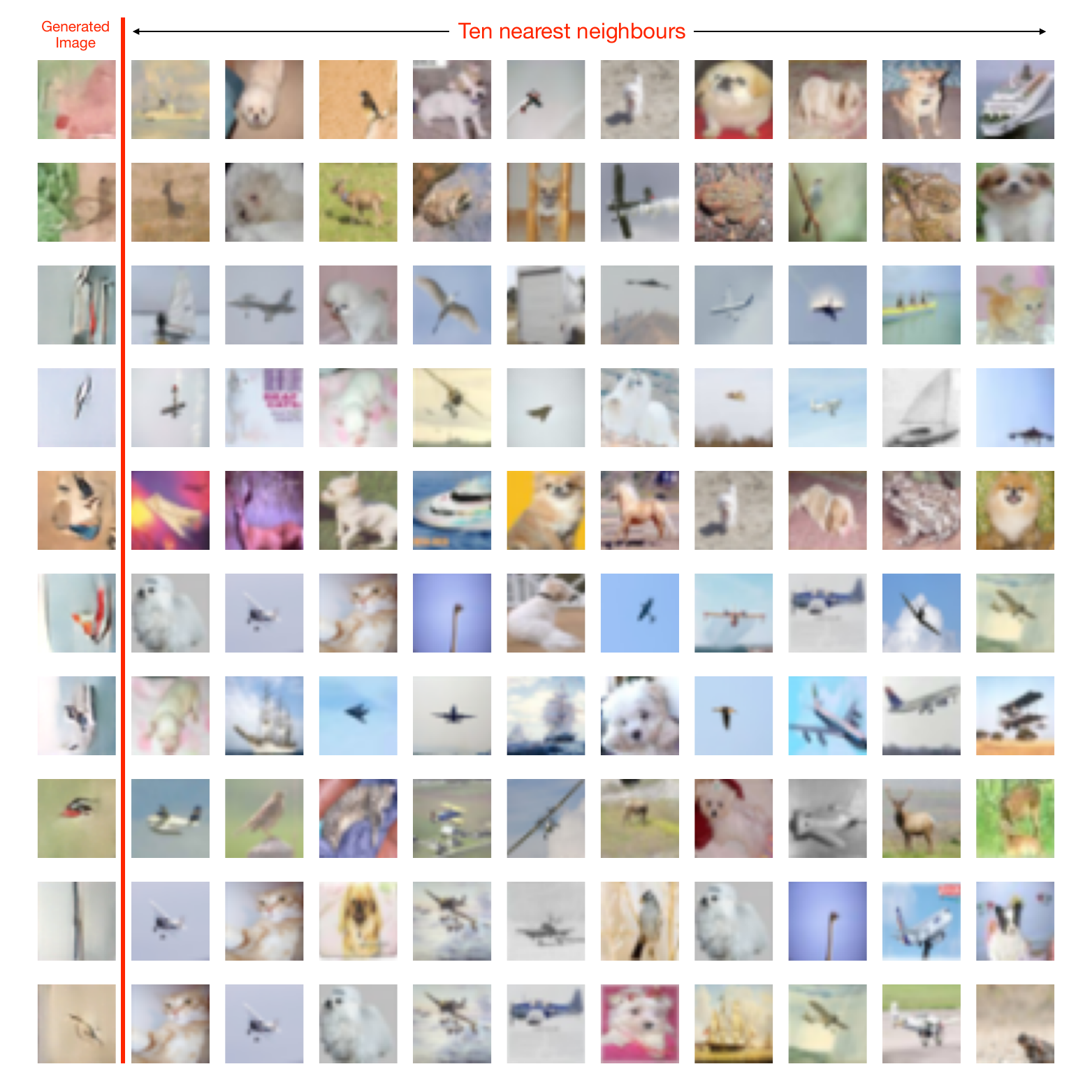}}
  \caption{$10$ nearest neighbours (calculated using Euclidean distance on InceptionV3 features and raw images respectively) from the CIFAR-10 dataset for samples generated with class averaged image passed through forward process as initialization using unconstrained multiplicative sampler \cref{eqn:gbmsde_backward_sampling}.}
\end{figure}

\begin{figure}[htbp]
  \centering
  \subfloat[InceptionV3 feature]{\includegraphics[page=1,width=0.49\textwidth] {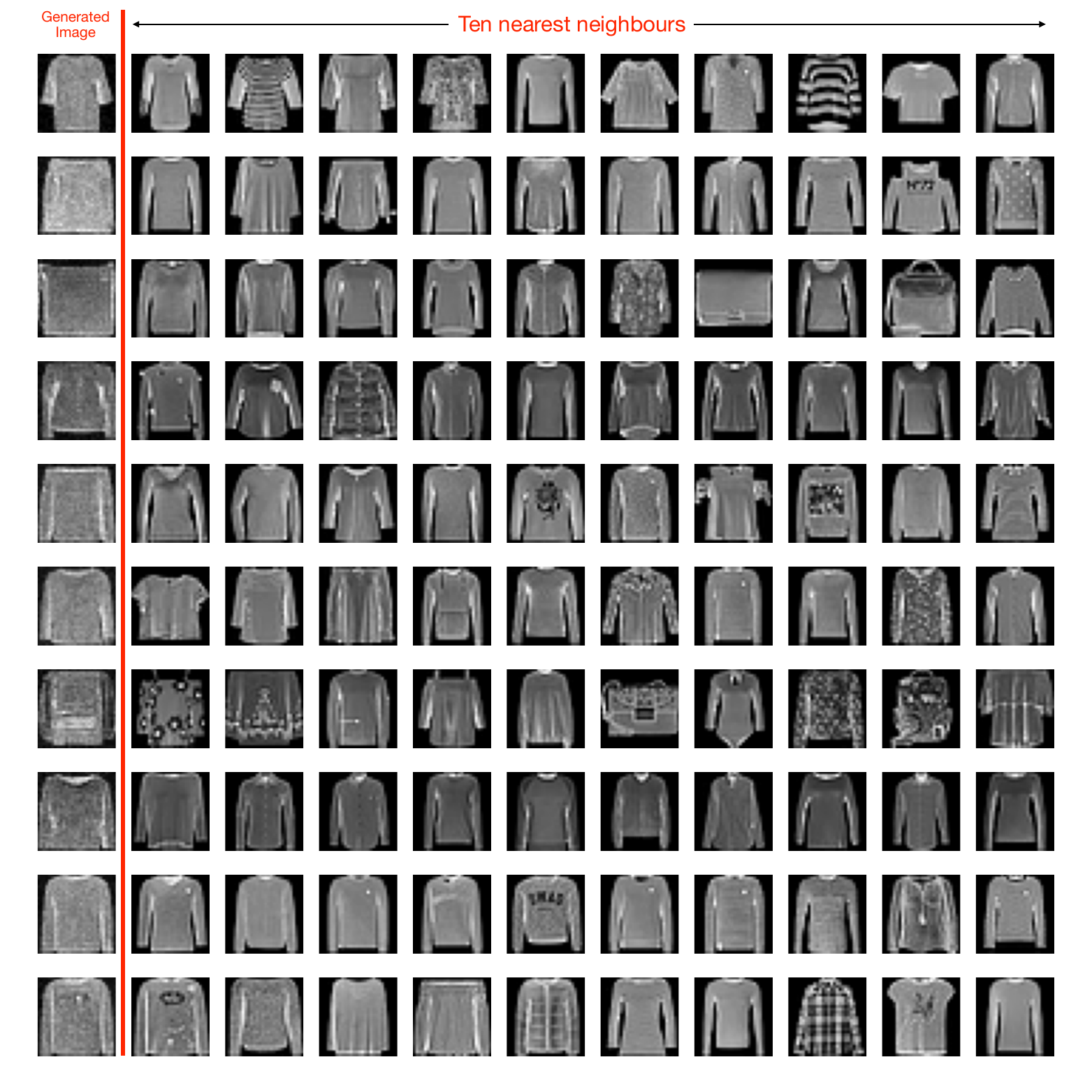}}
  \hfill
  \subfloat[Raw Image as feature]{\includegraphics[page=2,width=0.49\textwidth] {figures/non_lamperti/new_non_lamperti.pdf}}
  \caption{$10$ nearest neighbours (calculated using Euclidean distance on InceptionV3 features and raw images respectively) from the Fashion- MNIST dataset for samples generated with class averaged image passed through forward process as initialization using unconstrained multiplicative sampler \cref{eqn:gbmsde_backward_sampling}.}
\end{figure}

\begin{figure}[htbp]
  \centering
  \subfloat[InceptionV3 feature]{\includegraphics[page=3,width=0.49\textwidth] {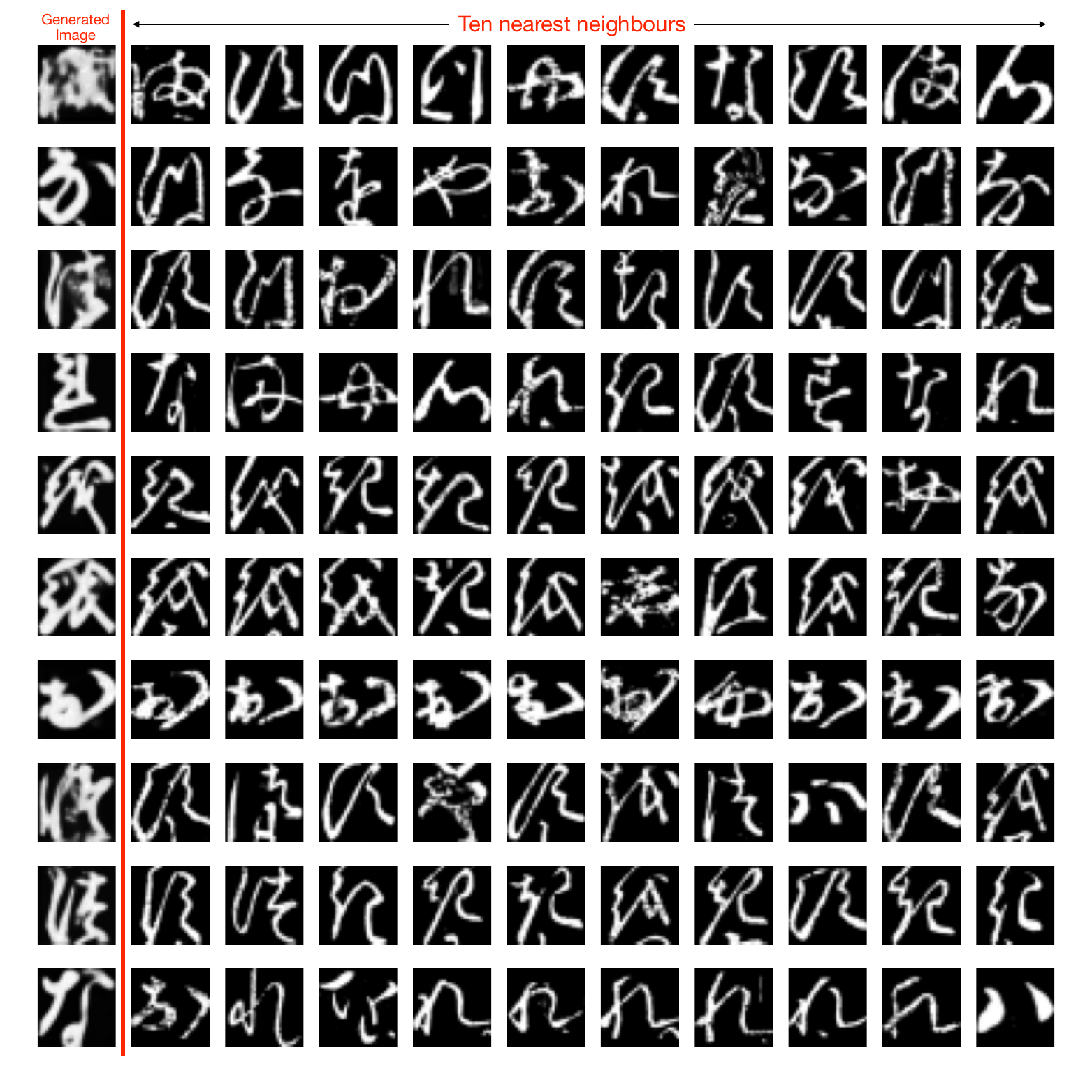}}
  \hfill
  \subfloat[Raw Image as feature]{\includegraphics[page=4,width=0.49\textwidth] {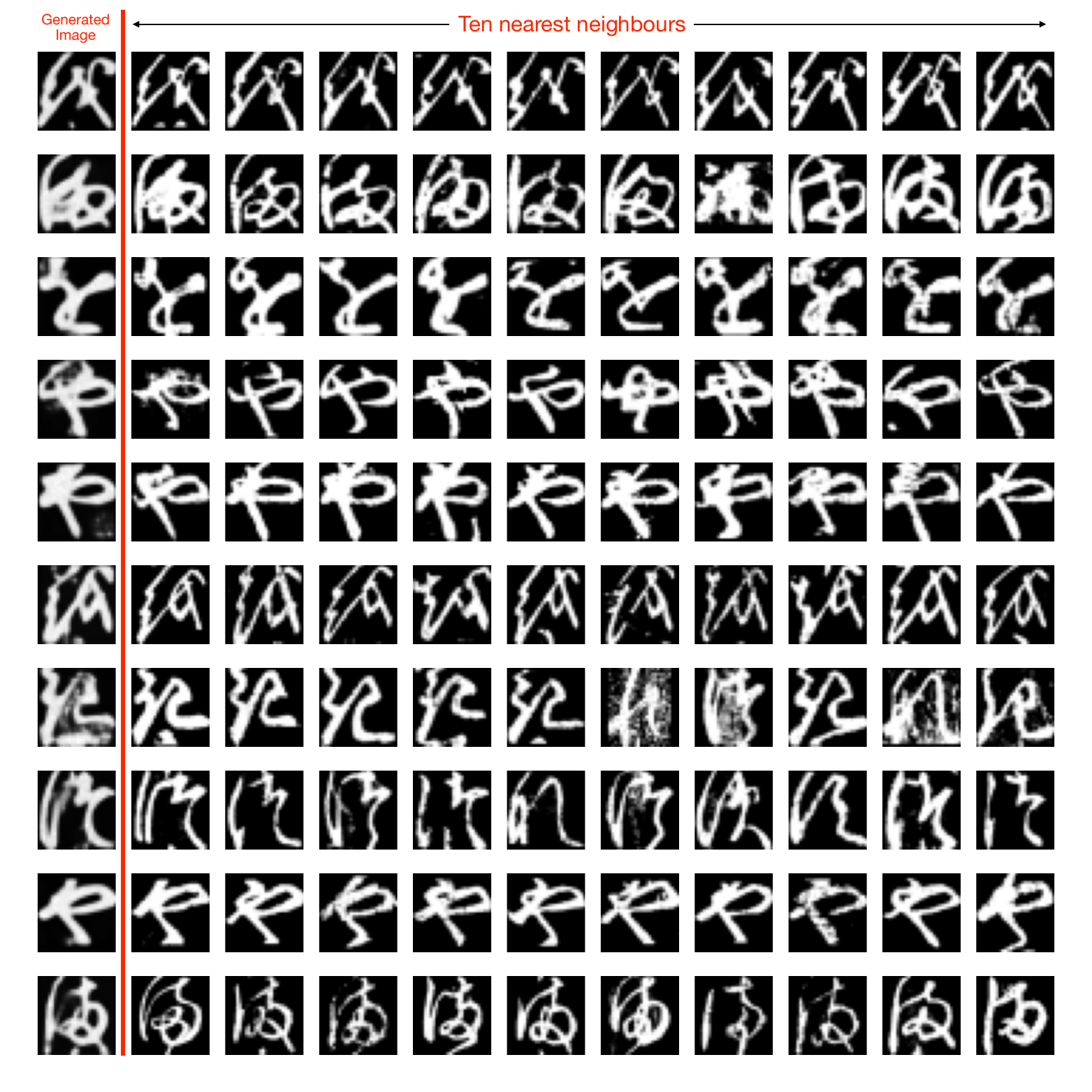}}
  \caption{$10$ nearest neighbours (calculated using Euclidean distance on InceptionV3 features and raw images respectively) from the Kuzushiji MNIST dataset for samples generated with class averaged image passed through forward process as initialization using unconstrained multiplicative sampler \cref{eqn:gbmsde_backward_sampling}.}
\end{figure}

\begin{figure}[htbp]
  \centering
  \subfloat[InceptionV3 feature]{\includegraphics[page=5,width=0.49\textwidth] {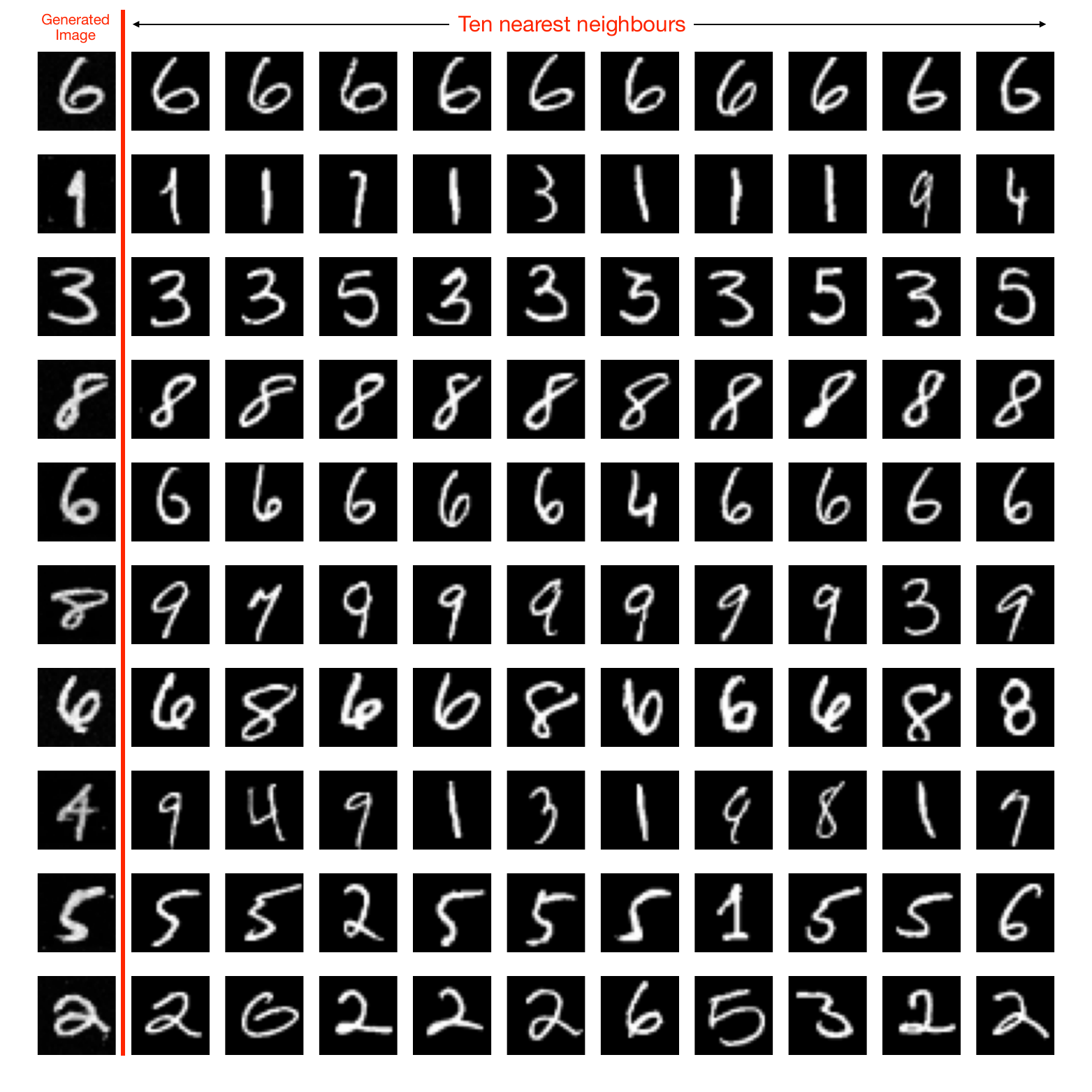}}
  \hfill
  \subfloat[Raw Image as feature]{\includegraphics[page=6,width=0.49\textwidth] {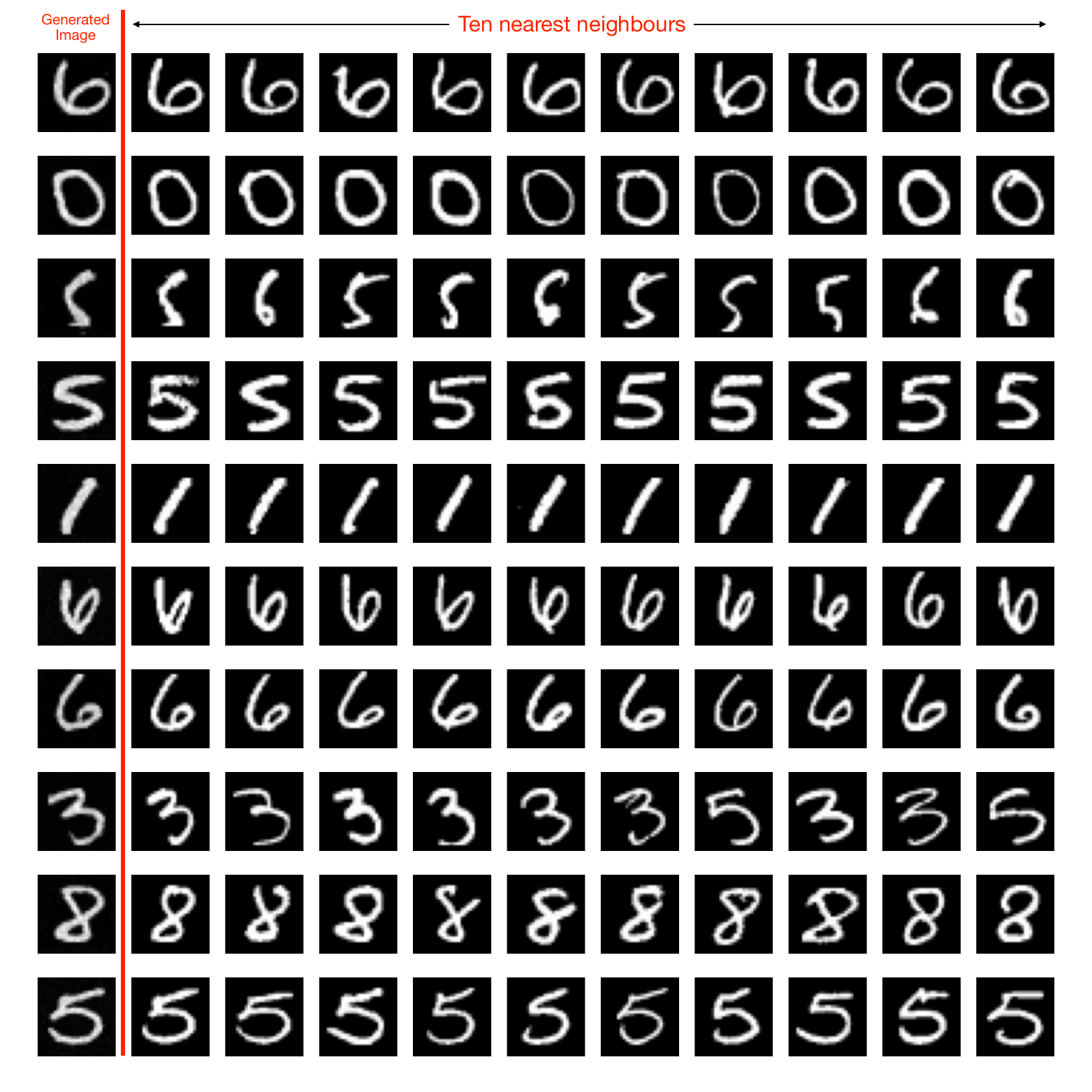}}
  \caption{$10$ nearest neighbours (calculated using Euclidean distance on InceptionV3 features and raw images respectively) from the MNIST dataset for samples generated with class averaged image passed through forward process as initialization using unconstrained multiplicative sampler \cref{eqn:gbmsde_backward_sampling}.}
\end{figure}

\begin{figure}[htbp]
  \centering
  \subfloat[InceptionV3 feature]{\includegraphics[page=16,width=0.49\textwidth] {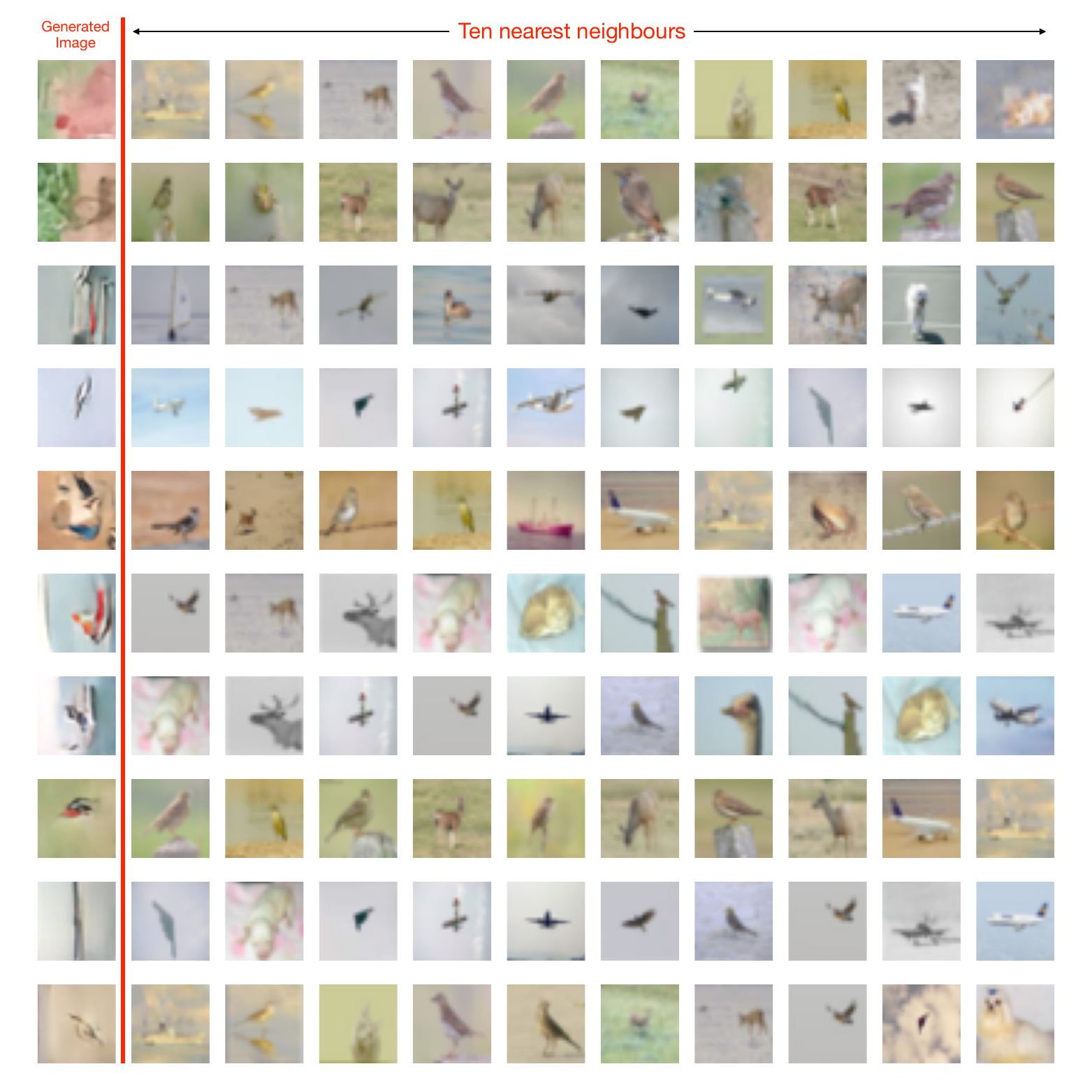}}
  \hfill
  \subfloat[Raw Image as feature]{\includegraphics[page=17,width=0.49\textwidth] {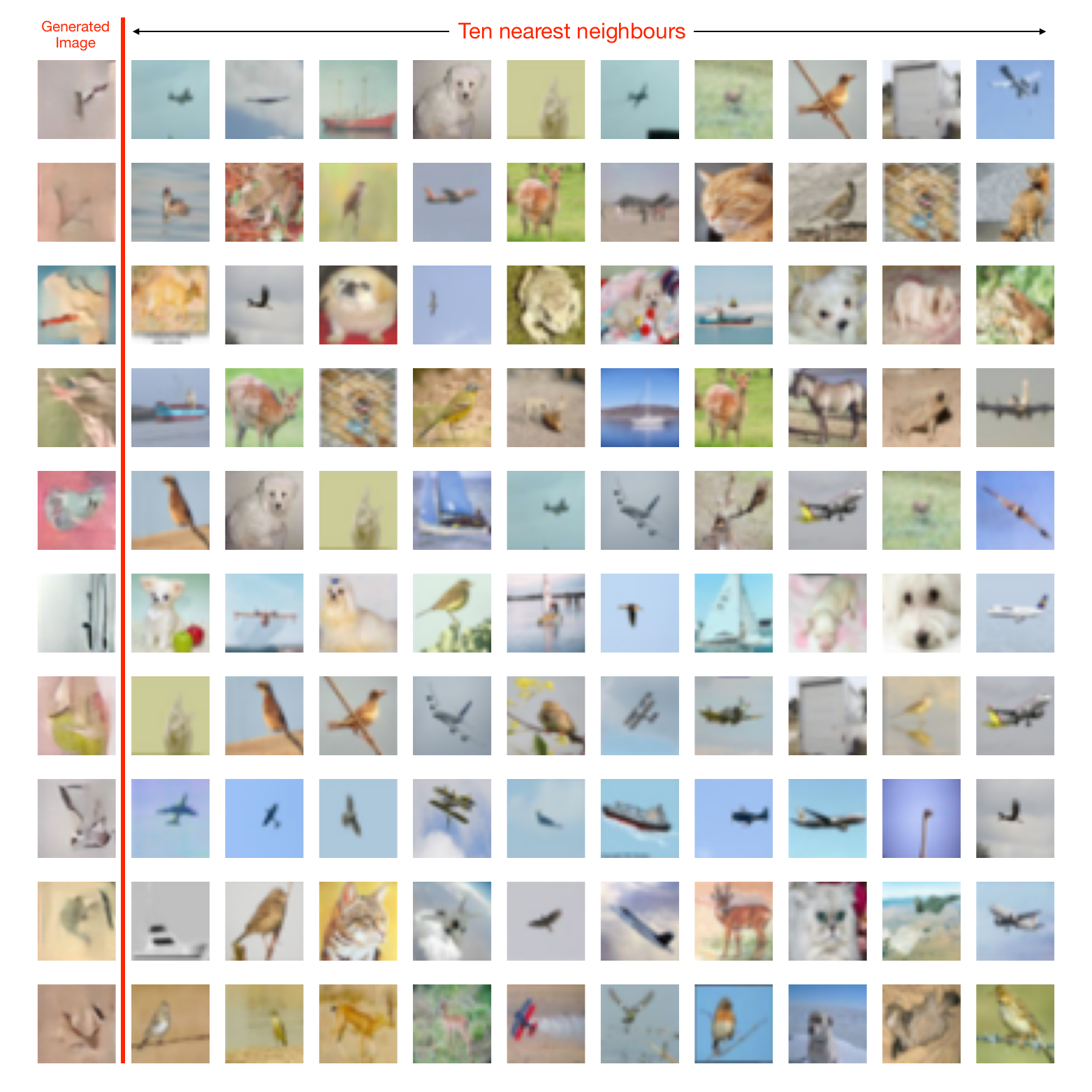}}

  \caption{$10$ nearest neighbours (calculated using Euclidean distance on InceptionV3 features and raw images respectively) from the CIFAR-10 dataset for samples generated with lognormal noise initialization using unconstrained multiplicative sampler \cref{eqn:gbmsde_backward_sampling} . }
\end{figure}

\begin{figure}[htbp]
  \centering
  \subfloat[InceptionV3 feature]{\includegraphics[page=1,width=0.49\textwidth] {figures/non_lamperti/new_non_lamperti.pdf}}
  \hfill
  \subfloat[Raw Image as feature]{\includegraphics[page=2,width=0.49\textwidth] {figures/non_lamperti/new_non_lamperti.pdf}}

  \caption{$10$ nearest neighbours (calculated using Euclidean distance on InceptionV3 features and raw images respectively) from the Fashion MNIST dataset for samples generated with lognormal noise initialization using unconstrained multiplicative sampler \cref{eqn:gbmsde_backward_sampling} . }
\end{figure}

\begin{figure}[htbp]
  \centering
  \subfloat[InceptionV3 feature]{\includegraphics[page=3,width=0.49\textwidth] {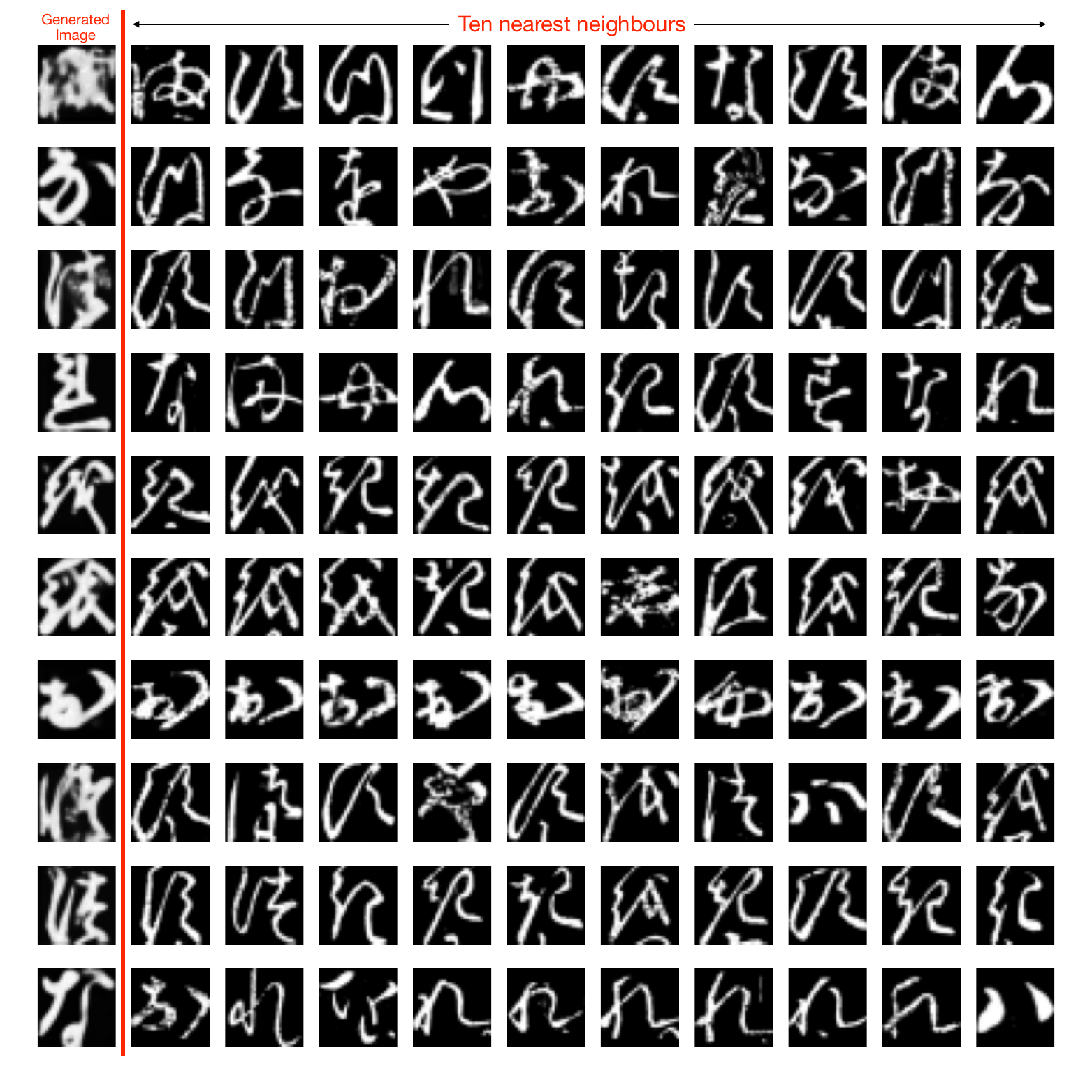}}
  \hfill
  \subfloat[Raw Image as feature]{\includegraphics[page=4,width=0.49\textwidth] {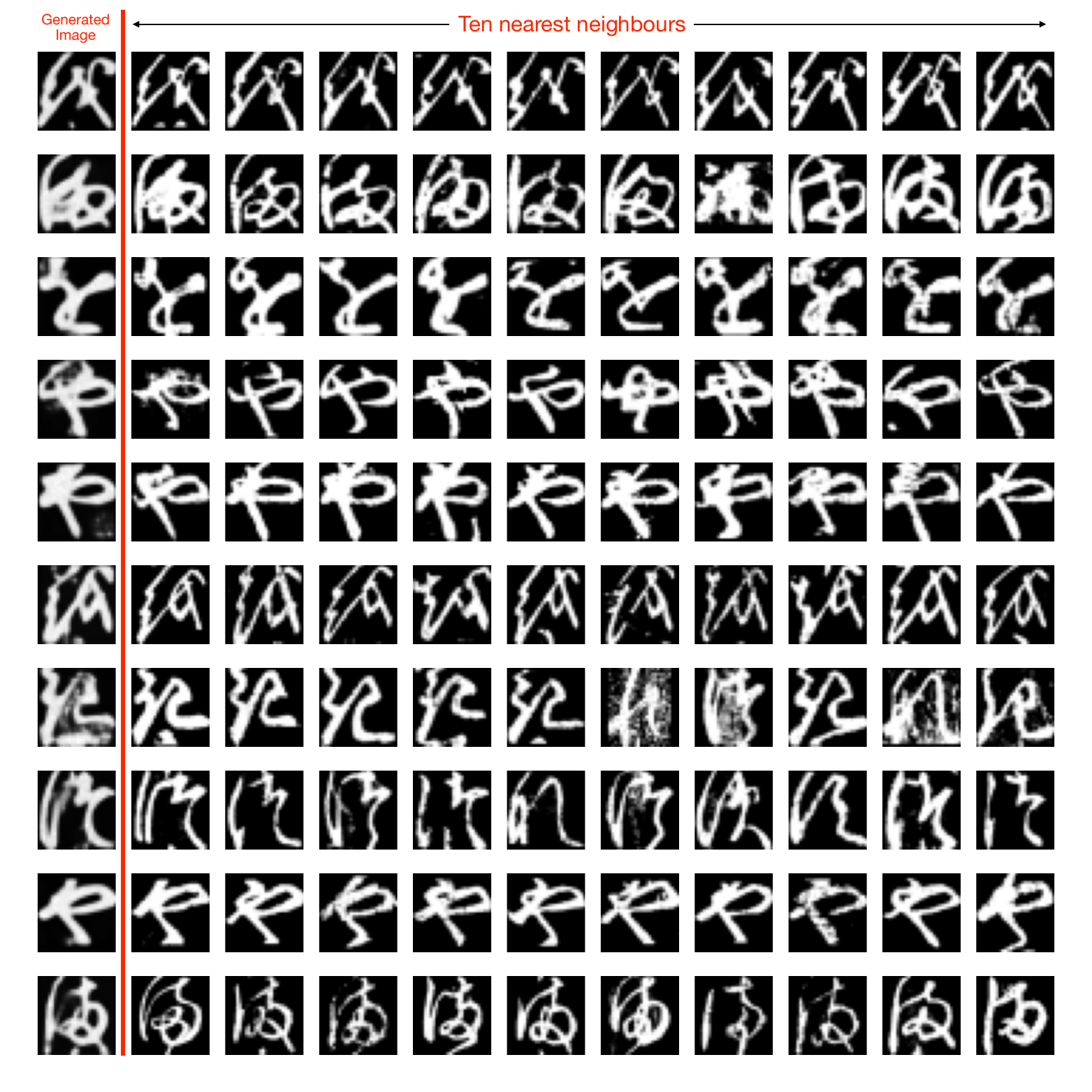}}
  \caption{$10$ nearest neighbours (calculated using Euclidean distance on InceptionV3 features and raw images respectively) from the Kuzushiji MNIST dataset for samples generated with lognormal noise initialization using unconstrained multiplicative sampler \cref{eqn:gbmsde_backward_sampling}.}
\end{figure}

\begin{figure}[htbp]
  \centering
  \subfloat[InceptionV3 feature]{\includegraphics[page=5,width=0.49\textwidth] {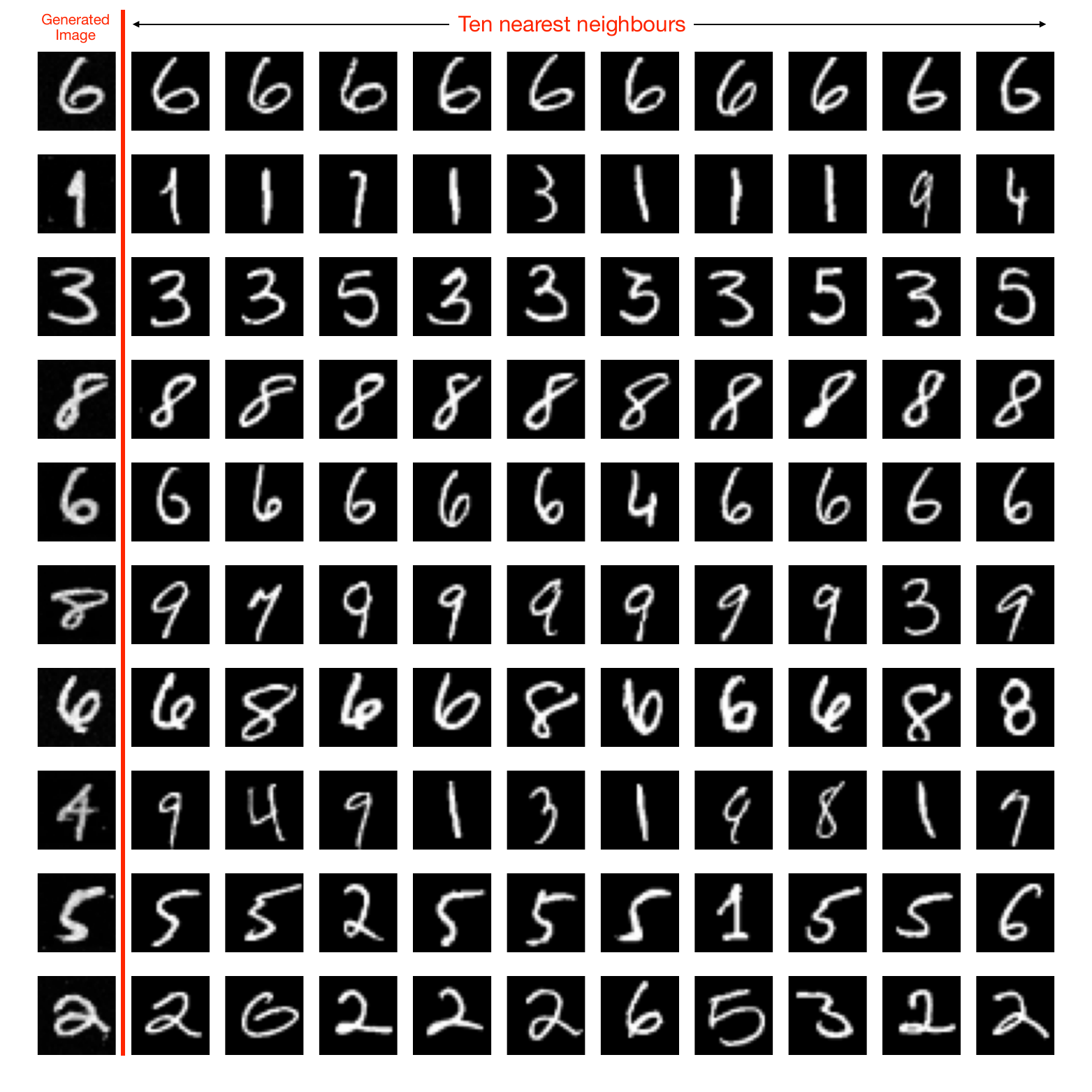}}
  \hfill
  \subfloat[Raw Image as feature]{\includegraphics[page=6,width=0.49\textwidth] {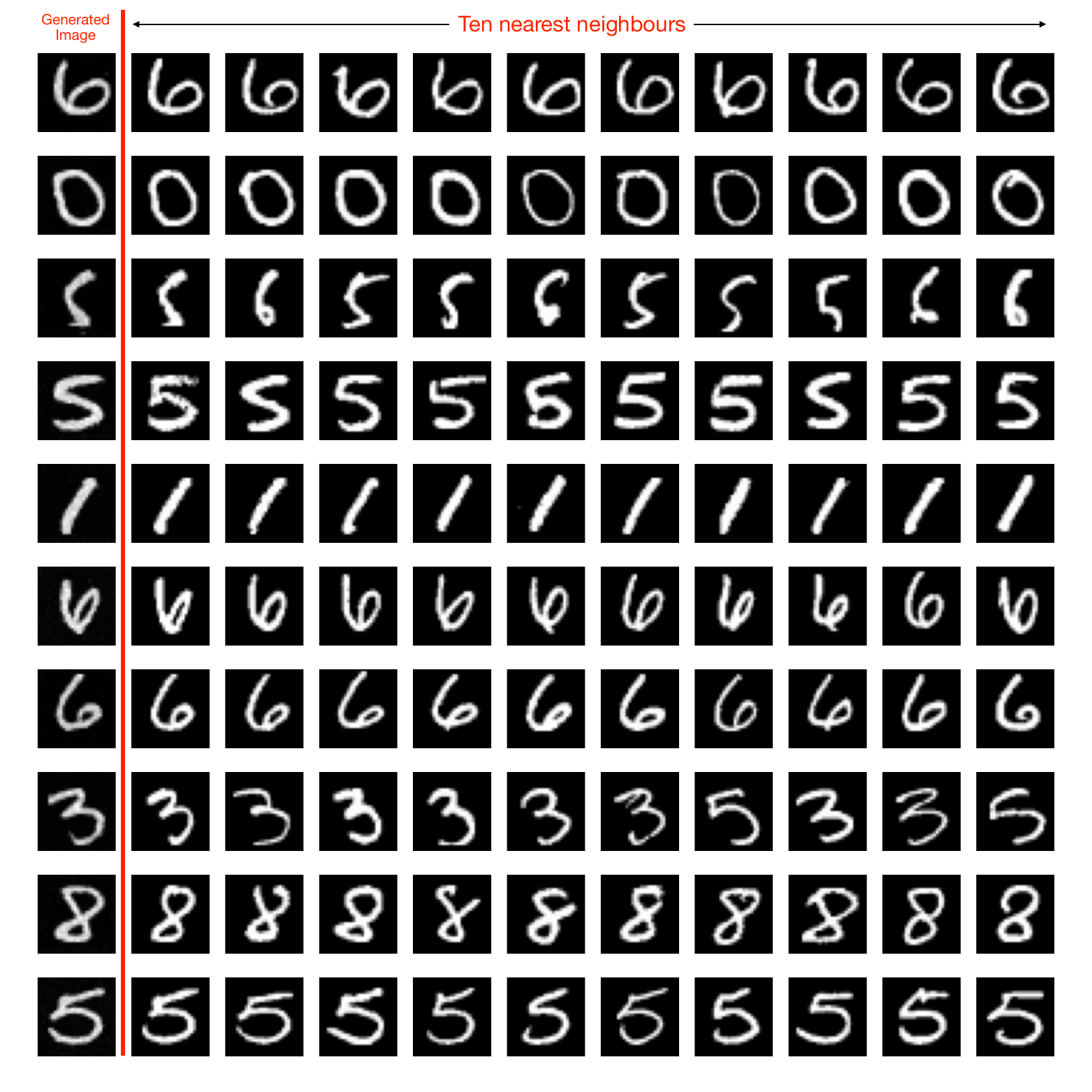}}
  \caption{$10$ nearest neighbours (calculated using Euclidean distance on InceptionV3 features and raw images respectively) from the MNIST dataset for samples generated with lognormal noise initialization using unconstrained multiplicative sampler~\cref{eqn:gbmsde_backward_sampling}. }
\end{figure}

\FloatBarrier

\subsection{Effects of annealing in generation of samples}

We investigate how the annealing factor $\chi$ influences the sampling dynamics across different initialization strategies. These strategies include initializing the reverse process by passing class-averaged images through the forward diffusion process, as well as a specific lognormal noise initialization. For the latter, we first estimate the $\mu$ and $\sigma$ parameters from the class-averaged images that have undergone the forward process, and then initialize the noise directly from this estimated parameters. To evaluate these approaches, we conduct an ablation study analyzing the interplay between the annealing factor and the sampler architecture while holding $L$ and the step-size $\delta$ constant. This impact is rigorously tested across the CIFAR-10, Fashion-MNIST, Kuzushiji MNIST, and MNIST datasets using factors of $\chi = 0.995, 0.9995,$ and $1.0$. We have observed a trade-off between sample fidelity and diversity with the parameter $\chi$, while $\chi = 1.0$ produces diverse samples with the cost of having noise, where as samples generated using $\chi = 0.995$ yield samples with mode collapse with no noise. samples generated using $\chi = 0.9995$ act as a middle ground as it generates samples with less noise and with less mode collapse.

\begin{figure}[htbp]
  \centering
  \begin{subfigure}[b]{0.49\textwidth}
    \includegraphics[width=\textwidth,height=0.45\textheight, keepaspectratio]{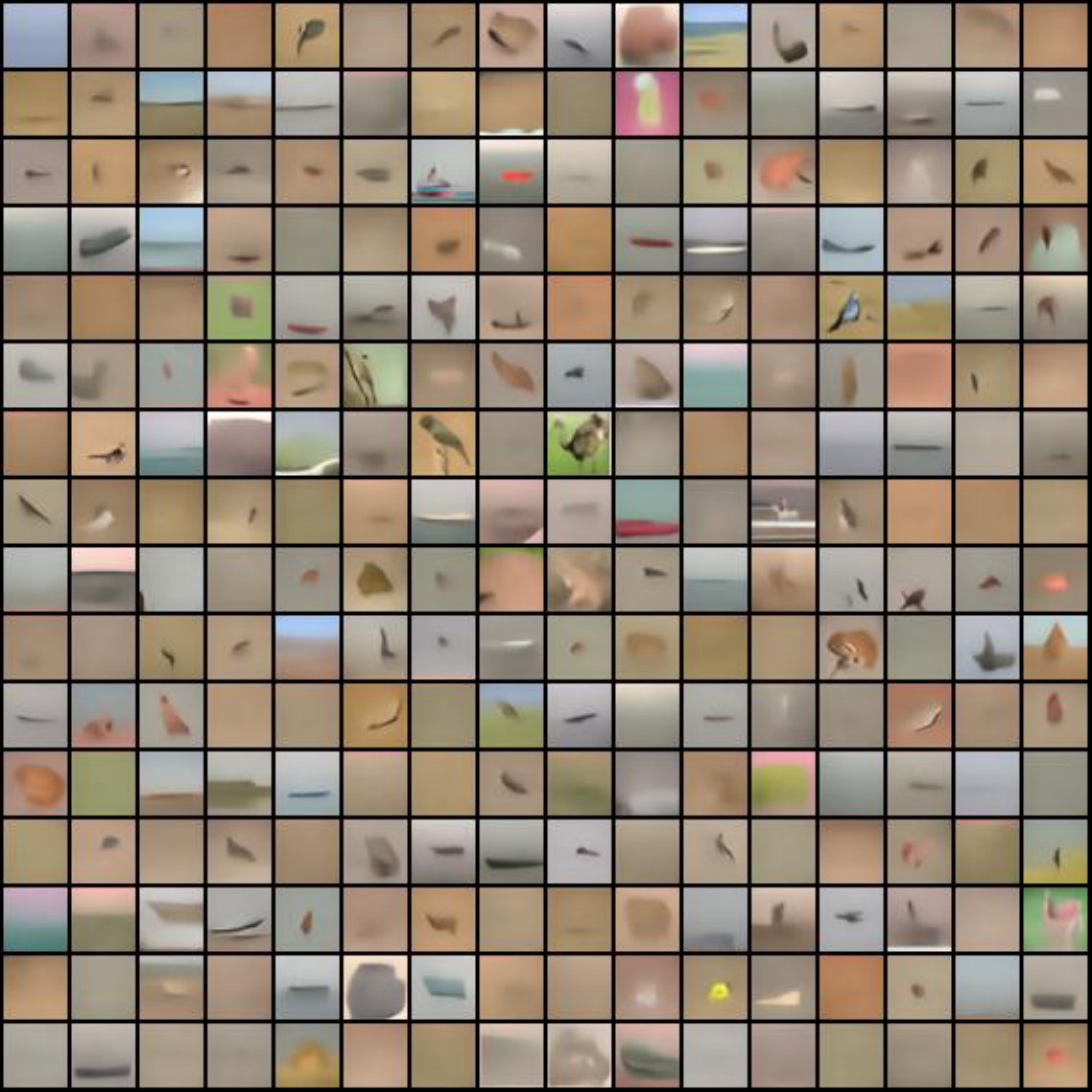}
    \caption{CIFAR-10 ($\chi = 0.995$)}
  \end{subfigure}
  \begin{subfigure}[b]{0.49\textwidth}
    \includegraphics[width=\textwidth,height=0.45\textheight, keepaspectratio]{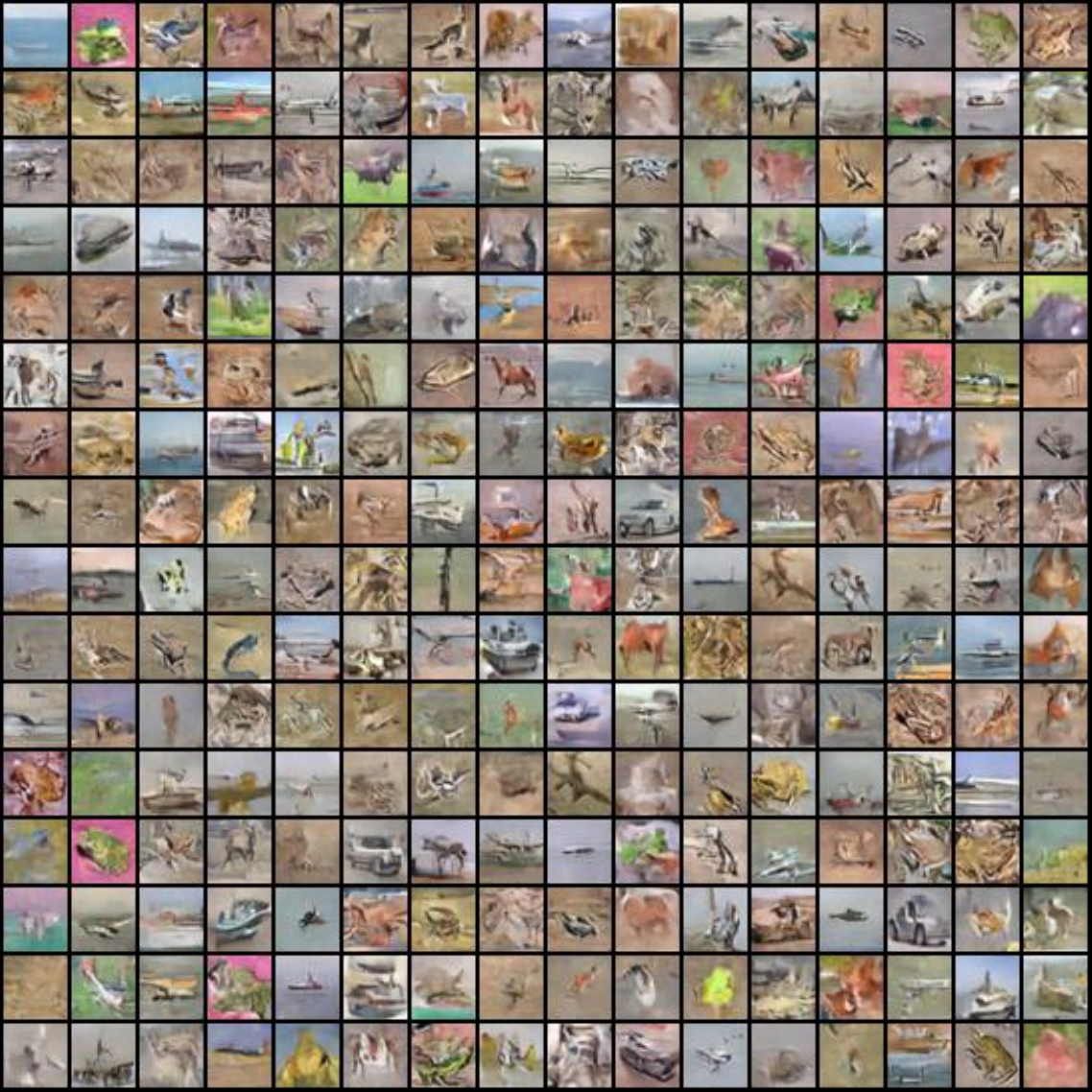}
    \caption{ CIFAR-10 ($\chi = 0.9995$)}
  \end{subfigure}
  \vspace{1.5em}
  \begin{subfigure}[b]{0.49\textwidth}
    \includegraphics[width=\textwidth,height=0.45\textheight, keepaspectratio]{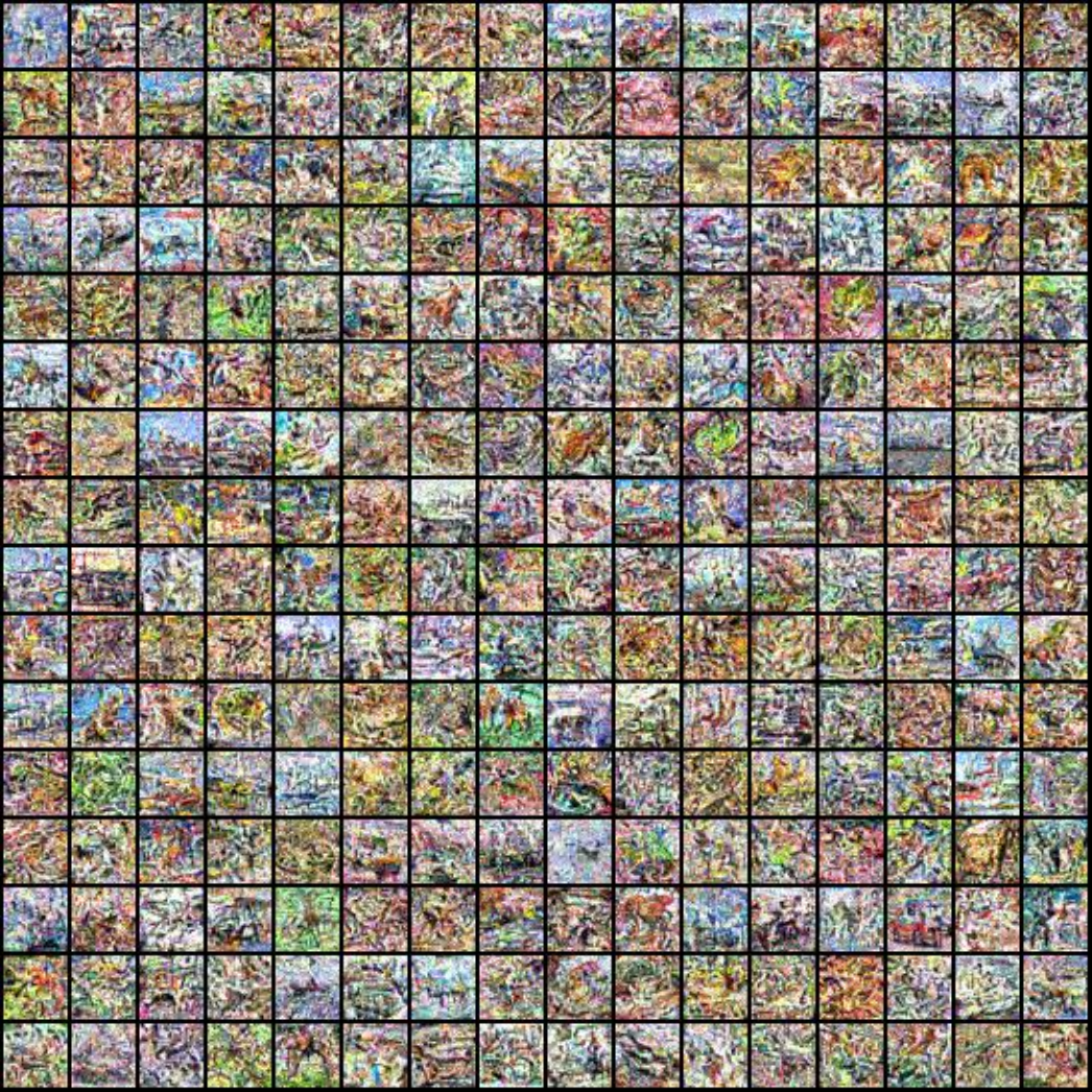}
    \caption{CIFAR-10 ($\chi = 1.0$)}
  \end{subfigure}

  \caption{Uncurated samples generated by the Dale-Langevin sampler \eqref{eqn:gbmsde_sampling_a} using the Annealed algorithm \ref{algo:gbm_sampler_anneal} on the CIFAR-10 dataset. Results are presented for annealing factors $\chi \in \{0.995, 0.9995, 1.0\}$. Initialized from the terminal state of the forward process applied to class-averaged images, the sampling follows the configurations in \cref{tab:parameters_sampling}.}
\end{figure}

\begin{figure}[htbp]
  \centering
  \begin{subfigure}[b]{0.49\textwidth}
    \includegraphics[width=\textwidth,height=0.45\textheight, keepaspectratio]{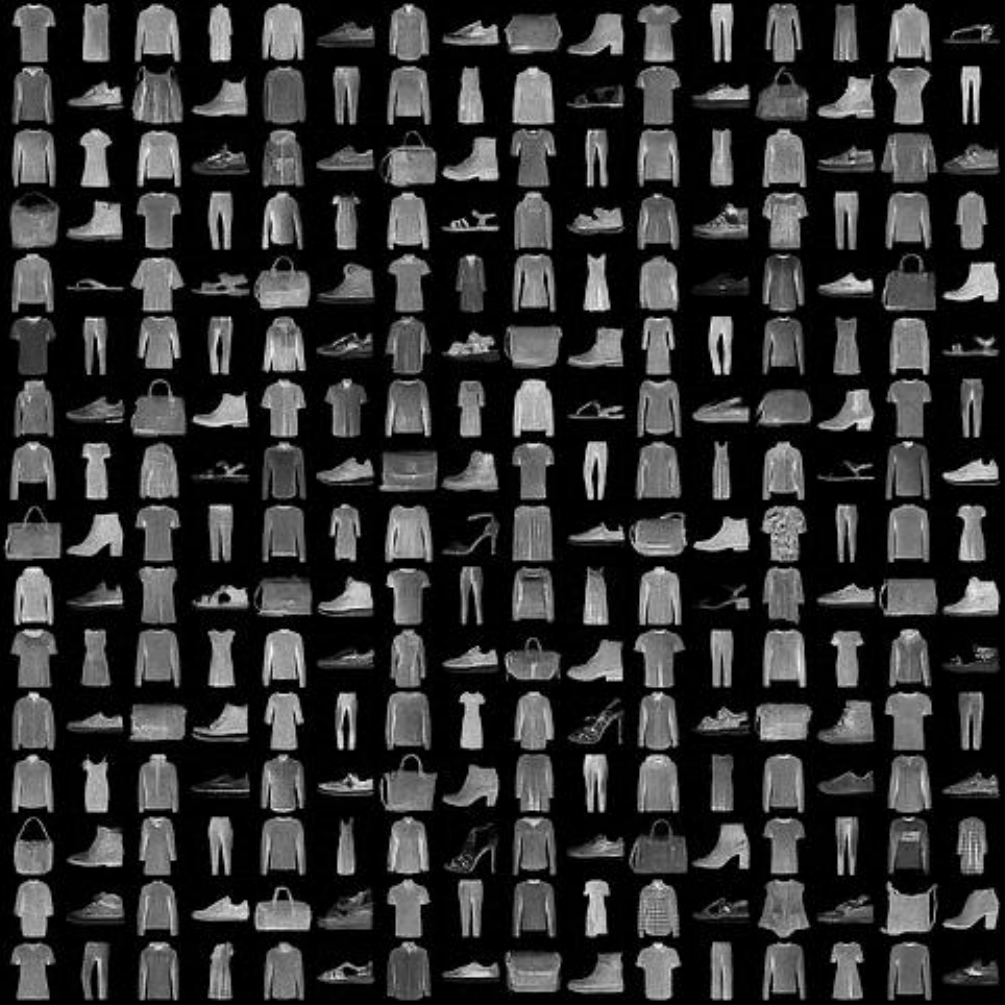}
    \caption{Fashion-MNIST}
  \end{subfigure}
  \begin{subfigure}[b]{0.49\textwidth}
    \includegraphics[width=\textwidth,height=0.45\textheight, keepaspectratio]{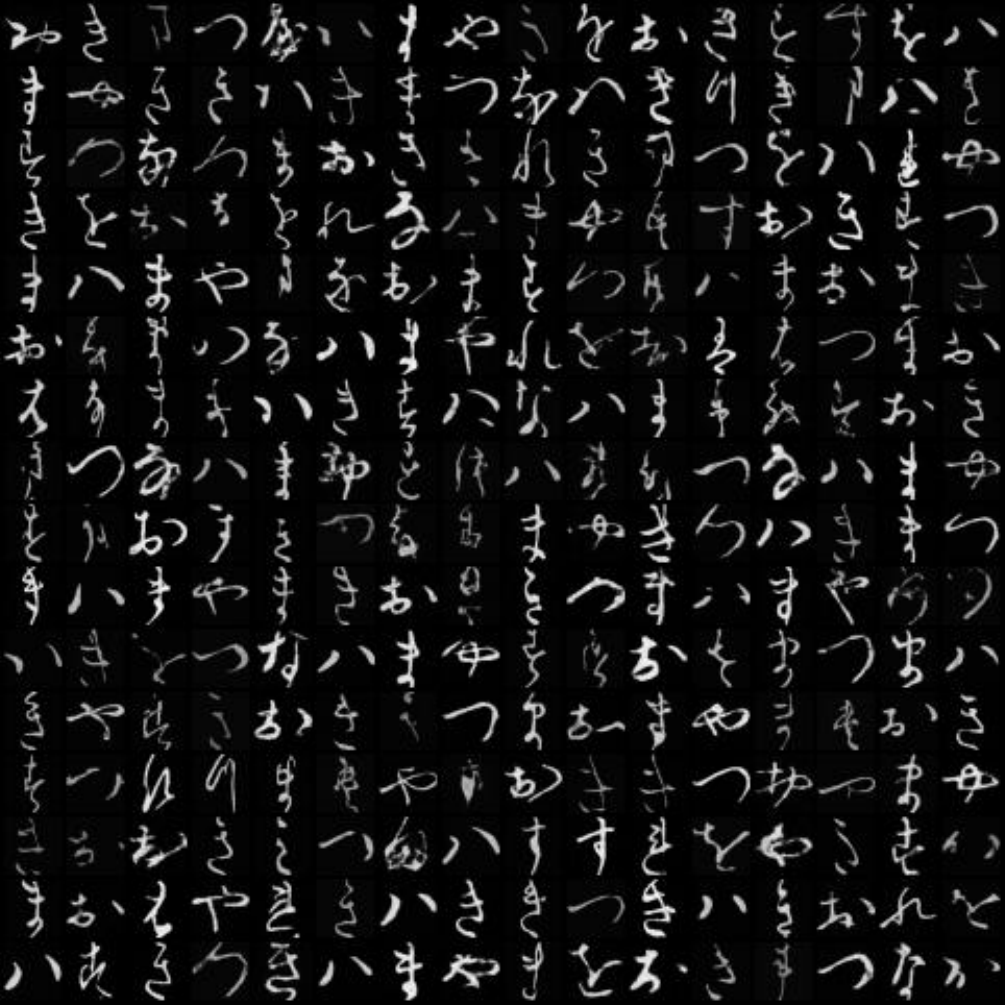}
    \caption{Kuzushiji MNIST}
  \end{subfigure}
  \vspace{1.5em}
  \begin{subfigure}[b]{0.49\textwidth}
    \includegraphics[width=\textwidth,height=0.45\textheight, keepaspectratio]{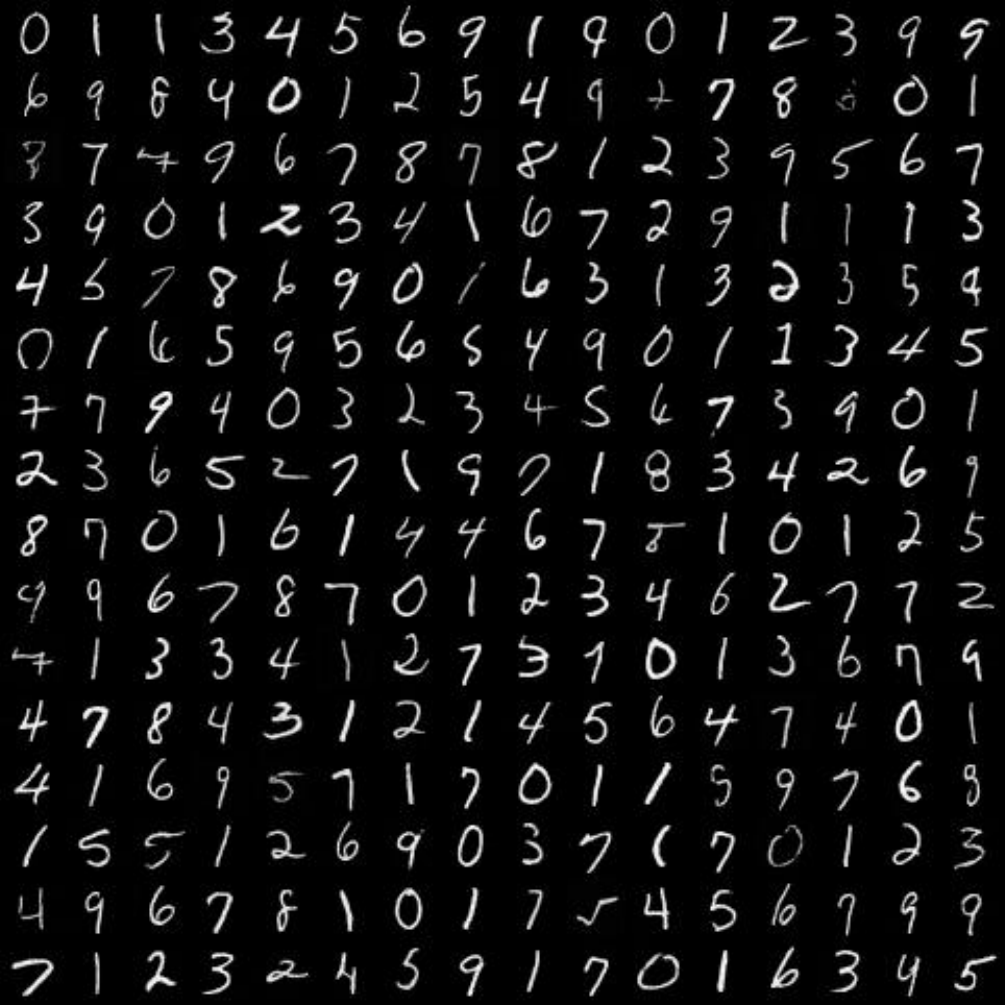}
    \caption{MNIST}
  \end{subfigure}

  \caption{Uncurated samples produced by the Dale-Langevin sampler \eqref{eqn:gbmsde_sampling_a} using the Annealed Dale-Langevin algorithm \ref{algo:gbm_sampler_anneal} with $\chi = 0.995$. Results for Fashion-MNIST, Kuzushiji MNIST, and MNIST were initialized by applying the forward process to class-averaged images, as detailed in the hyperparameters of \cref{tab:parameters_sampling}.}
\end{figure}

\begin{figure}[htbp]
  \centering
  \begin{subfigure}[b]{0.49\textwidth}
    \includegraphics[width=\textwidth,height=0.45\textheight, keepaspectratio]{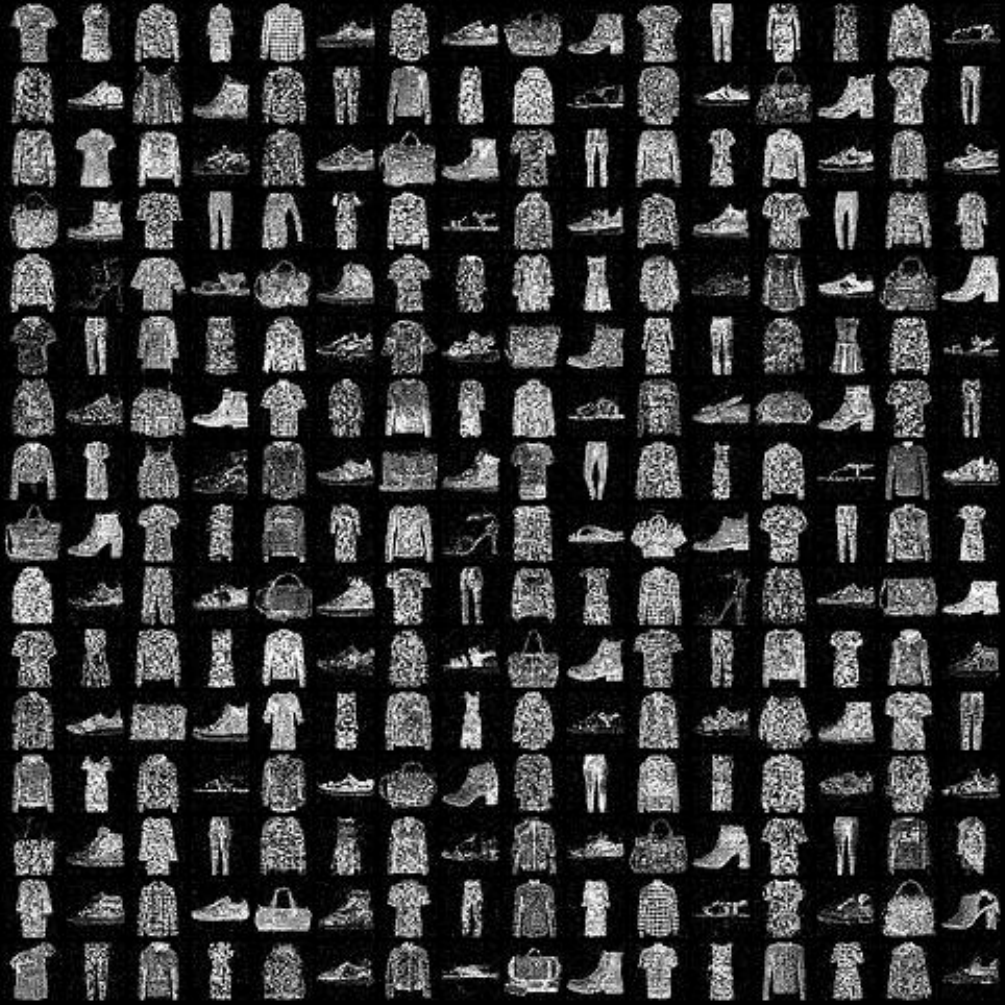}
    \caption{Fashion-MNIST}
  \end{subfigure}
  \begin{subfigure}[b]{0.49\textwidth}
    \includegraphics[width=\textwidth,height=0.45\textheight, keepaspectratio]{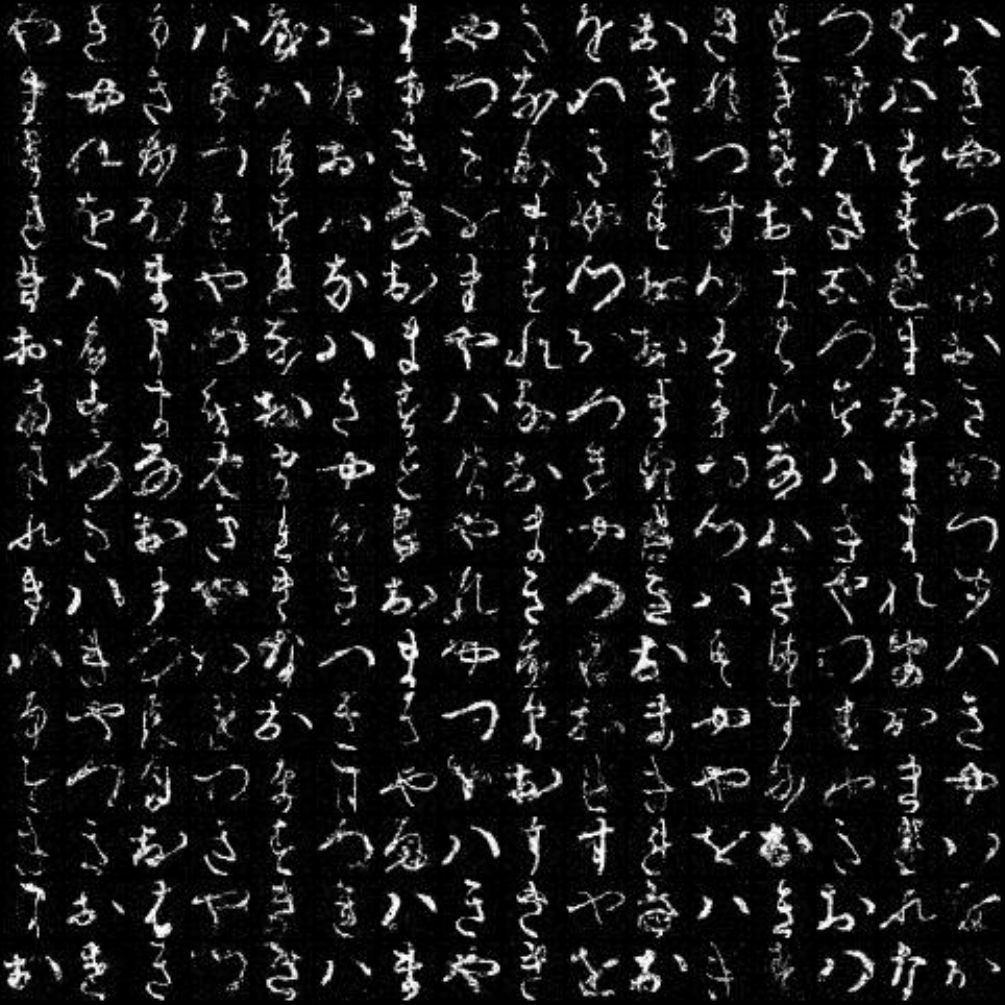}
    \caption{Kuzushiji MNIST}
  \end{subfigure}
  \vspace{1.5em}
  \begin{subfigure}[b]{0.49\textwidth}
    \includegraphics[width=\textwidth,height=0.45\textheight, keepaspectratio]{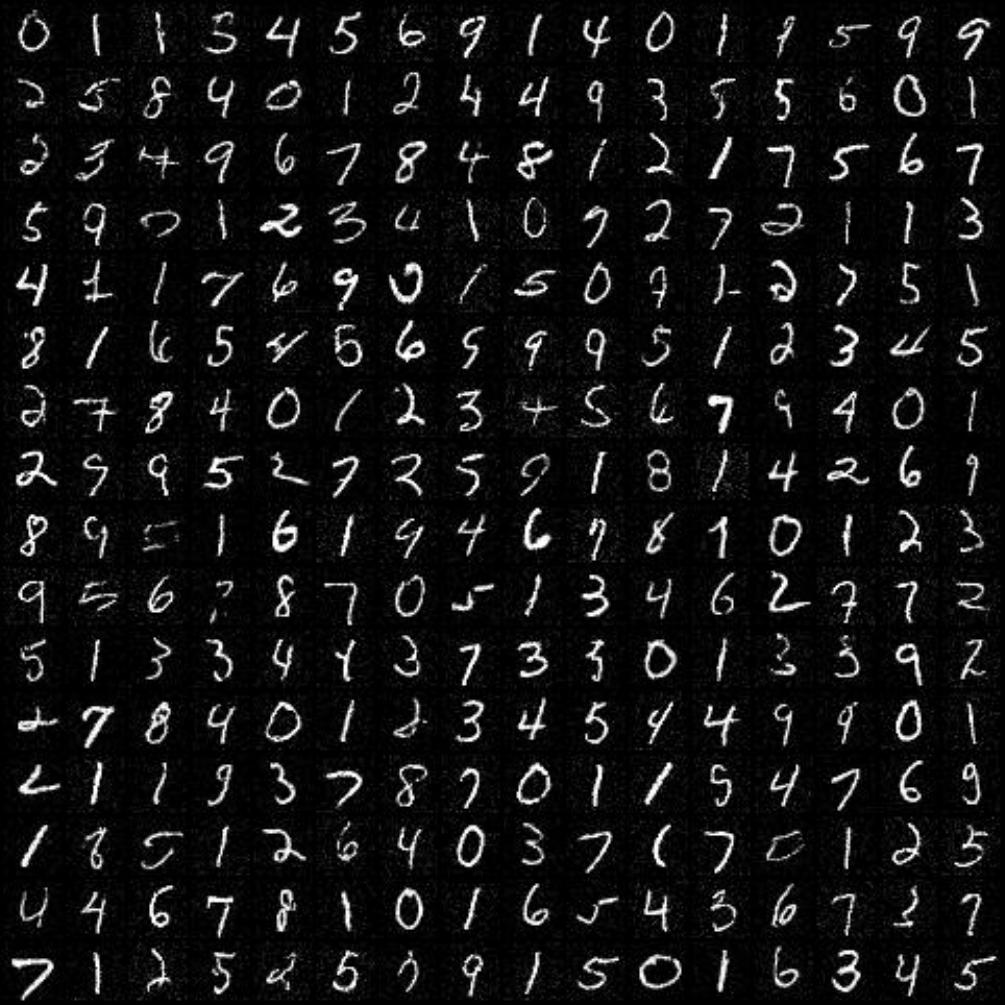}
    \caption{MNIST}
  \end{subfigure}

  \caption{Uncurated samples produced by the Dale-Langevin sampler \eqref{eqn:gbmsde_sampling_a} using the Annealed Dale-Langevin algorithm \ref{algo:gbm_sampler_anneal} with $\chi = 0.9995$. Results for Fashion-MNIST, Kuzushiji MNIST, and MNIST were initialized by applying the forward process to class-averaged images, as detailed in the hyperparameters of \cref{tab:parameters_sampling}.}
\end{figure}

\begin{figure}[htbp]
  \centering
  \begin{subfigure}[b]{0.49\textwidth}
    \includegraphics[width=\textwidth,height=0.45\textheight, keepaspectratio]{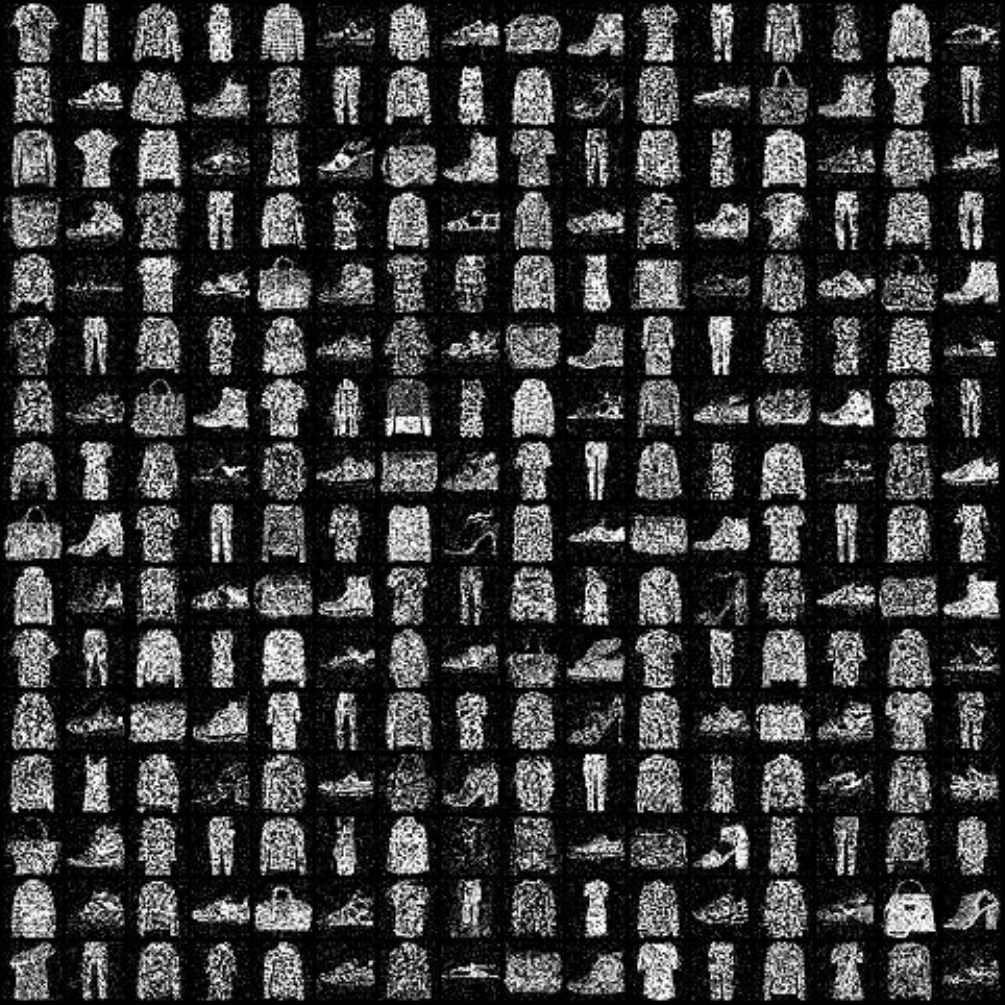}
    \caption{Fashion-MNIST}
  \end{subfigure}
  \begin{subfigure}[b]{0.49\textwidth}
    \includegraphics[width=\textwidth,height=0.45\textheight, keepaspectratio]{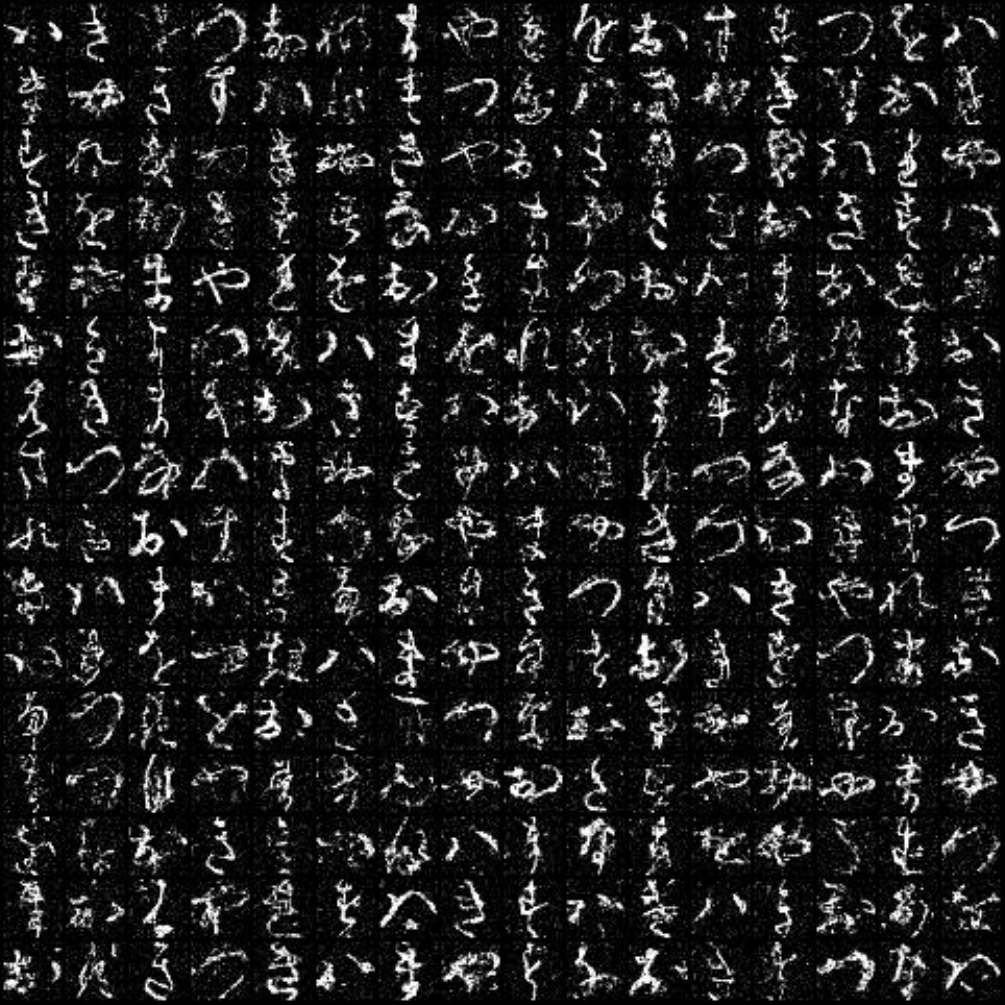}
    \caption{Kuzushiji MNIST}
  \end{subfigure}
   \vspace{1.5em}
  \begin{subfigure}[b]{0.49\textwidth}
    \includegraphics[width=\textwidth,height=0.45\textheight, keepaspectratio]{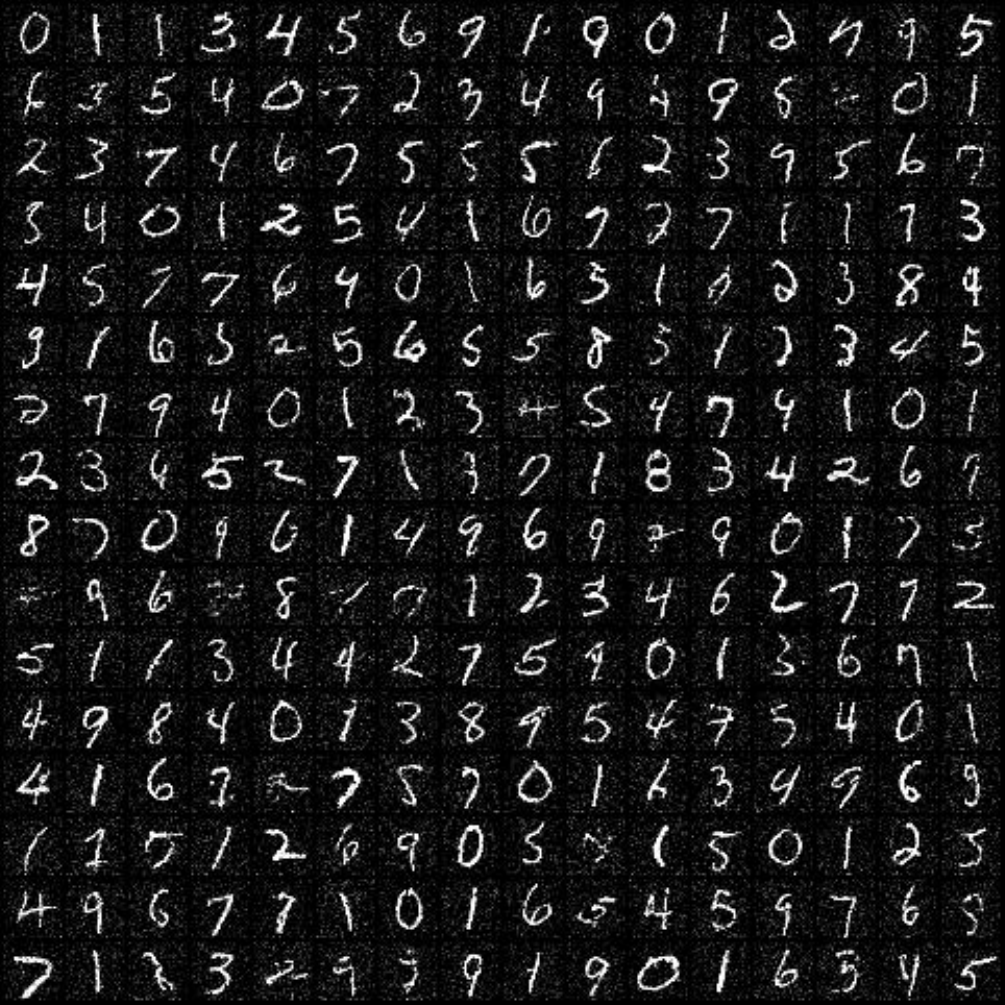}
    \caption{MNIST}
  \end{subfigure}

  \caption{Uncurated samples produced by the Dale-Langevin sampler \eqref{eqn:gbmsde_sampling_a} using the Annealed Dale-Langevin algorithm \ref{algo:gbm_sampler_anneal} with $\chi = 1.0$. Results for Fashion-MNIST, Kuzushiji MNIST, and MNIST were initialized by applying the forward process to class-averaged images, as detailed in the hyperparameters of \cref{tab:parameters_sampling}.}
\end{figure}

\begin{figure}[htbp]
  \centering
  \begin{subfigure}[b]{0.49\textwidth}
    \includegraphics[width=\textwidth,height=0.45\textheight, keepaspectratio]{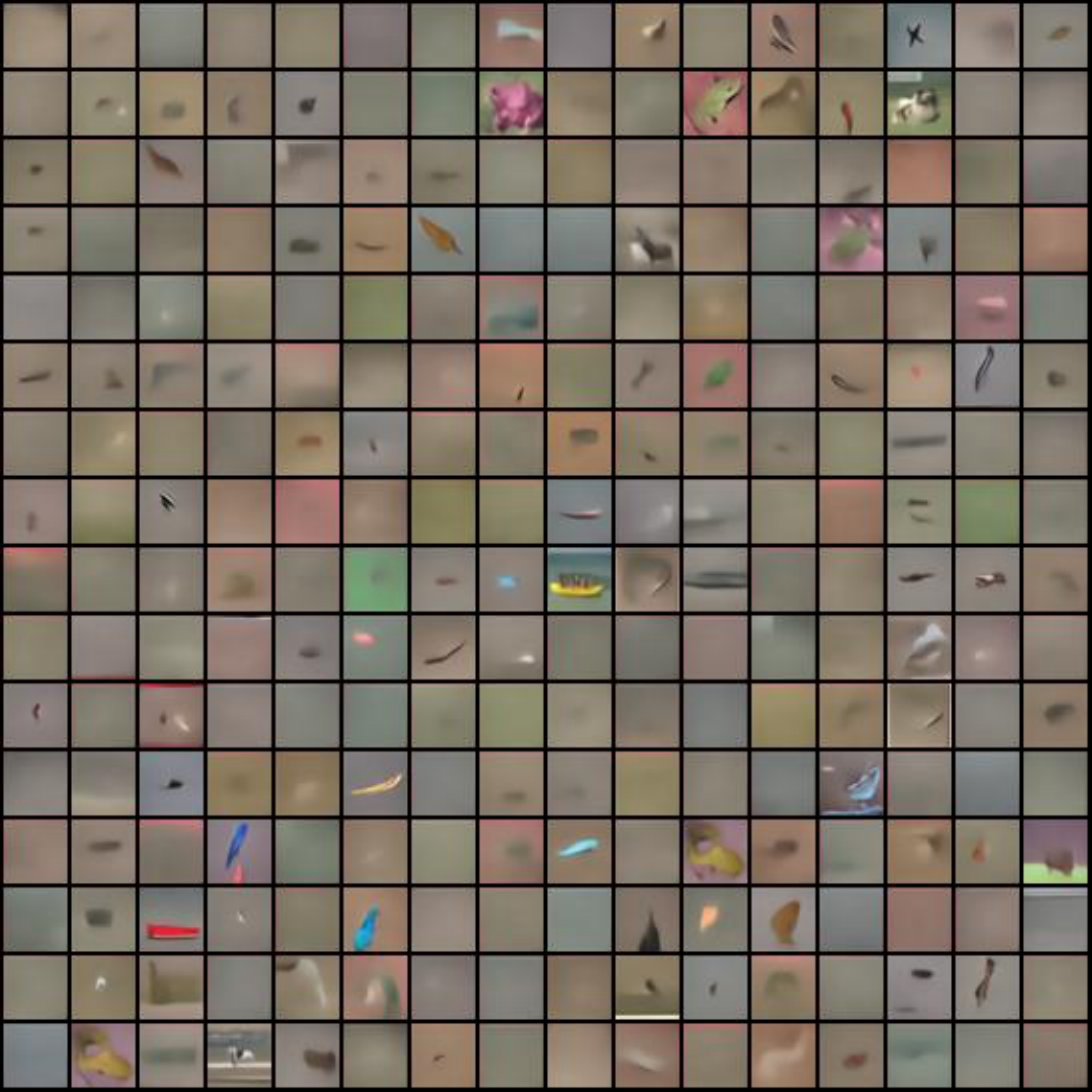}
    \caption{CIFAR-10 ($\chi = 0.995$)}
  \end{subfigure}
  \begin{subfigure}[b]{0.49\textwidth}
    \includegraphics[width=\textwidth,height=0.45\textheight, keepaspectratio]{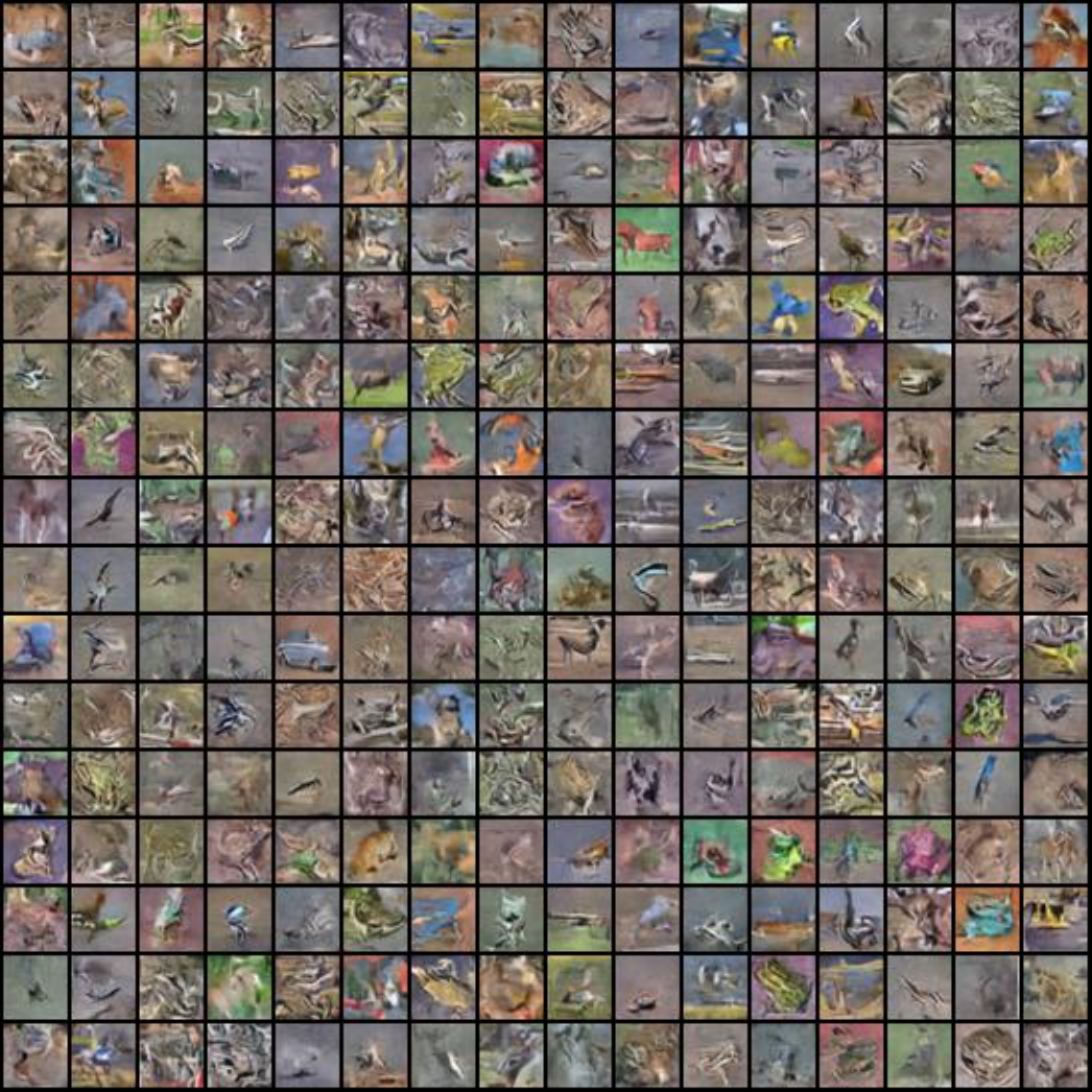}
    \caption{CIFAR-10 ($\chi = 0.9995$)}
  \end{subfigure}
   \vspace{1.5em}
  \begin{subfigure}[b]{0.49\textwidth}
    \includegraphics[width=\textwidth,height=0.45\textheight, keepaspectratio]{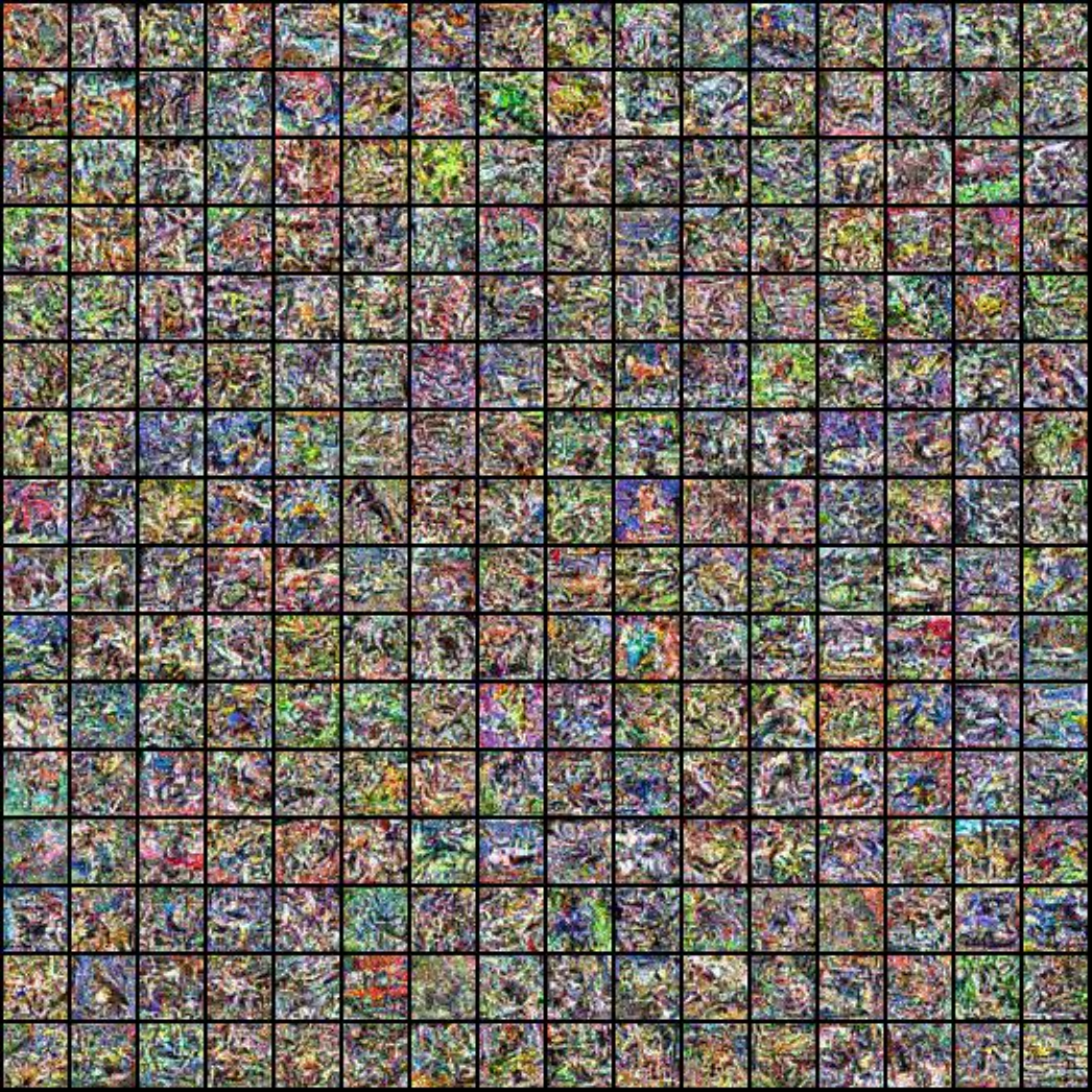}
    \caption{CIFAR-10 ($\chi = 1.0$)}
  \end{subfigure}

  \caption{Uncurated samples generated by the Dale-Langevin sampler \eqref{eqn:gbmsde_sampling_a} following the Annealed algorithm \ref{algo:gbm_sampler_anneal} for the CIFAR-10 dataset. Results are shown for various annealing factors, specifically $\chi \in \{0.995, 0.9995, 1.0\}$. Initialized with lognormal noise, the sampling process follows the configurations ($L,\delta$) specified in \cref{tab:parameters_sampling}.}
\end{figure}

\begin{figure}[htbp]
  \centering
  \begin{subfigure}[b]{0.49\textwidth}
    \includegraphics[width=\textwidth,height=0.45\textheight, keepaspectratio]{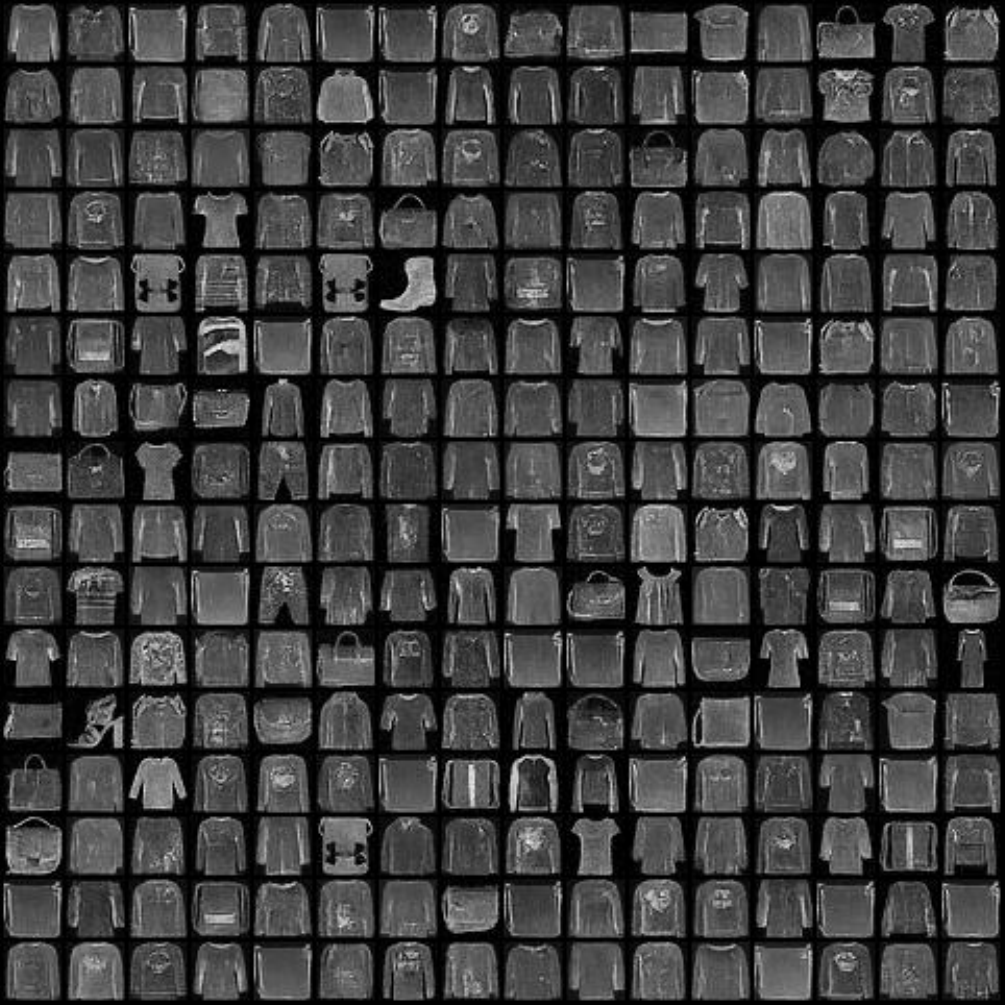}
    \caption{Fashion-MNIST}
  \end{subfigure}
  \begin{subfigure}[b]{0.49\textwidth}
    \includegraphics[width=\textwidth,height=0.45\textheight, keepaspectratio]{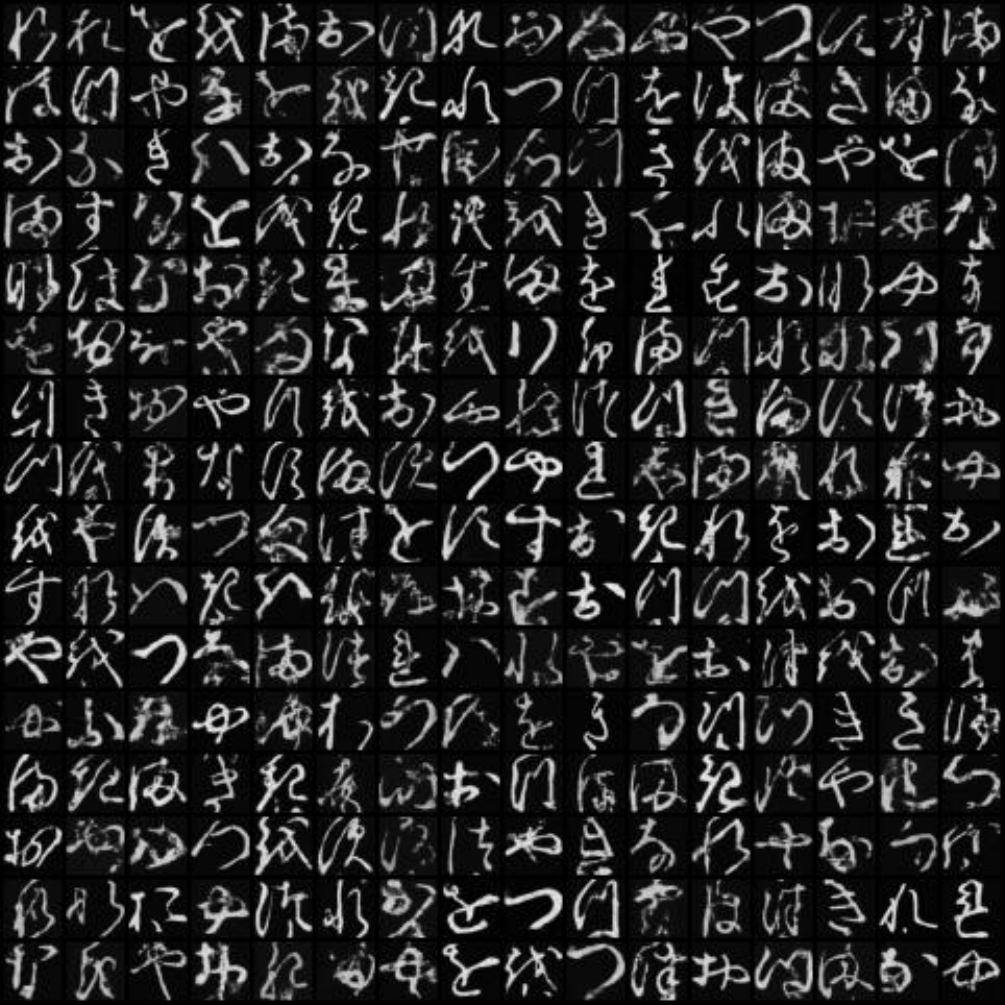}
    \caption{Kuzushiji MNIST}
  \end{subfigure}
  \vspace{1.5em}
  \begin{subfigure}[b]{0.49\textwidth}
    \includegraphics[width=\textwidth,height=0.45\textheight, keepaspectratio]{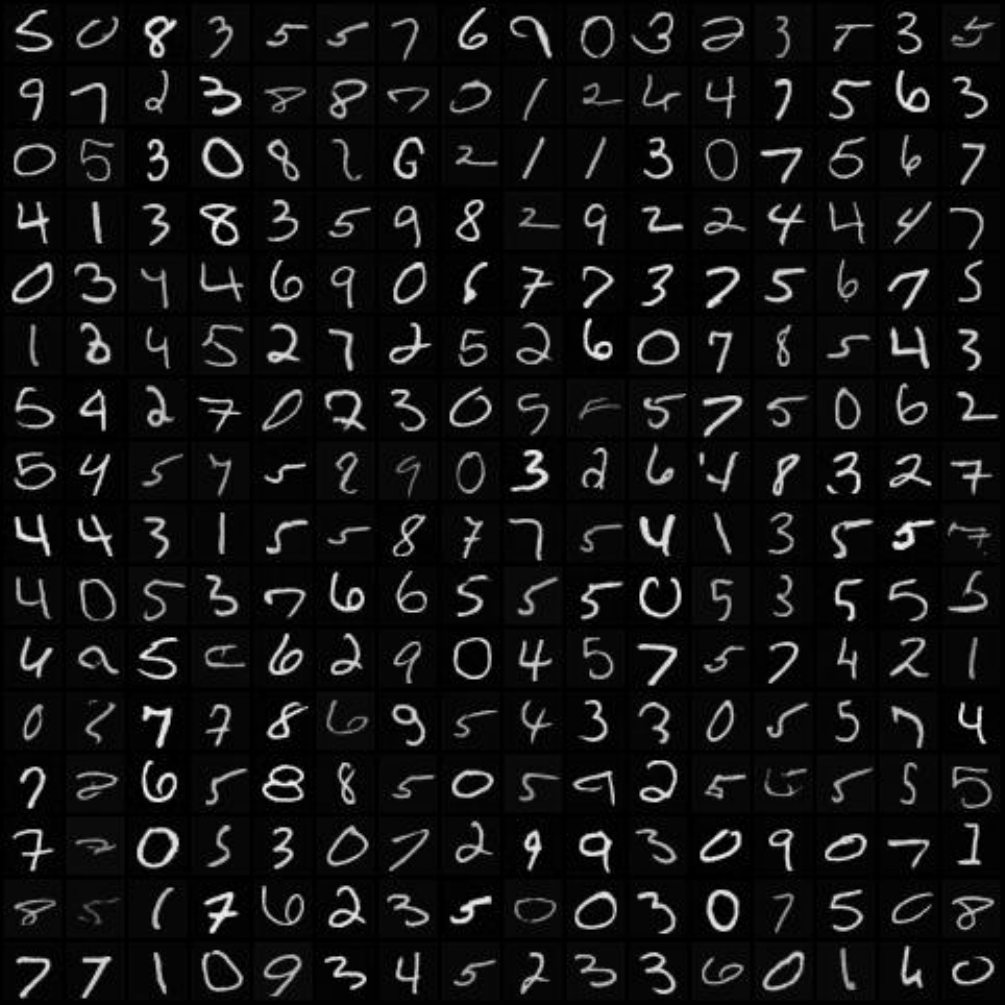}
    \caption{MNIST}
  \end{subfigure}

  \caption{Uncurated samples generated by the Dale-Langevin sampler \eqref{eqn:gbmsde_sampling_a} following the Annealed algorithm \ref{algo:gbm_sampler_anneal} with an annealing factor $\chi = 0.995$. Results are presented for Fashion-MNIST, Kuzushiji MNIST, and MNIST, respectively. Initialized with lognormal noise, the sampling follows the configurations ($L,\delta$) specified in \cref{tab:parameters_sampling}.}
\end{figure}

\begin{figure}[htbp]
  \centering
  \begin{subfigure}[b]{0.49\textwidth}
    \includegraphics[width=\textwidth,height=0.45\textheight, keepaspectratio]{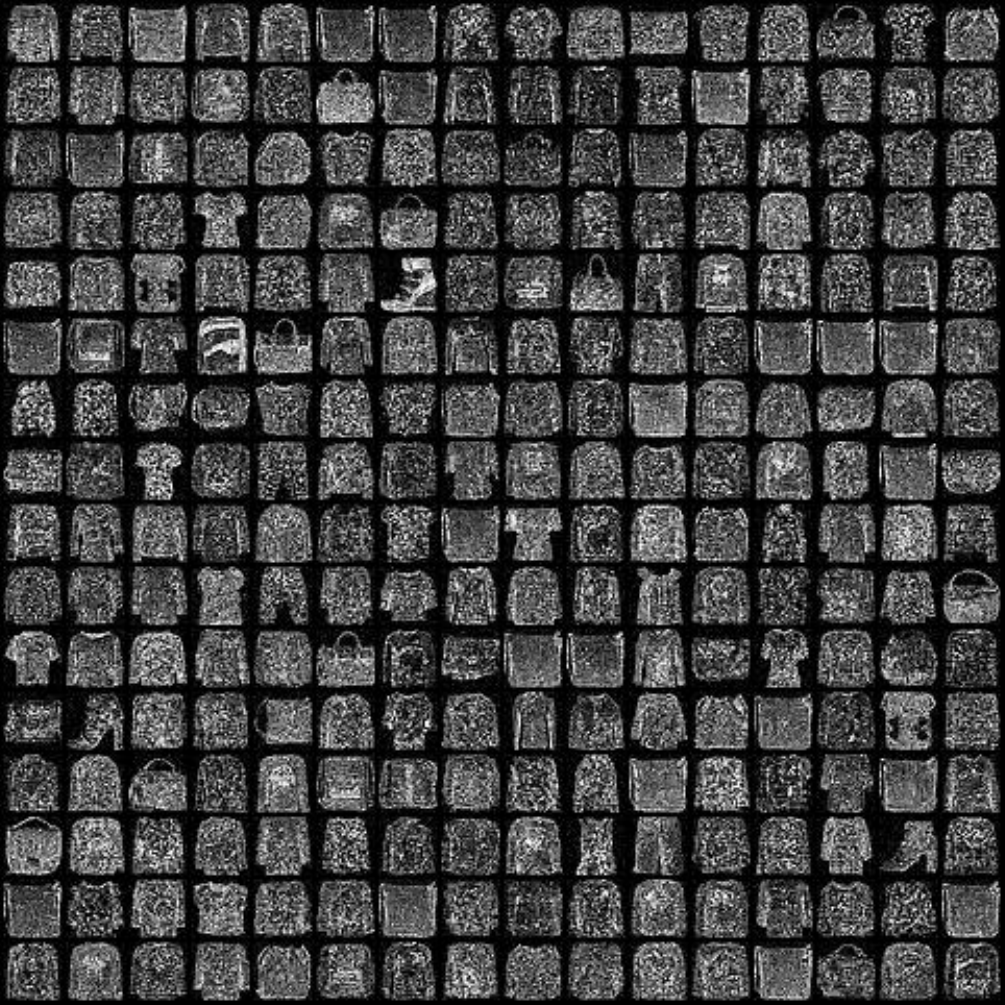}
    \caption{Fashion-MNIST}
  \end{subfigure}
  \begin{subfigure}[b]{0.49\textwidth}
    \includegraphics[width=\textwidth,height=0.45\textheight, keepaspectratio]{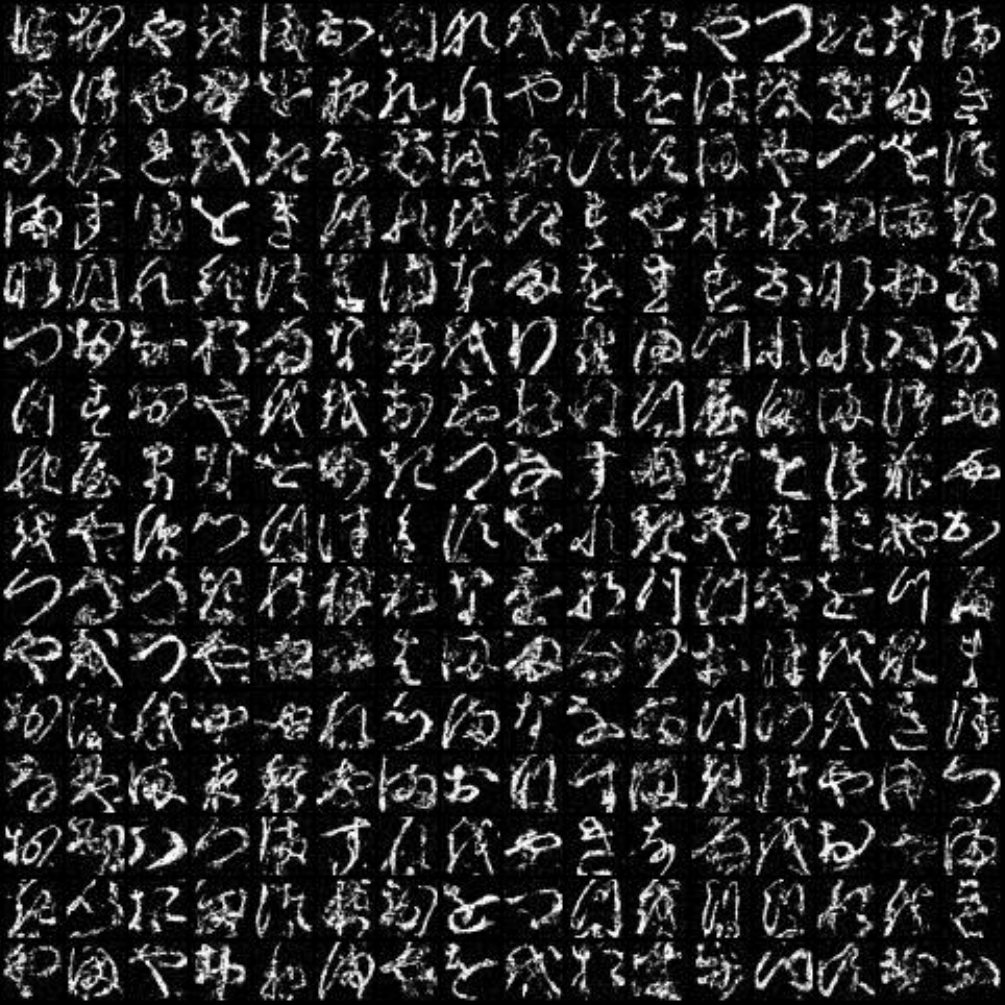}
    \caption{Kuzushiji MNIST}
  \end{subfigure}
  \vspace{1.5em}
  \begin{subfigure}[b]{0.49\textwidth}
    \includegraphics[width=\textwidth,height=0.45\textheight, keepaspectratio]{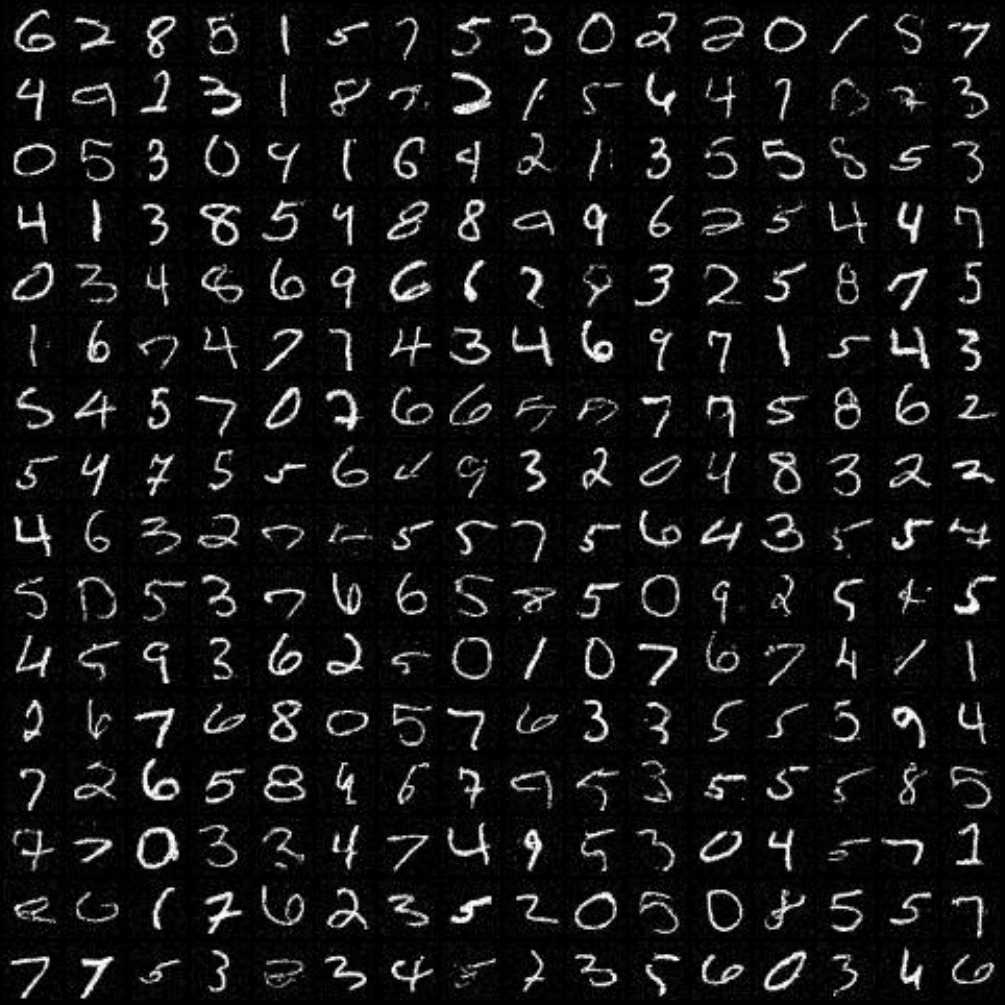}
    \caption{MNIST}
  \end{subfigure}
  \hfill
  \caption{Uncurated samples generated by the Dale-Langevin sampler \eqref{eqn:gbmsde_sampling_a} following the Annealed algorithm \ref{algo:gbm_sampler_anneal} with an annealing factor $\chi = 0.9995$. Results are presented for Fashion-MNIST, Kuzushiji MNIST, and MNIST, respectively. Initialized with lognormal noise, the sampling follows the configurations ($L,\delta$) specified in \cref{tab:parameters_sampling}.}
\end{figure}

\begin{figure}[htbp]
  \centering
  \begin{subfigure}[b]{0.49\textwidth}
    \includegraphics[width=\textwidth,height=0.45\textheight, keepaspectratio]{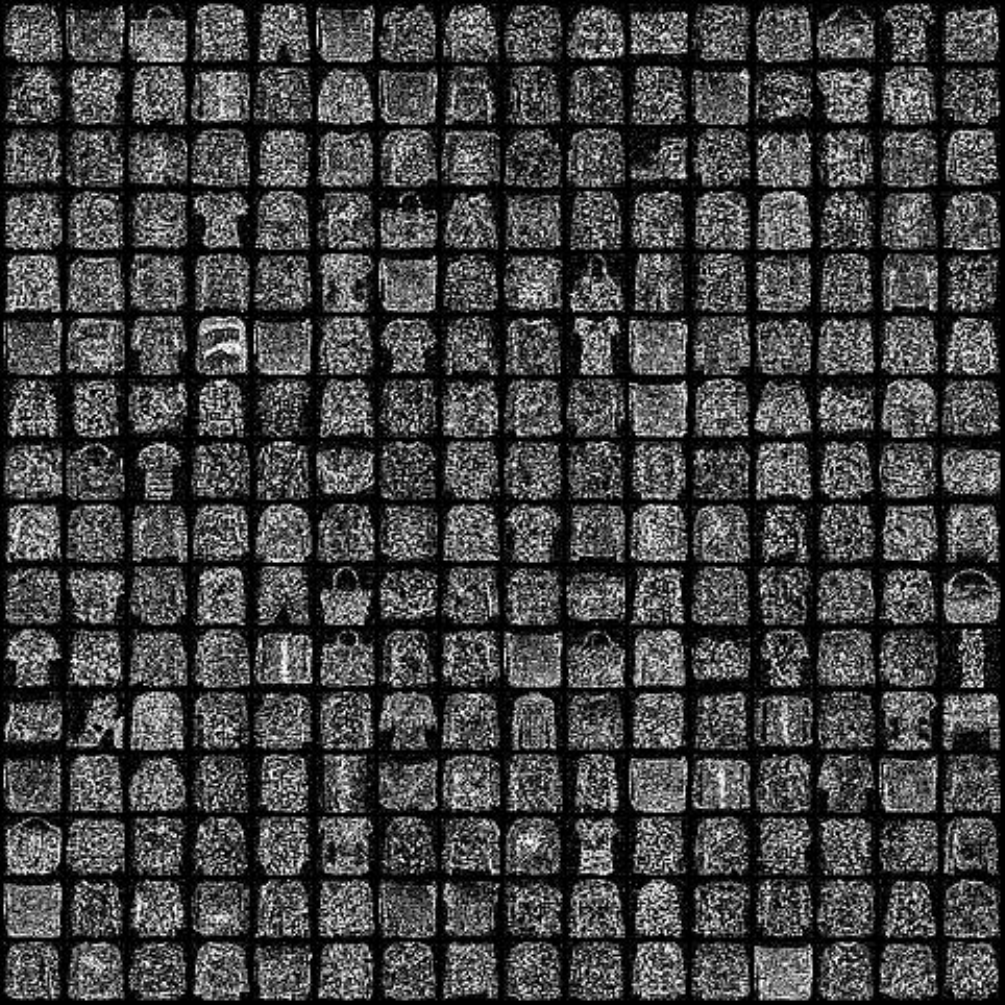}
    \caption{Fashion-MNIST}
  \end{subfigure}
  \begin{subfigure}[b]{0.49\textwidth}
    \includegraphics[width=\textwidth,height=0.45\textheight, keepaspectratio]{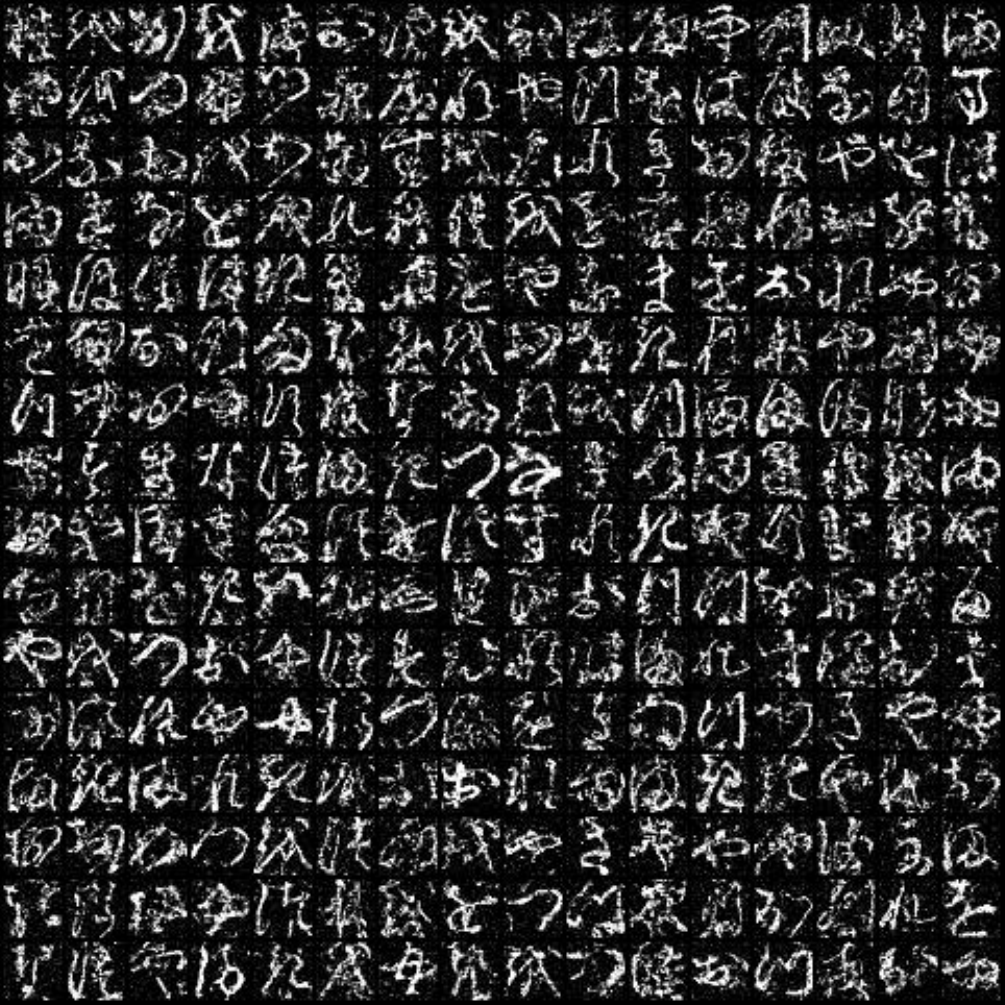}
    \caption{Kuzushiji MNIST}
  \end{subfigure}

  \vspace{1.5em}
  
  \begin{subfigure}[b]{0.49\textwidth}
    \includegraphics[width=\textwidth,]{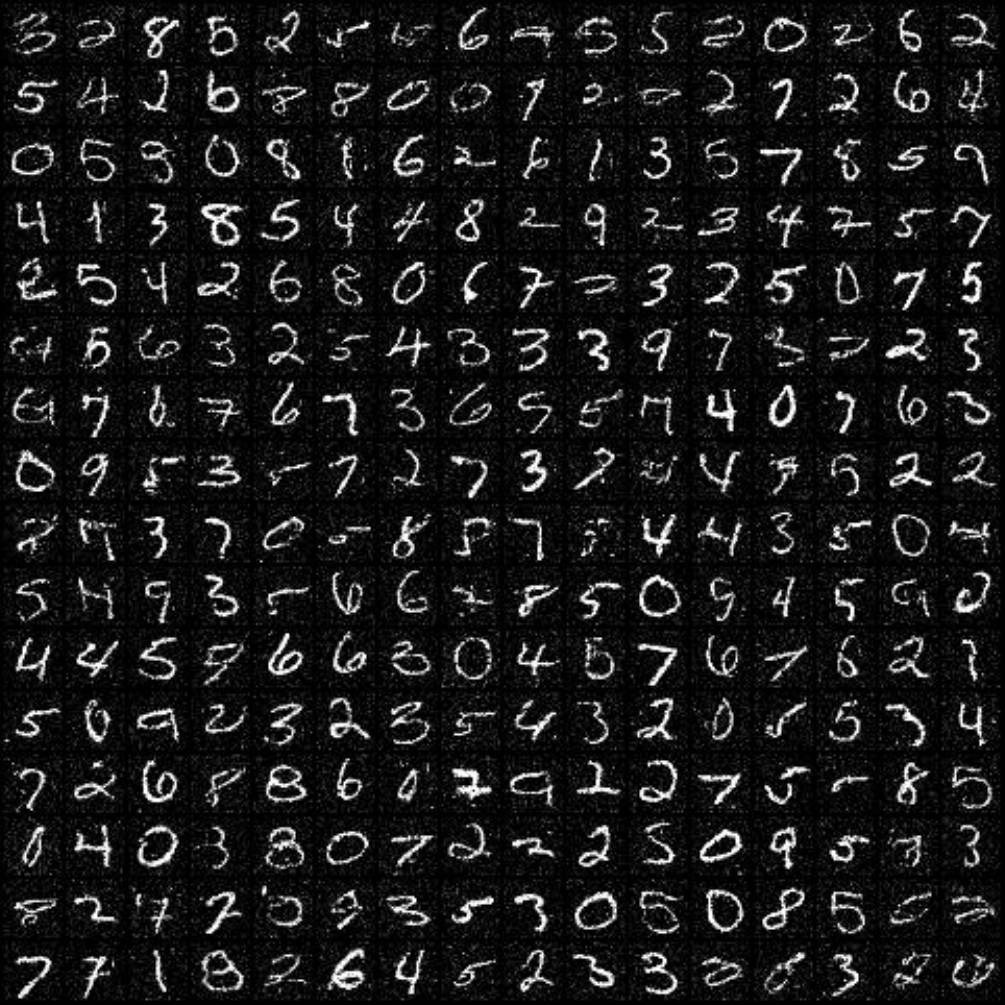}
    \caption{MNIST}
  \end{subfigure}
  \hfill
  \caption{Uncurated samples generated by the Dale-Langevin sampler \eqref{eqn:gbmsde_sampling_a} following the Annealed algorithm \ref{algo:gbm_sampler_anneal} with an annealing factor $\chi = 1.0$. Results are presented for Fashion-MNIST, Kuzushiji MNIST, and MNIST, respectively. Initialized with lognormal noise, the sampling follows the configurations ($L,\delta$) specified in \cref{tab:parameters_sampling}.}
\end{figure}

\begin{figure}[htbp]
  \centering
  \begin{subfigure}[b]{0.49\textwidth}
    \includegraphics[width=\textwidth,height=0.45\textheight, keepaspectratio]{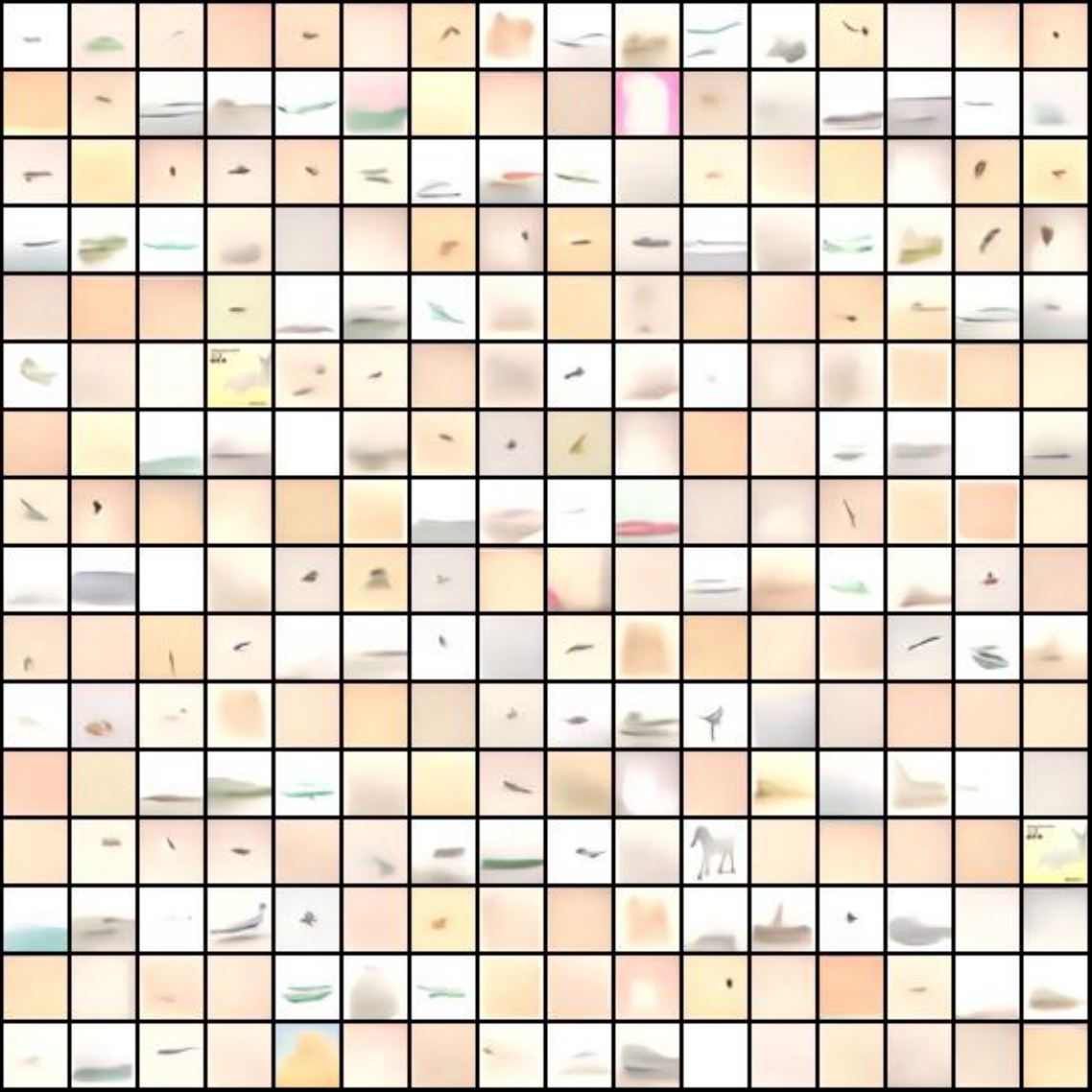}
    \caption{CIFAR-10 ($\chi=0.995$)}
  \end{subfigure}
  \begin{subfigure}[b]{0.49\textwidth}
    \includegraphics[width=\textwidth,height=0.45\textheight, keepaspectratio]{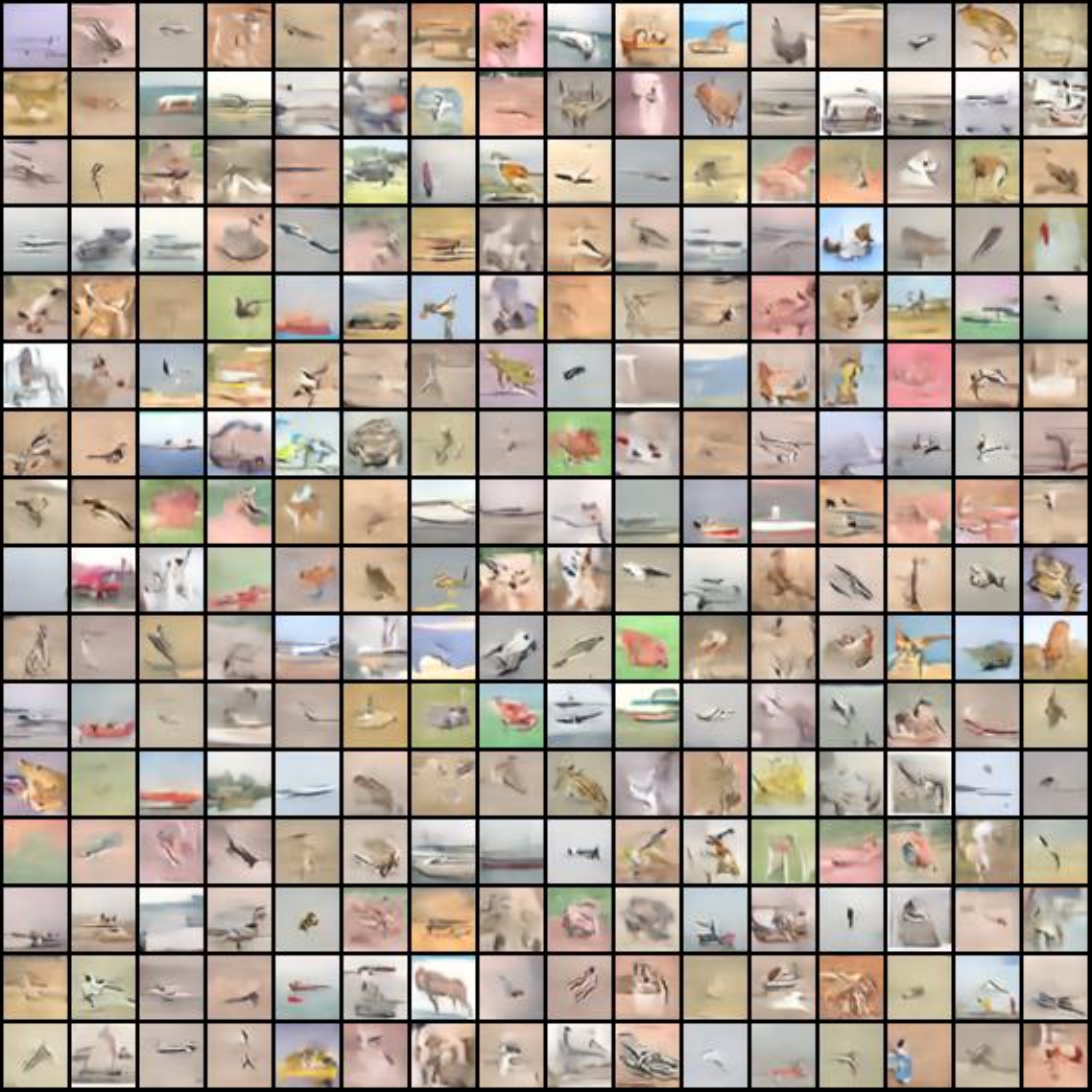}
    \caption{CIFAR-10 ($\chi=0.9995$)}
  \end{subfigure}
  \vspace{1.5em}
  \begin{subfigure}[b]{0.49\textwidth}
    \includegraphics[width=\textwidth,height=0.45\textheight, keepaspectratio]{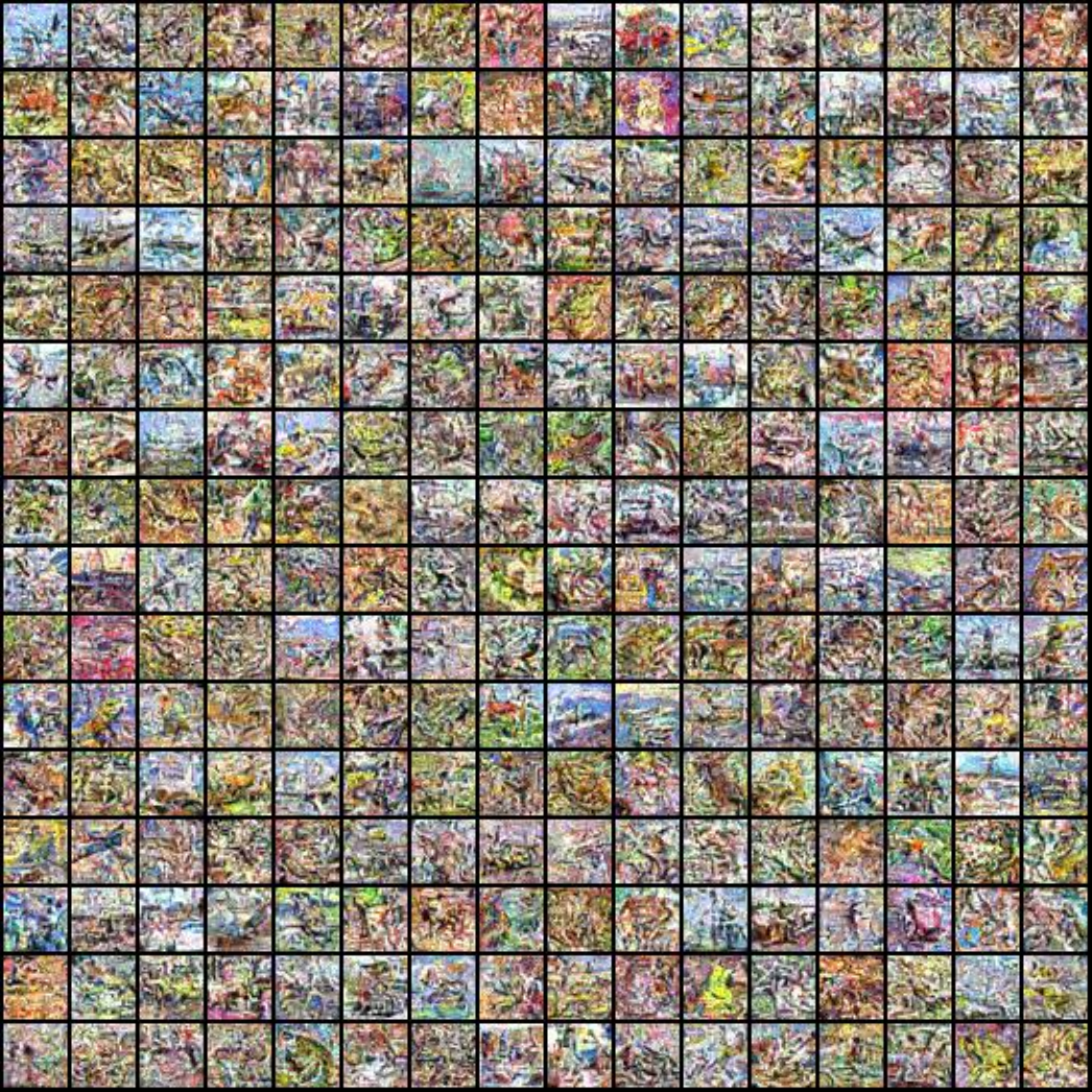}
    \caption{CIFAR-10 ($\chi=1.0$)}
  \end{subfigure}

  \caption{Uncurated samples generated by the sign-agnostic multiplicative sampler \cref{eqn:gbmsde_backward_sampling} using the Annealed~\cref{algo:gbm_sampler_nlamp_anneal} for the CIFAR-10 dataset. Results are presented for annealing factors $\chi \in \{0.995, 0.9995, 1.0\}$. Initialized from class-averaged images passed through the forward process, the sampling follows the configurations ($L,\delta$) detailed in \cref{tab:parameters_sampling}.}
\end{figure}

\begin{figure}[htbp]
  \centering
  \begin{subfigure}[b]{0.49\textwidth}
    \includegraphics[width=\textwidth,height=0.45\textheight, keepaspectratio]{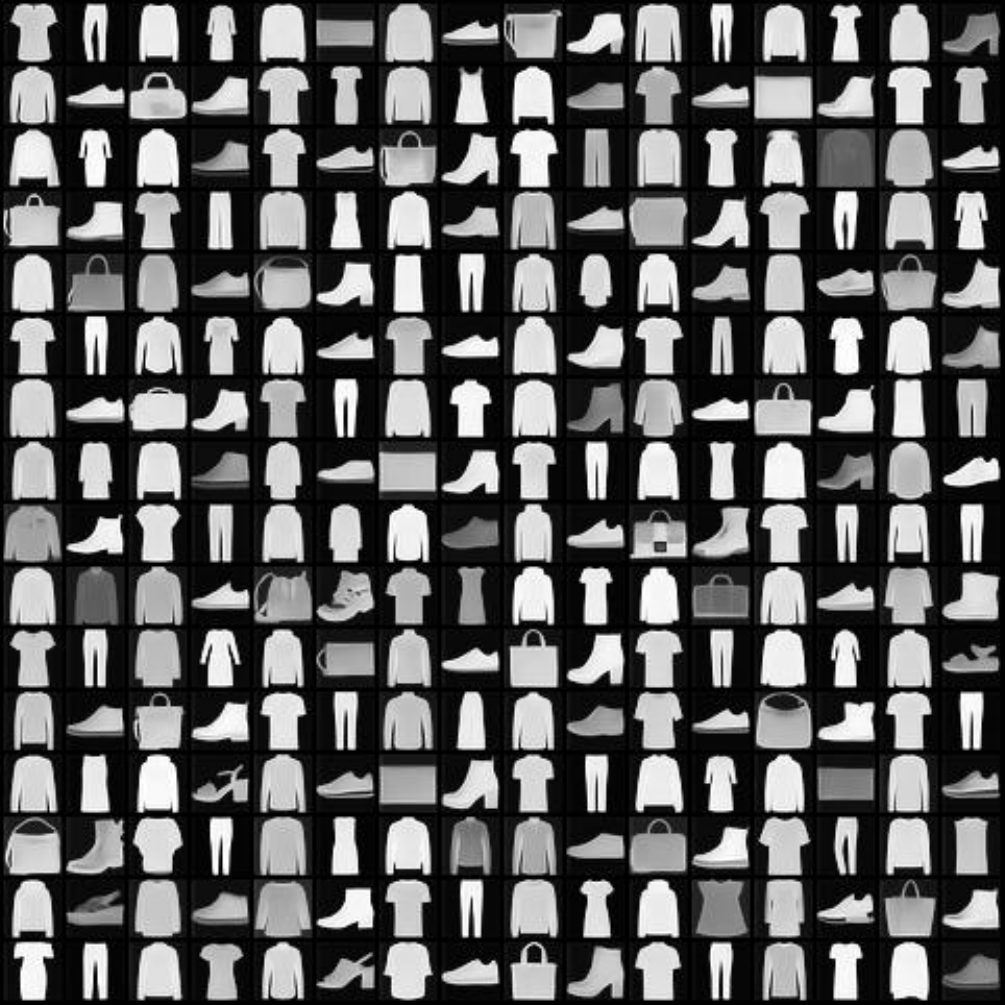}
    \caption{Fashion-MNIST}
  \end{subfigure}
  \begin{subfigure}[b]{0.49\textwidth}
    \includegraphics[width=\textwidth,height=0.45\textheight, keepaspectratio]{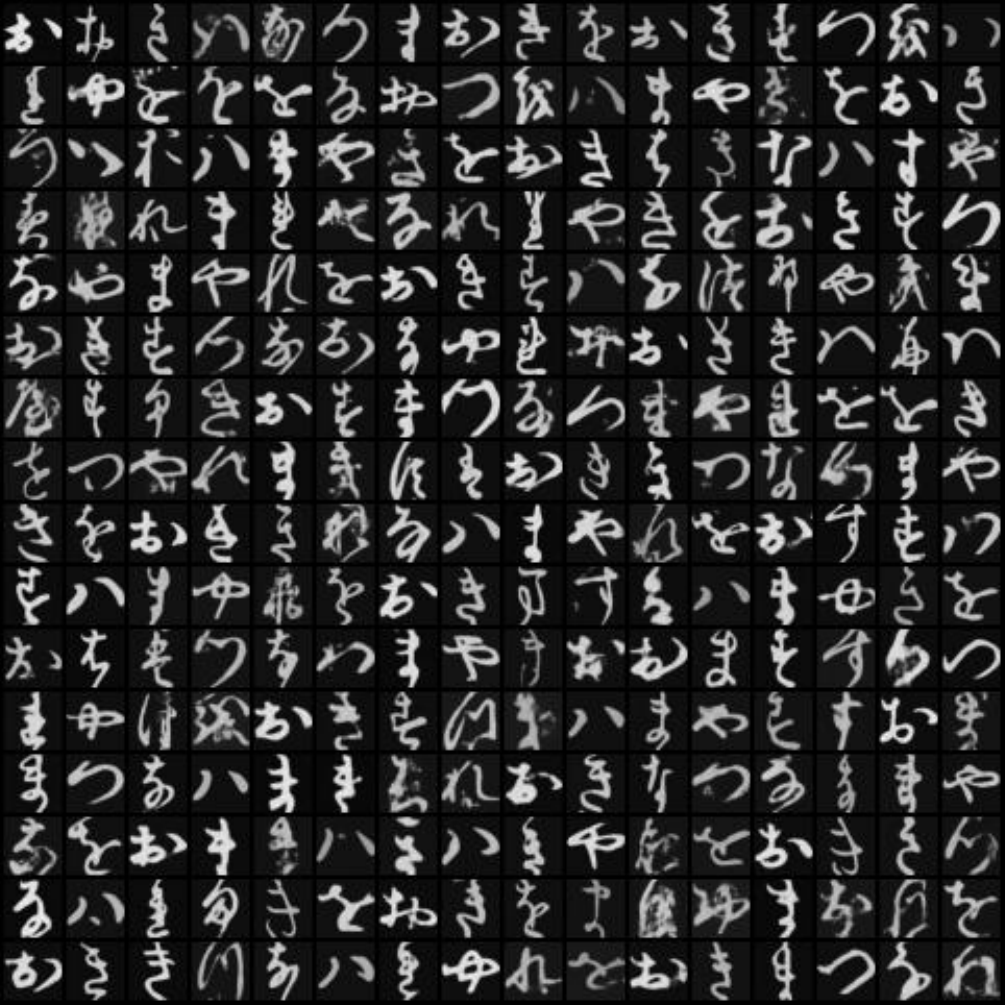}
    \caption{Kuzushiji MNIST}
  \end{subfigure}
  \vspace{1.5em}
  \begin{subfigure}[b]{0.49\textwidth}
    \includegraphics[width=\textwidth,height=0.45\textheight, keepaspectratio]{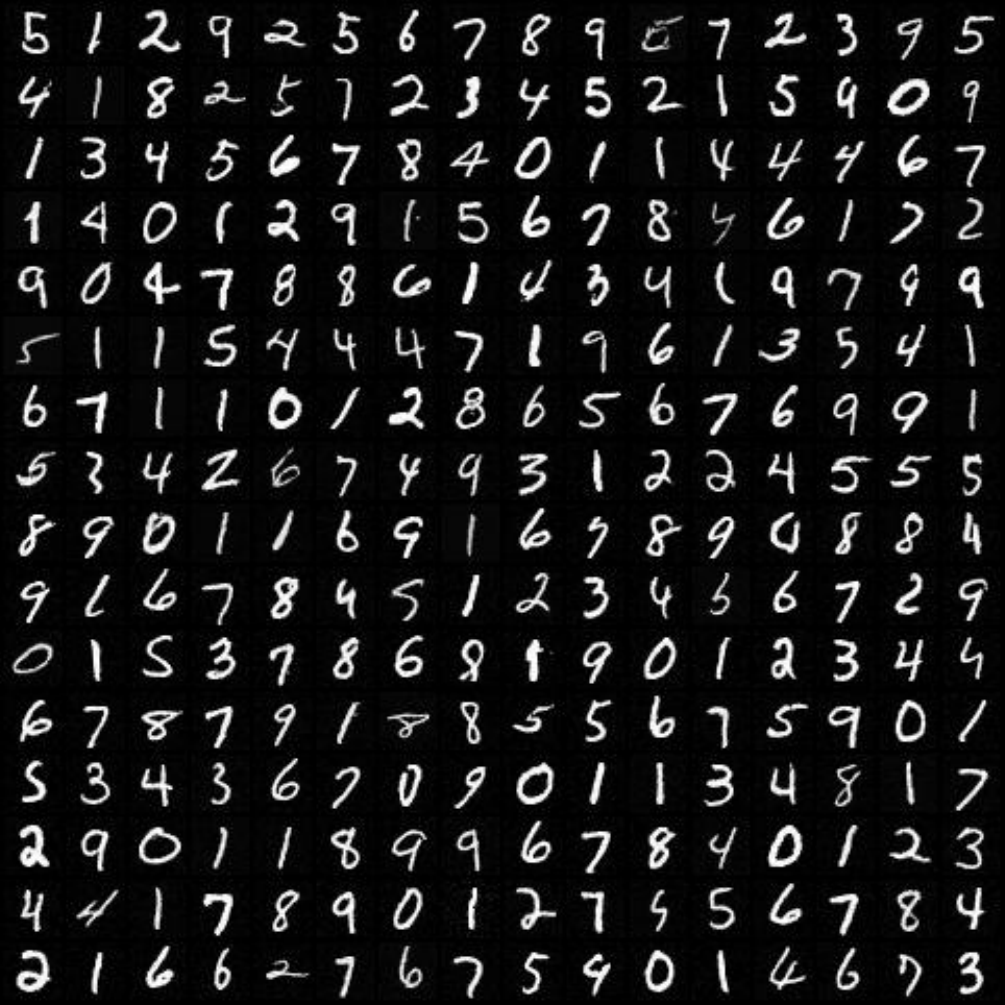}
    \caption{MNIST}
  \end{subfigure}

  \caption{Uncurated samples generated by the sign-agnostic multiplicative sampler \cref{eqn:gbmsde_backward_sampling} using the Annealed algorithm \ref{algo:gbm_sampler_nlamp_anneal} with an annealing factor $\chi = 0.995$. Results are displayed for Fashion-MNIST, Kuzushiji MNIST, and MNIST, respectively. Each process was initialized by passing a class-averaged image through the forward process, following the configurations ($L,\delta$) in \cref{tab:sign_agnostic_class_avg}.}
\end{figure}

\begin{figure}[htbp]
  \centering
  \begin{subfigure}[b]{0.49\textwidth}
    \includegraphics[width=\textwidth]{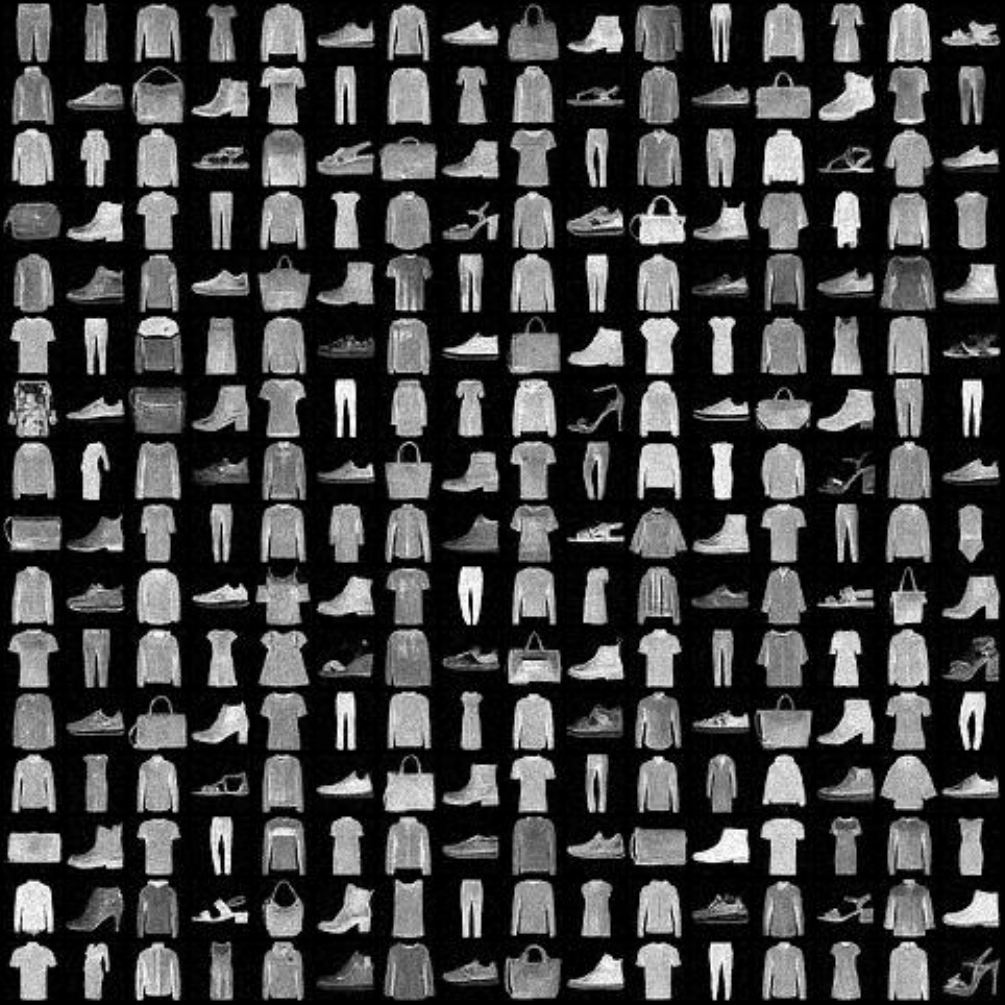}
    \caption{Fashion-MNIST}
  \end{subfigure}
  \begin{subfigure}[b]{0.49\textwidth}
    \includegraphics[width=\textwidth,height=0.45\textheight, keepaspectratio]{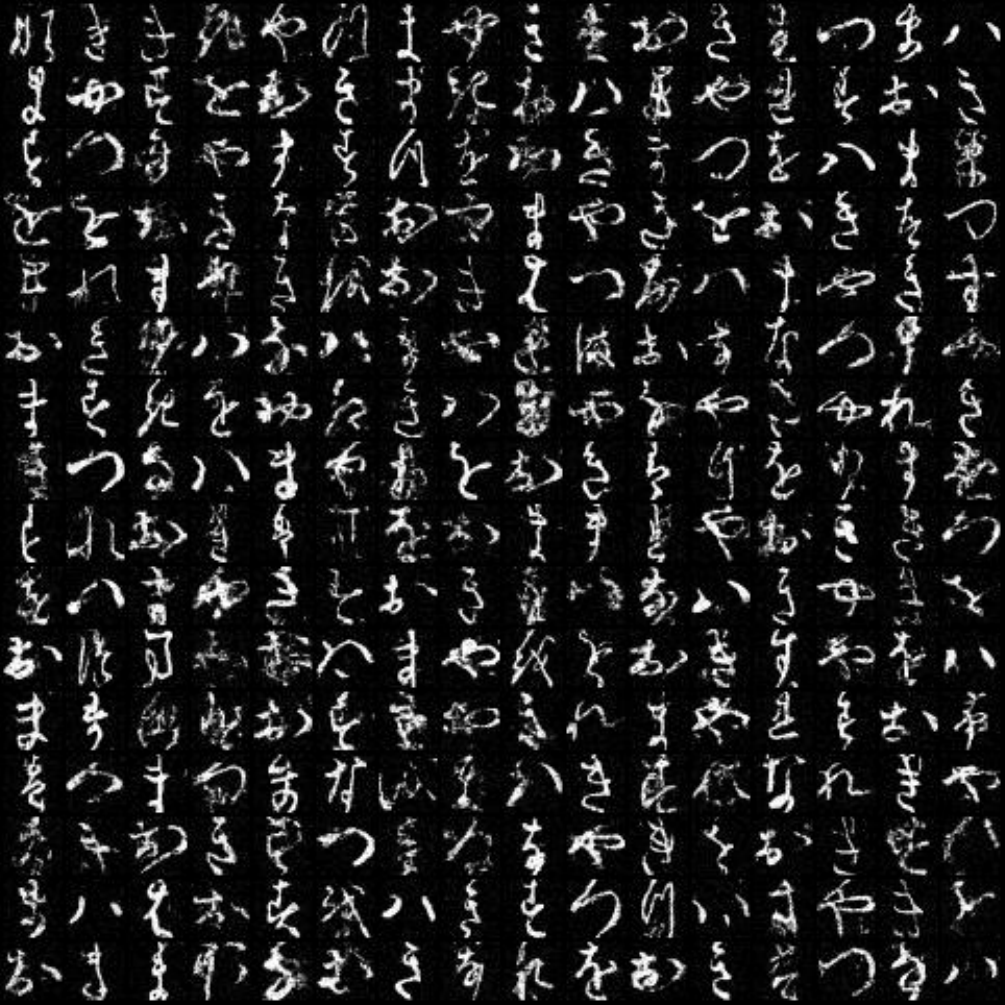}
    \caption{Kuzushiji MNIST}
  \end{subfigure}
  \vspace{1.5em}
  \begin{subfigure}[b]{0.49\textwidth}
    \includegraphics[width=\textwidth,height=0.45\textheight, keepaspectratio]{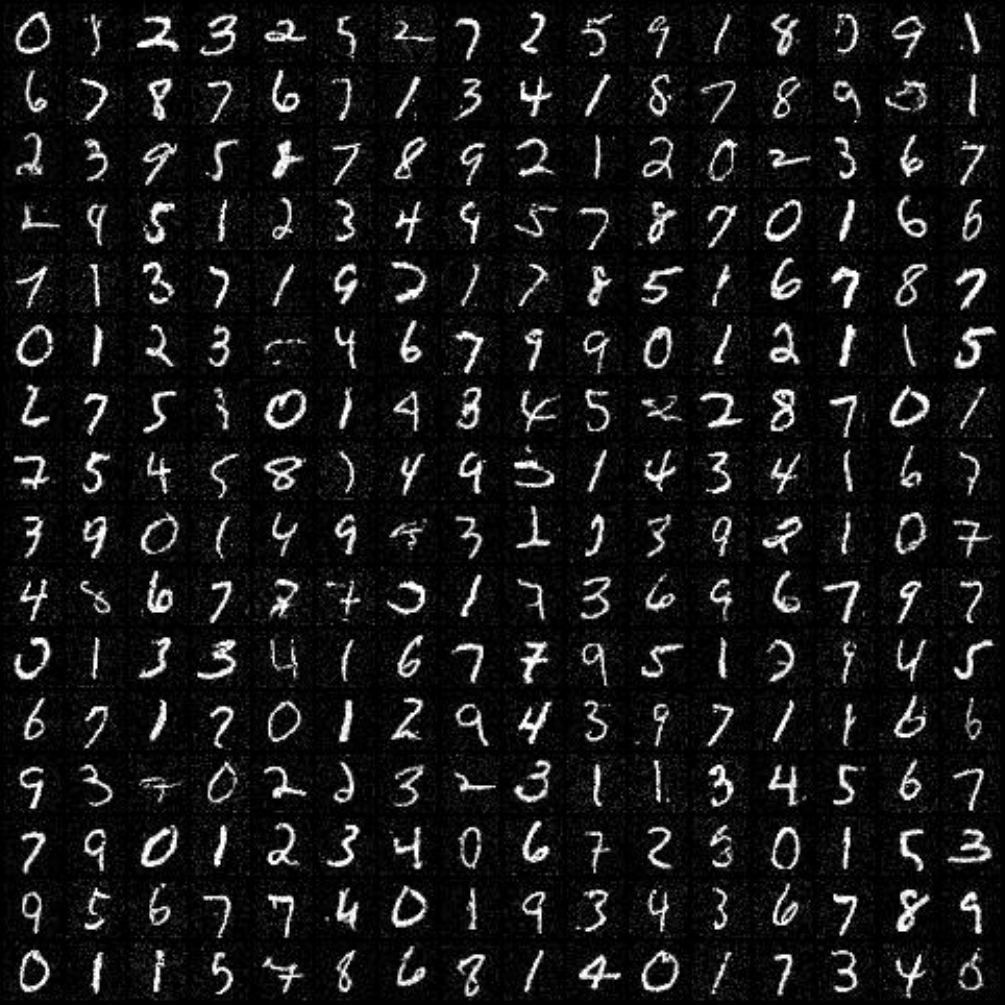}
    \caption{MNIST}
  \end{subfigure}

  \caption{Uncurated samples generated by the sign-agnostic multiplicative sampler \cref{eqn:gbmsde_backward_sampling} using the Annealed~\cref{algo:gbm_sampler_nlamp_anneal} with an annealing factor $\chi = 0.9995$. Results are displayed for Fashion-MNIST, Kuzushiji MNIST, and MNIST, respectively. Each process was initialized by passing a class-averaged image through the forward process, following the configurations ($L,\delta$) in \cref{tab:sign_agnostic_class_avg}.}
\end{figure}

\begin{figure}[htbp]
  \centering
  \begin{subfigure}[b]{0.49\textwidth}
    \includegraphics[width=\textwidth,height=0.45\textheight, keepaspectratio]{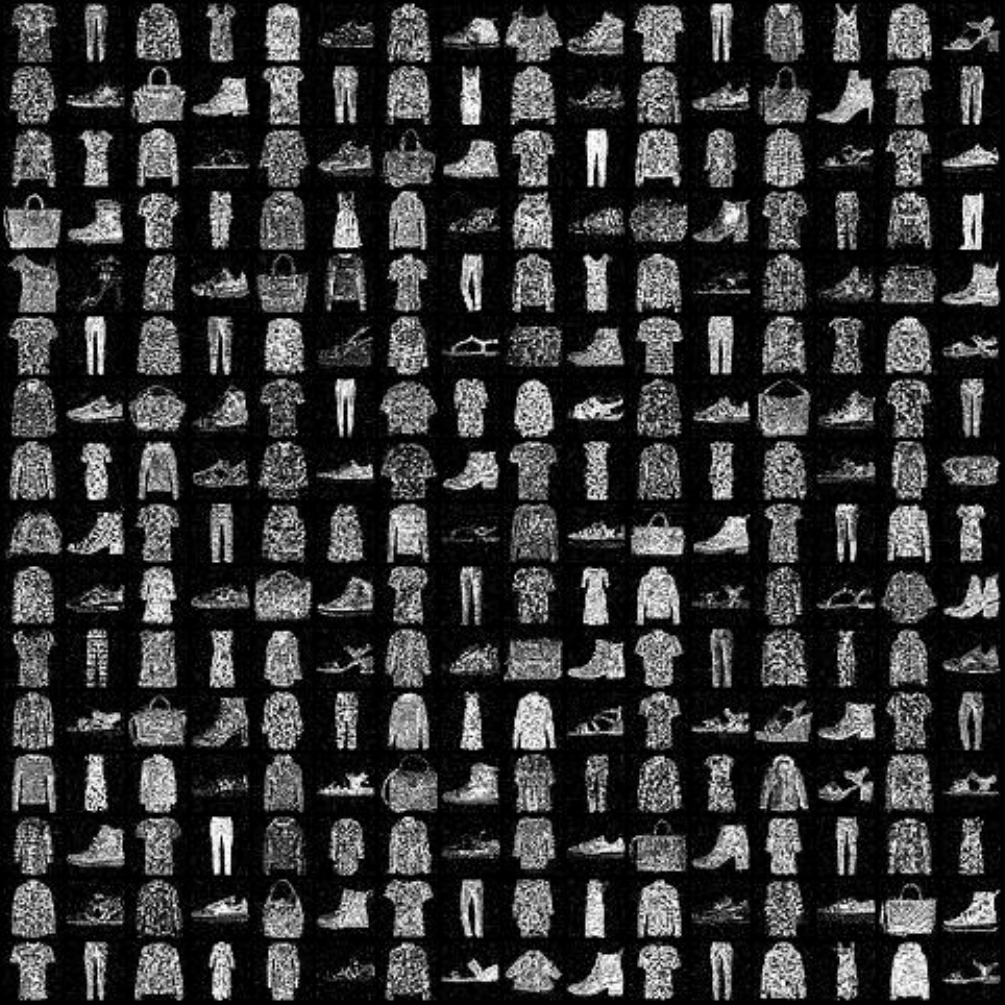}
    \caption{Fashion-MNIST}
  \end{subfigure}
  \begin{subfigure}[b]{0.49\textwidth}
    \includegraphics[width=\textwidth,height=0.45\textheight, keepaspectratio]{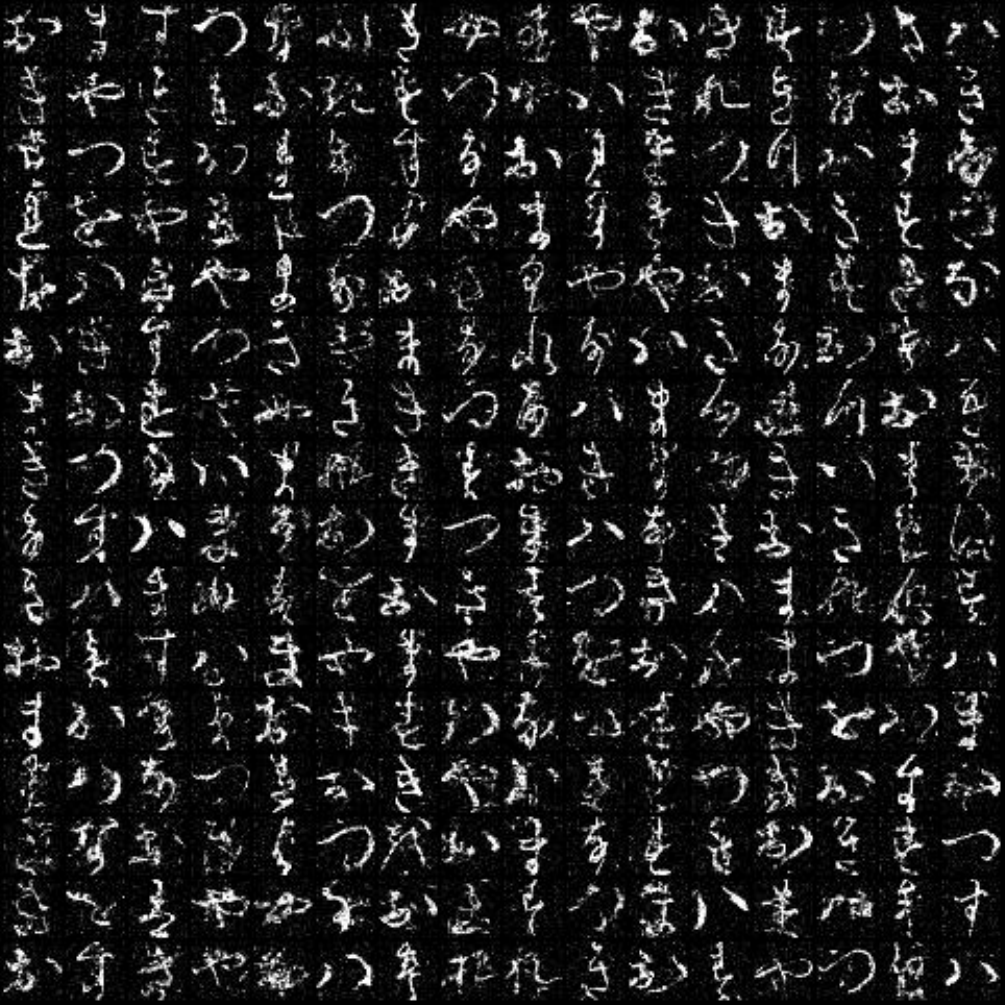}
    \caption{Kuzushiji MNIST}
  \end{subfigure}
  \vspace{1.5em}
  \begin{subfigure}[b]{0.49\textwidth}
    \includegraphics[width=\textwidth,height=0.45\textheight, keepaspectratio]{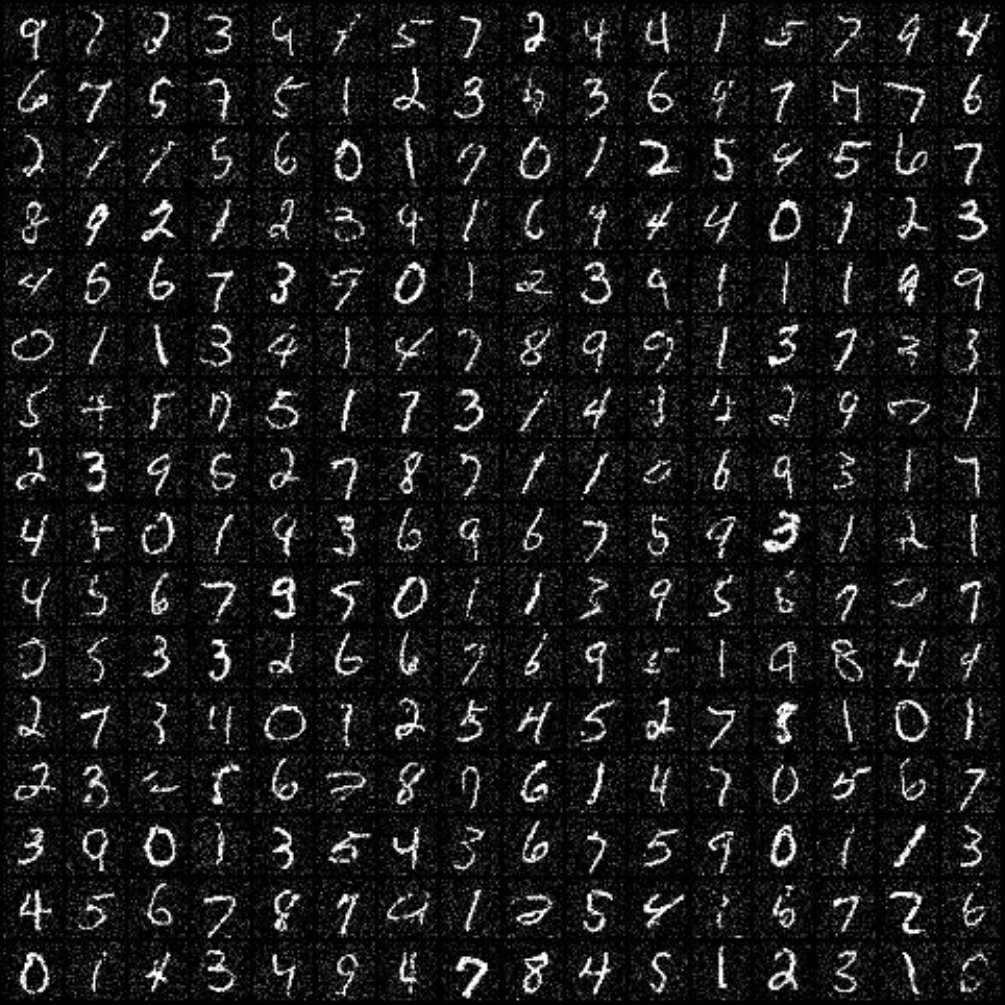}
    \caption{MNIST}
  \end{subfigure}
  
  \caption{Uncurated samples generated by the sign-agnostic multiplicative sampler \cref{eqn:gbmsde_backward_sampling} using the Annealed~\cref{algo:gbm_sampler_nlamp_anneal} with an annealing factor $\chi = 1.0$. Results are displayed for Fashion-MNIST, Kuzushiji MNIST, and MNIST, respectively. Each process was initialized by passing a class-averaged image through the forward process, following the configurations ($L,\delta$) in \cref{tab:sign_agnostic_class_avg}.}
\end{figure}

\begin{figure}[htbp]
  \centering
  \begin{subfigure}[b]{0.49\textwidth}
    \includegraphics[width=\textwidth,height=0.45\textheight, keepaspectratio]{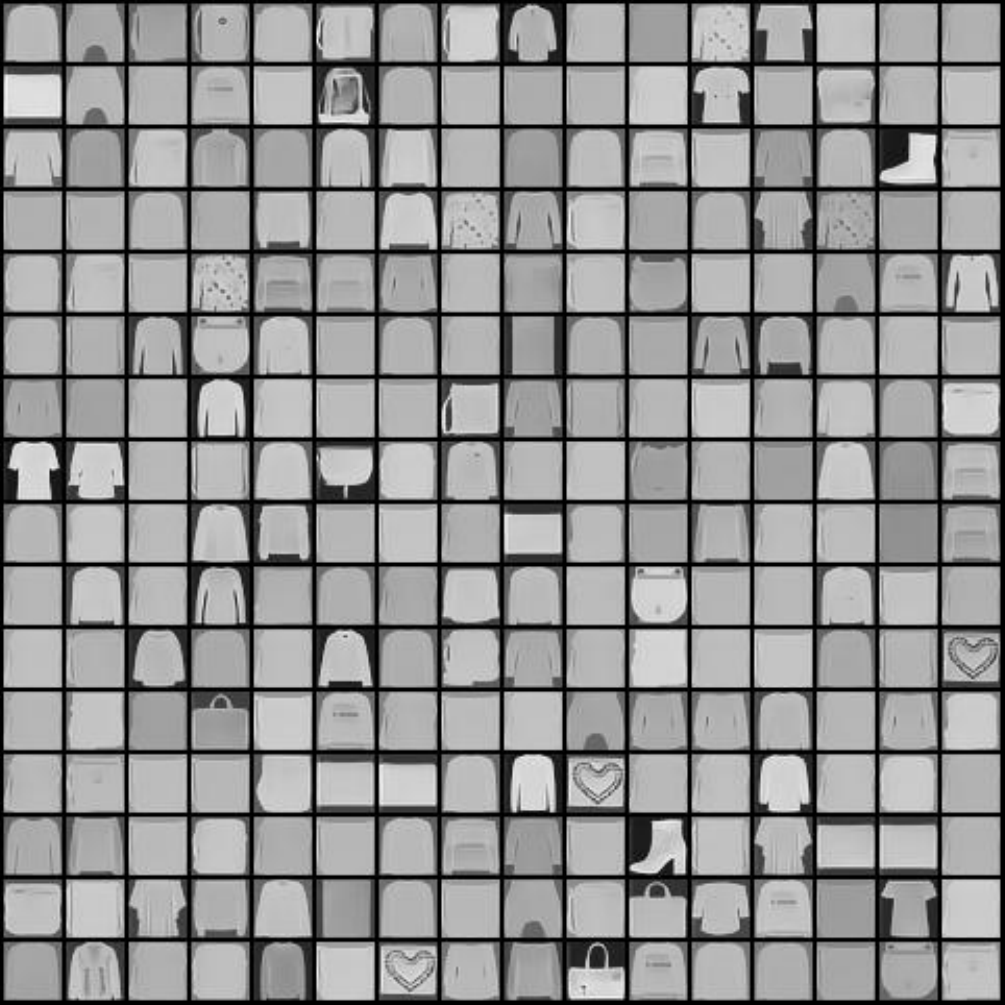}
    \caption{Fashion-MNIST}
  \end{subfigure}
  \begin{subfigure}[b]{0.49\textwidth}
    \includegraphics[width=\textwidth,height=0.45\textheight, keepaspectratio]{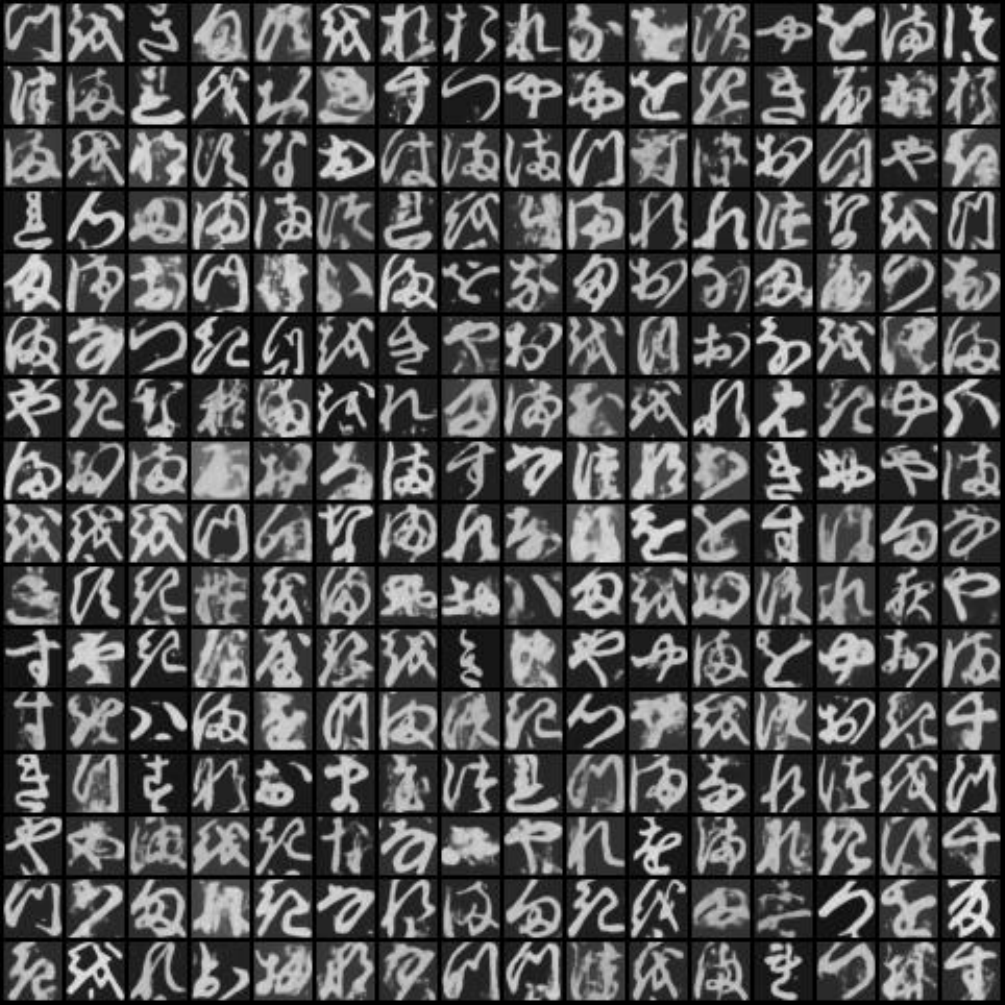}
    \caption{Kuzushiji MNIST}

  \end{subfigure}
   \vspace{1.5em}
  \begin{subfigure}[b]{0.49\textwidth}
    \includegraphics[width=\textwidth,height=0.45\textheight, keepaspectratio]{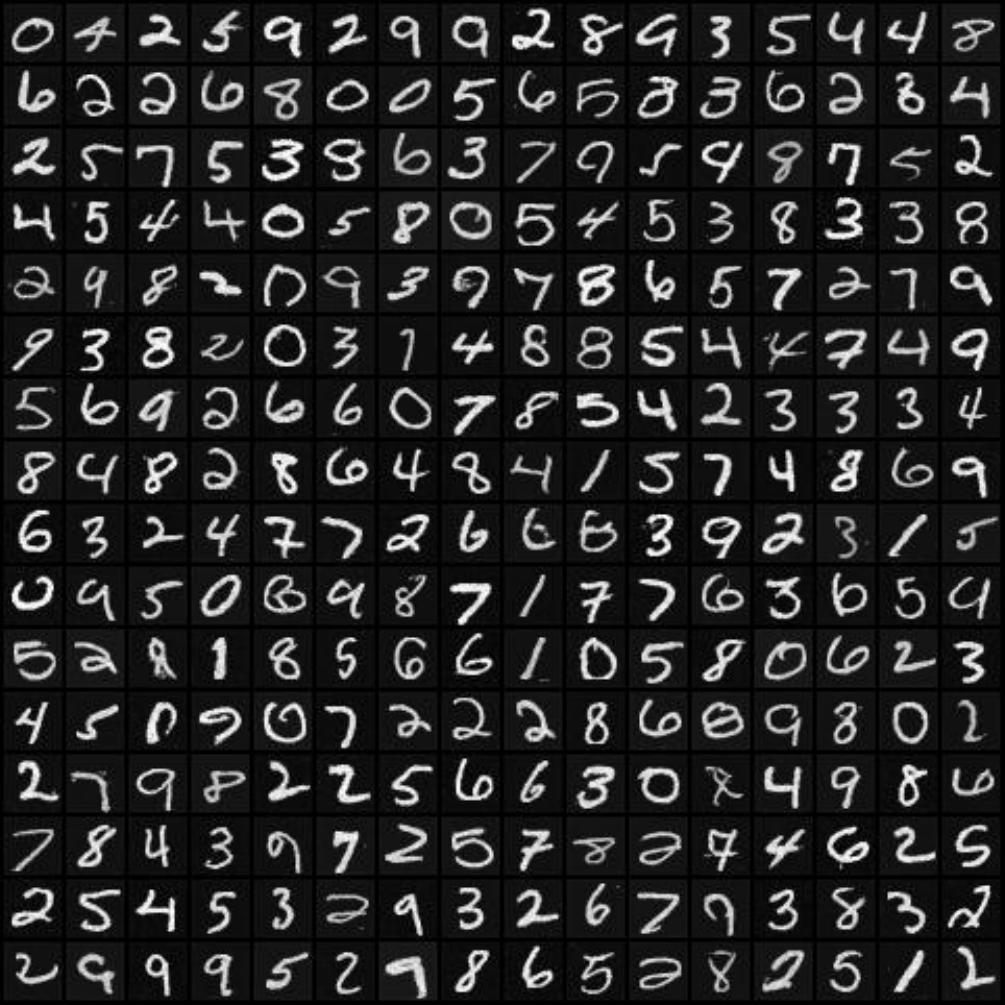}
    \caption{MNIST}
  \end{subfigure}

  \caption{Uncurated samples generated by the sign-agnostic multiplicative sampler \cref{eqn:gbmsde_backward_sampling} using the Annealed~\cref{algo:gbm_sampler_nlamp_anneal} with an annealing factor $\chi = 0.995$. Results are displayed for Fashion-MNIST, Kuzushiji MNIST, and MNIST, respectively,  initialized with lognormal noise, following the configurations ($L,\delta$) detailed in \cref{tab:parameters_sampling}.}
\end{figure}

\begin{figure}[htbp]
  \centering
  \begin{subfigure}[b]{0.49\textwidth}
    \centering
    \includegraphics[width=\textwidth, height=0.45\textheight, keepaspectratio]{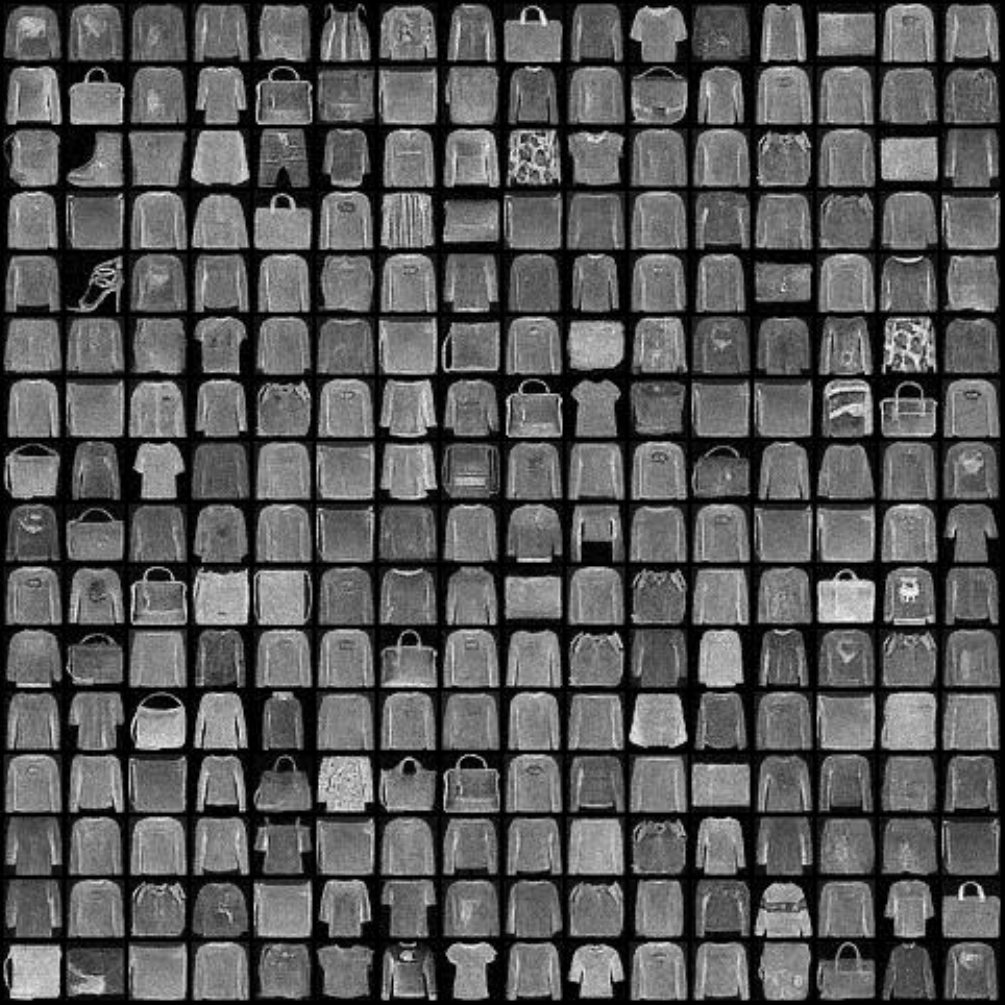}
    \caption{Fashion-MNIST}
  \end{subfigure}
  \hfill
  \begin{subfigure}[b]{0.49\textwidth}
    \includegraphics[width=\textwidth, height=0.45\textheight, keepaspectratio]{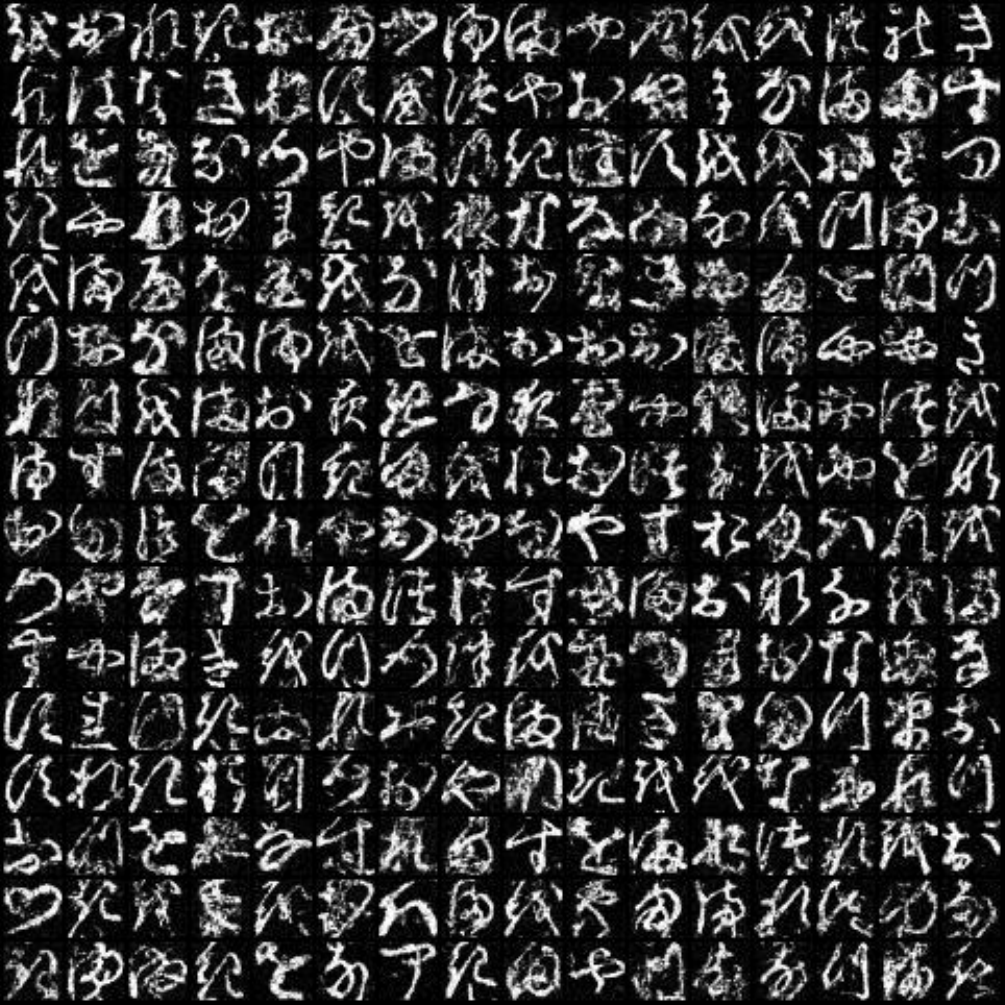}
    \caption{Kuzushiji MNIST}
  \end{subfigure}
  \vspace{1.5em}
  \begin{subfigure}[b]{0.49\textwidth}
    \includegraphics[width=\textwidth,height=0.45\textheight, keepaspectratio]{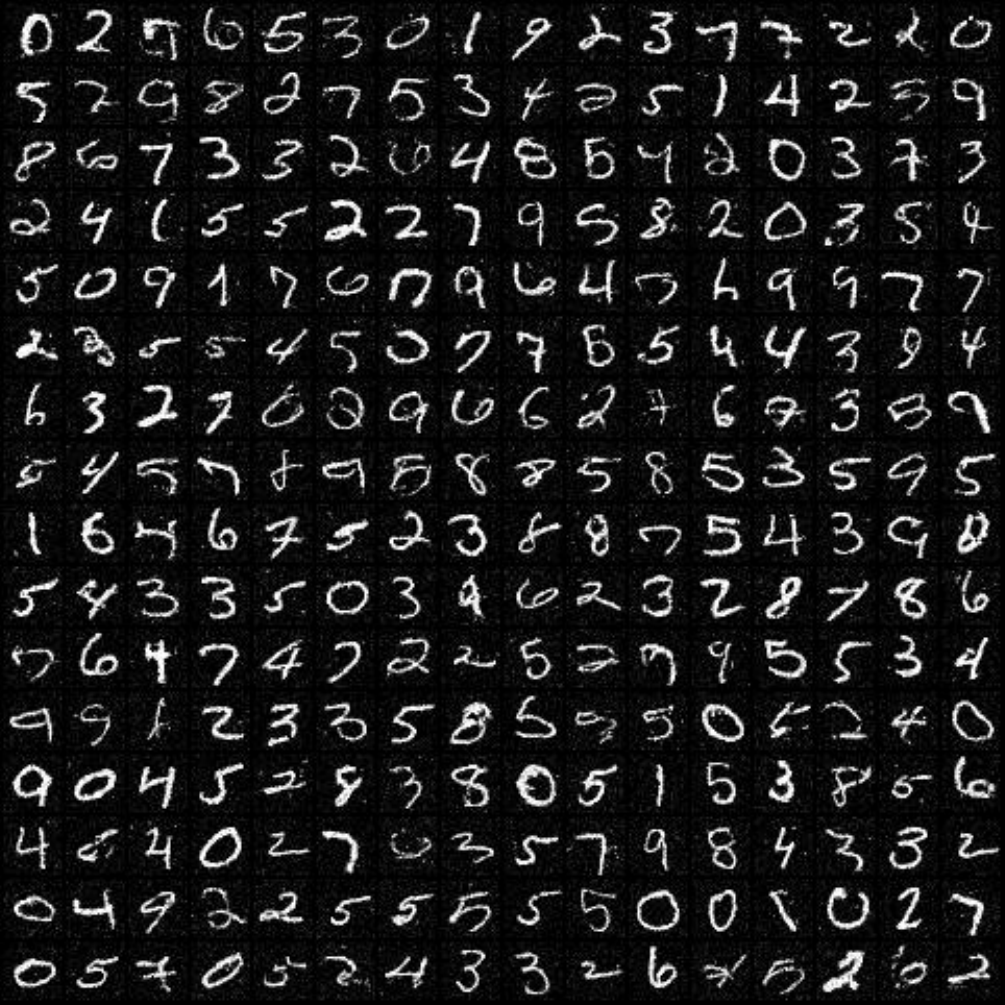}
    \caption{MNIST}
  \end{subfigure}

  \caption{Uncurated samples generated by the sign-agnostic multiplicative sampler \cref{eqn:gbmsde_backward_sampling} using the Annealed~\cref{algo:gbm_sampler_nlamp_anneal} with an annealing factor $\chi = 0.9995$. Results are displayed for Fashion-MNIST, Kuzushiji MNIST, and MNIST, respectively, initialized with lognormal noise, following the configurations ($L,\delta$) detailed in \cref{tab:parameters_sampling}.}
\end{figure}

\begin{figure}[htbp]
  \centering
  \begin{subfigure}[b]{0.49\textwidth}
    \centering
    \includegraphics[width=\textwidth, height=0.45\textheight, keepaspectratio ]{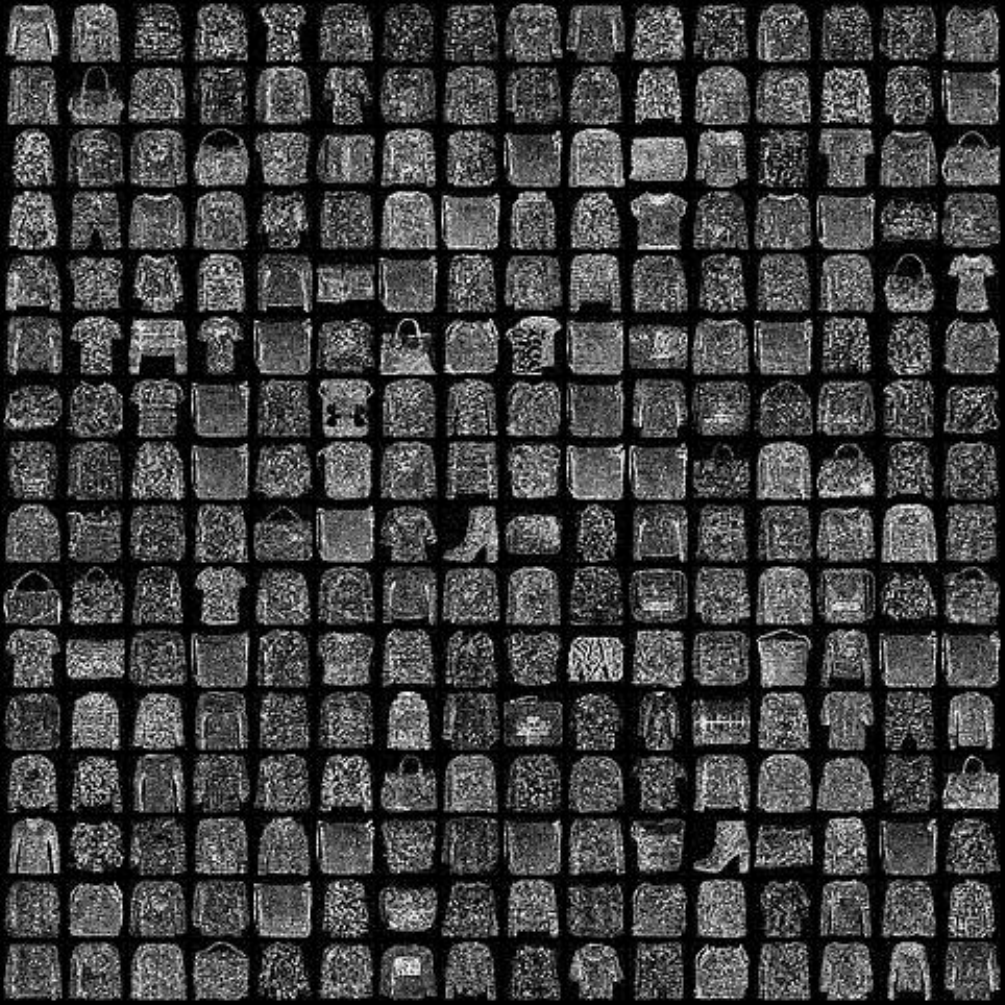}
    \caption{Fashion-MNIST}
  \end{subfigure}
  \hfill
  \begin{subfigure}[b]{0.49\textwidth}
    \includegraphics[width=\textwidth, height=0.45\textheight, keepaspectratio]{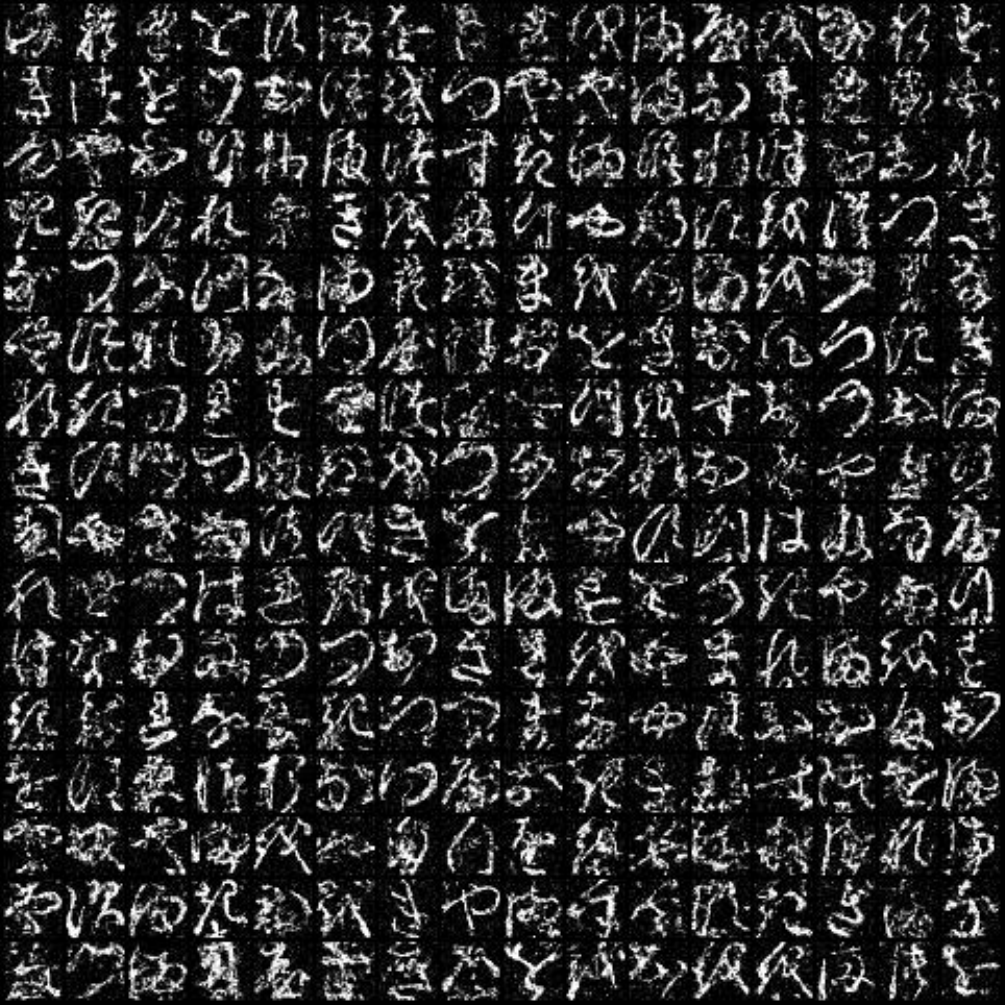}
    \caption{Kuzushiji MNIST}
  \end{subfigure}

  \vspace{1.5em}

  \begin{subfigure}[b]{0.49\textwidth}
    \includegraphics[width=\textwidth, height=0.45\textheight, keepaspectratio]{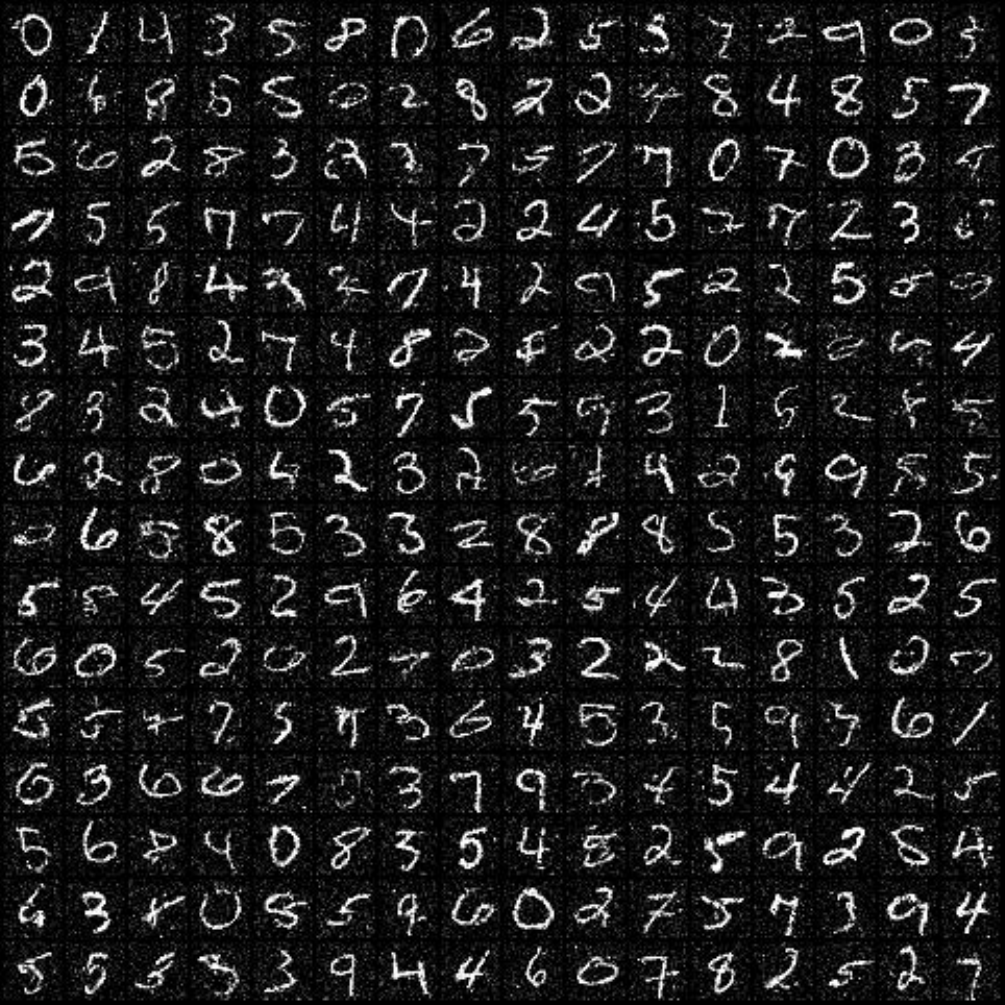}
    \caption{MNIST}
  \end{subfigure}

  \caption{Uncurated samples generated by the sign-agnostic multiplicative sampler \cref{eqn:gbmsde_backward_sampling} using the Annealed~\cref{algo:gbm_sampler_nlamp_anneal} with an annealing factor $\chi = 1.0$. Results are displayed for Fashion-MNIST, Kuzushiji MNIST, and MNIST, respectively, initialized with lognormal noise, following the configurations ($L,\delta$) detailed in \cref{tab:parameters_sampling}.}
\end{figure}

\begin{figure}[!ht]
  \centering
  
  \begin{subfigure}[b]{0.49\textwidth}
    \centering
    \includegraphics[width=\textwidth, height=0.45\textheight, keepaspectratio]{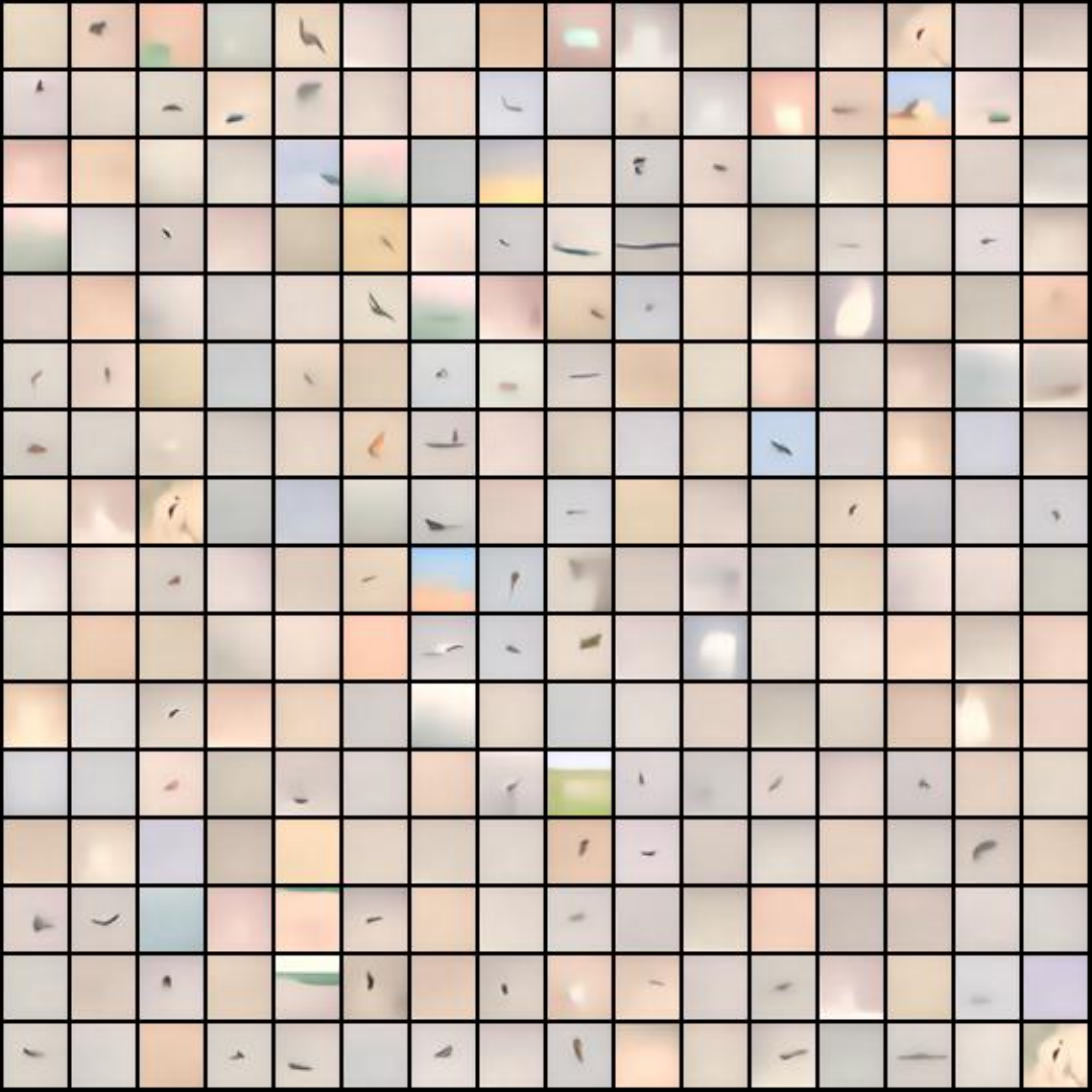}
    \caption{CIFAR-10 ($\chi=0.995$)}
  \end{subfigure}
  \hfill
  \begin{subfigure}[b]{0.49\textwidth}
    \centering
    \includegraphics[width=\textwidth, height=0.45\textheight, keepaspectratio]{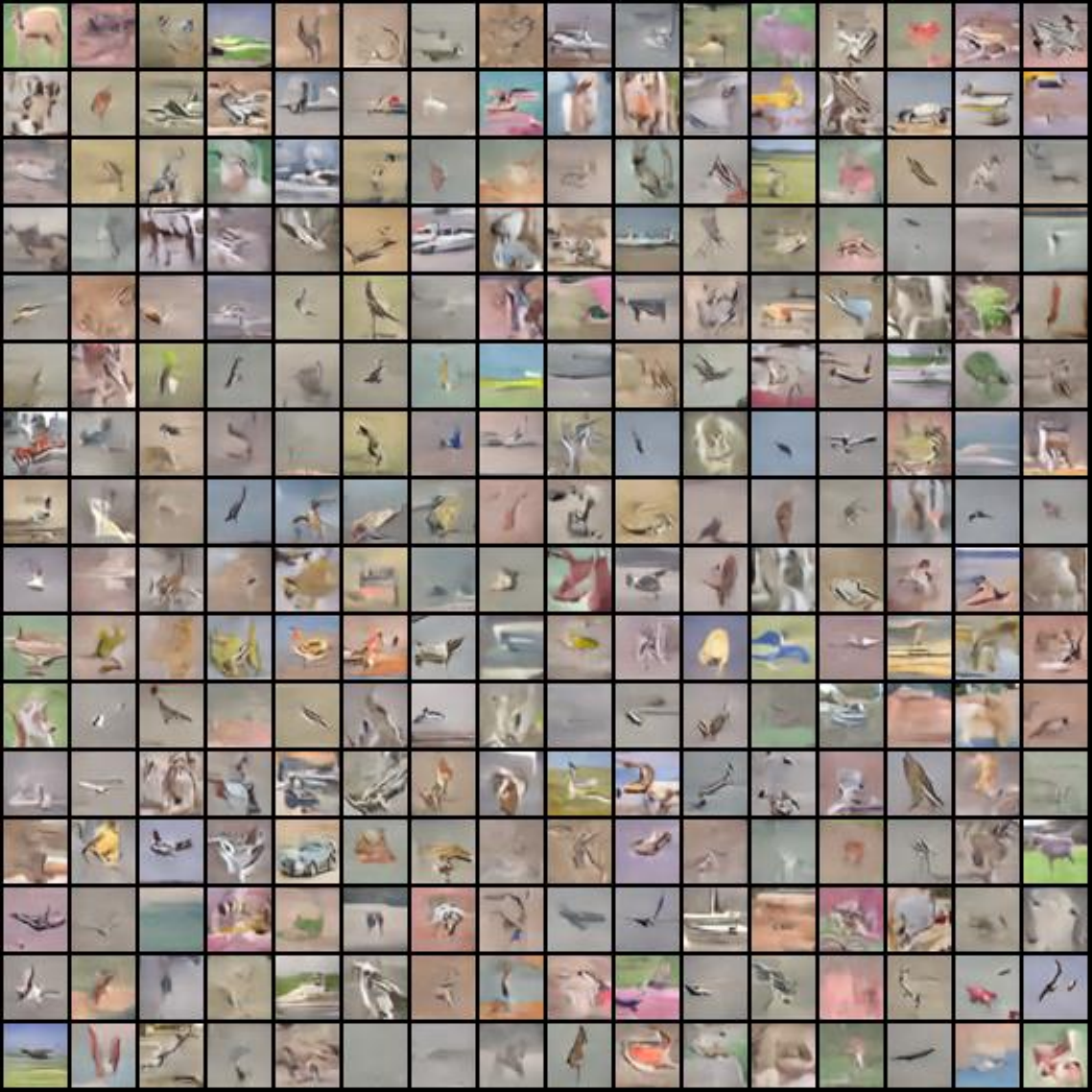}
    \caption{CIFAR-10 ($\chi=0.9995$)}
  \end{subfigure}
  
  \vspace{1.5em} 
  
  \begin{subfigure}[b]{0.49\textwidth}
    \centering
    \includegraphics[width=\textwidth, height=0.45\textheight, keepaspectratio]{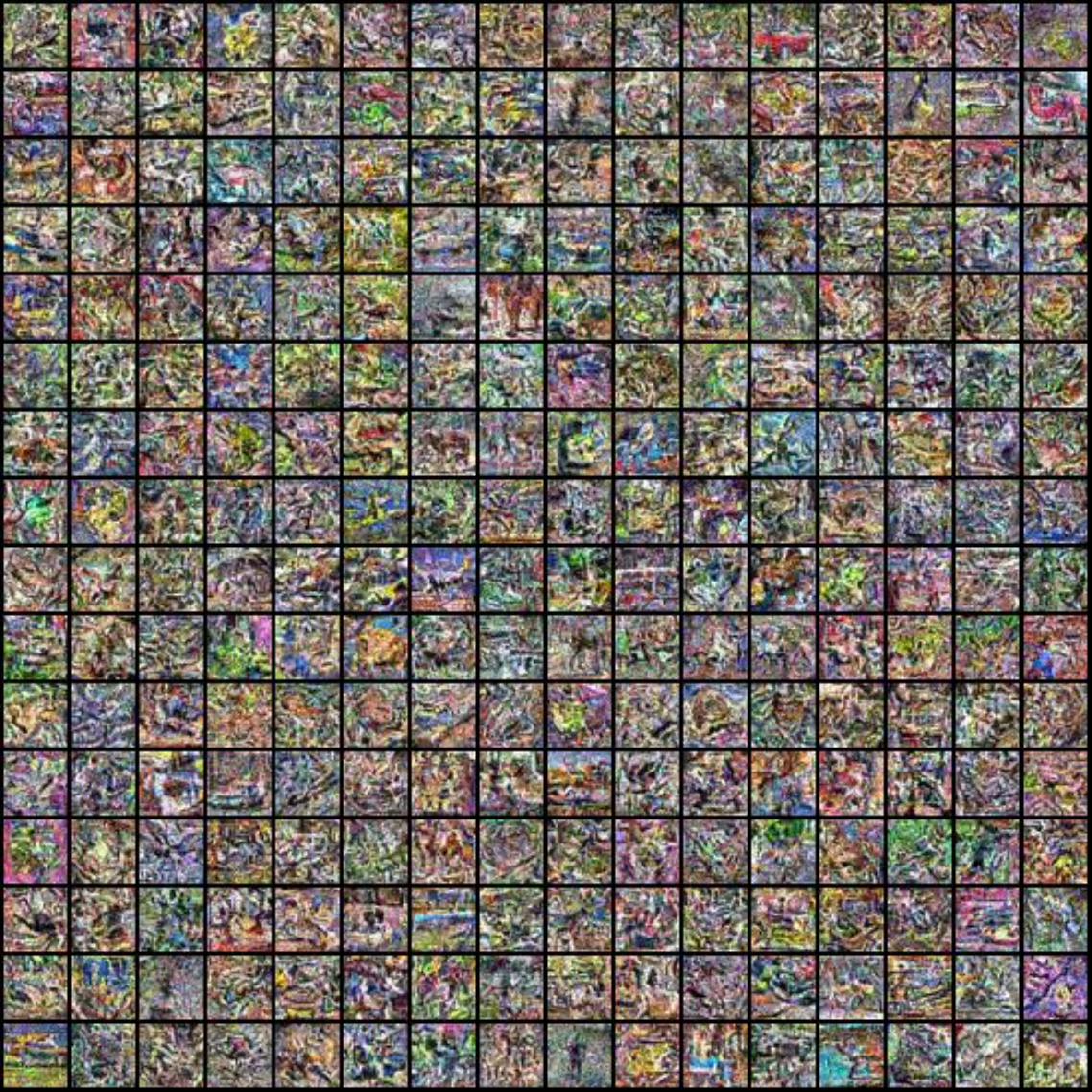}
    \caption{CIFAR-10 ($\chi=1.0$)}
  \end{subfigure}
  
  \caption{Uncurated samples generated by the sign-agnostic multiplicative sampler \cref{eqn:gbmsde_backward_sampling} using the Annealed~\cref{algo:gbm_sampler_nlamp_anneal} for the CIFAR-10 dataset. Results are presented for annealing factors $\chi \in \{0.995, 0.9995, 1.0\}$. Initialized from lognormal Noise the sampling follows the configurations ($L,\delta$) detailed in \cref{tab:parameters_sampling}.}
\end{figure}

\end{document}